\documentclass[12pt]{article} 

\usepackage[left= 0.8in, textwidth=6.9in, textheight=9.0in, top=1.0in]{geometry}

\usepackage{graphicx}
\usepackage{amsmath,amssymb,verbatim}
\usepackage{color}  
\usepackage{bm}                                  

\newcommand{\MT}{\left[ \begin{array}{rrrrrrrrrrrrrrrrrrrr}}
\newcommand{\EM}{\end{array}\right]}
\newcommand{\EQ}{\begin{equation}\begin{array}{lllllllllll}}
\newcommand{\EE}{\end{array}\end{equation}}
\newcommand{\Real}{\mathbb R}
\newcommand\norm[1]{\left\lVert#1\right\rVert}
\newcommand\abs[1]{\left|#1\right|}

\newtheorem{theorem}{Theorem}

\newtheorem{remark}{Remark}
\newtheorem{definition}{Definition}
\newtheorem{proposition}{Proposition}

\newtheorem{corollary}{Corollary}
\newtheorem{example}{Example}

\def\Fr{\ds \frac}
\def\ds{\displaystyle}
\def\calG{{\cal G}}
\def\calL{{\cal L}}
\def\calV{{\cal V}}
\def\calE{{\cal E}}

\def\calU{{\cal U}}
\def\Dlt{{\it \Delta}}
\def\bff{{\bf f}}
\def\bfg{{\bf g}}
\def\bfh{{\bf h}}
\def\bfx{{\bf x}}
\def\bfz{{\bf z}}
\def\bfu{{\bf u}}
\def\bfv{{\bf v}}
\def\bfw{{\bf w}}
\def\bfk{{\bf k}}

\def\bfpsi{{\boldsymbol\psi}}
\def\bfphi{{\boldsymbol\phi}}

\title{\LARGE \bf
Neural Network Approximations of Compositional Functions With Applications to Dynamical Systems
\thanks{This work was supported in part by U.S. Naval Research Laboratory - Monterey, CA}
}

\author{Wei Kang\thanks{Department of Applied Mathematics, Naval Postgraduate School, Monterey, CA, USA; wkang@nps.edu}
\and Qi Gong\thanks{Department of Applied Mathematics, University of California, Santa Cruz, Santa Cruz, CA, USA; qgong@ucsc.edu}
}

\begin{document}

\maketitle

\begin{abstract}
As demonstrated in many areas of real-life applications, neural networks have the capability of dealing with high dimensional data. In the fields of optimal control and dynamical systems, the same capability was studied and verified in many published results in recent years. Towards the goal of revealing the underlying reason why neural networks are capable of solving some high dimensional problems, we develop an algebraic framework and an approximation theory for compositional functions and their neural network approximations. The theoretical foundation is developed in a way so that it supports the error analysis for not only functions as input-output relations, but also numerical algorithms. This capability is critical because it enables the analysis of approximation errors for problems for which analytic solutions are not available, such as differential equations and optimal control. We identify a set of key features of compositional functions and the relationship between the features and the complexity of neural networks. In addition to function approximations, we prove several formulae of error upper bounds for neural networks that approximate the solutions to differential equations, optimization, and optimal control. 
\end{abstract}

\section{Introduction}
In recent years, the research at the intersection of the fields of optimal control, dynamical systems and machine learning have attracted rapidly increasing attention.  Progresses have been made at a fast pace, for instance \cite{hamzi0,grune,hamzi1,han1,han2,kang,zimmerer1,zimmerer2,raissi}. In these papers, machine learning based on artificial neural networks are used as an effective tool for finding approximate solutions to high dimensional problems of optimal control and partial differential equations. Many of these problems are not solvable using other numerical methods due to the curse of dimensionality, i.e. the solution's complexity grows exponentially when the dimension of the problem increases. Why do neural networks work well? Despite of years of research efforts, the fundamental reasons are far from clear. Reviews of some related topics and results can be found in \cite{weinan0,kainen}. It is proved in \cite{barron0} that the neural network approximation of a function whose gradient has an integrable Fourier transform can achieve an error upper bound $O(n^{-1/2})$, where $n>0$ is the number of neurons, i.e., the complexity of the neural network. This rate has a constant exponent that is independent of the problem's dimension, $d$. Therefore, the curse of dimensionality is mitigated. However, it is pointed out in \cite{barron0} that a constant $C$ in $O(n^{-1/2})$ can indirectly depend on $d$.  If $C$ is exponentially large in $d$, then an exponentially large value of the neural network's complexity would be required for the approximation to be bounded by a given error tolerance. Similar issues are also brought up in \cite{kainen} in a review of various neural network approximation methods. If $L^\infty$ norm is used to measure errors, which is the norm adopted in this paper, the complexity of deep neural networks for the approximation of compositional functions is bounded above by $O(\epsilon^{-r})$ for some constant $r>0$, where $\epsilon$ is the error tolerance \cite{mhaskar1,mhaskar2,poggio}. This theory is based on the assumption that complicated functions can be represented by the compositions of relatively simple functions. The upper bound of complexity, $O(\epsilon^{-r})$, has an unknown constant which can grow with the increase of the function's input dimension. The growth rate depends on various factors or features associated with the function. The relationship is not fully understood. What makes the problem even more challenging is that some functions studied in this paper are associated with dynamical systems such as the trajectory of differential equations or the feedback law of optimal control. For these problems, most existing results including those mentioned above are not directly applicable to prove or guarantee error upper bounds. 

In this paper, we aim to reveal the relationship between the complexity of neural networks and the internal structure of the function to be approximated. A theoretical foundation is developed that supports the error analysis of neural networks for not only functions as input-output relations, but also functions generated based on the trajectories of differential equations and the feedback law of optimal control. By the structure of a function, we mean compositional structure. In engineering applications, it is quite common that a complicated function with a high input dimension is the result of a sequence of compositions of simple functions that have low input dimensions. These simple functions as well as the network structure determine the neural network's complexity that is required to accurately approximate the compositional function. The main contributions of the paper include: 1) We develop an algebraic framework and an approximation theory for compositional functions. These results form the theoretical foundation for the proofs of several theorems on the error upper bound and neural network complexity in the approximation of compositional functions. The theoretical foundation is developed in a way so that it supports the error analysis for numerical algorithms. This capability is critical because it enables the analysis of approximation errors for problems for which analytic solutions are not available, such as differential equations and optimal control. 2) We identify a set of key features of compositional functions. These key features are quantitative. Their value determines an upper bound of the complexity of neural networks when approximating compositional functions. These key features serve as sufficient conditions for overcoming the curse of dimensionality. 3) We prove several formulae of error upper bounds for neural networks that approximate the solutions to differential equations, optimization, and optimal control.  

\section{Composition vs. projection: what is the difference?}
A widely used idea of representing functions is to project a function to a sequence of finite dimensional subspaces. For instance, a large family of functions can be represented by Fourier series  
\EQ
f(x) = \Fr{a_0}{2}+\ds\sum_{n=1}^\infty \left(a_n\cos (2n\pi x) + b_n\sin(2n\pi x) \right), &x\in \Real.
\EE
In this representation, the $n$-th terms, $a_n\cos (2n\pi x) + b_n\sin(2n\pi x)$, are considered to be the elementary units of the function. Because the projection operator is linear, this representation is closed under linear combination in the following sense: If both $f$ and $g$ are Fourier series, then so is $h=af+bg$ for any constants $a$ and $b$; and the $n$-th term of $h$ is determined by the $n$-th term of $f$ and $g$, i.e., the elementary units of $f$ and $g$ are the elementary units of $h$ as well. However, the Fourier series representation is not closed under some other important algebraic operations such as function composition. It is obvious that $h=f\circ g$, or $h(x)=f(g(x))$, is not in the form of Fourier series. Although one can find a Fourier series representation for $h$, its $n$-th term is not determined by the $n$-terms of $f$ and $g$, i.e., the $n$-th terms of $f$ and $g$  are not elementary units in the Fourier series of $h$. 

\begin{figure}[!ht]
\centering
\includegraphics[width = 2.25in]{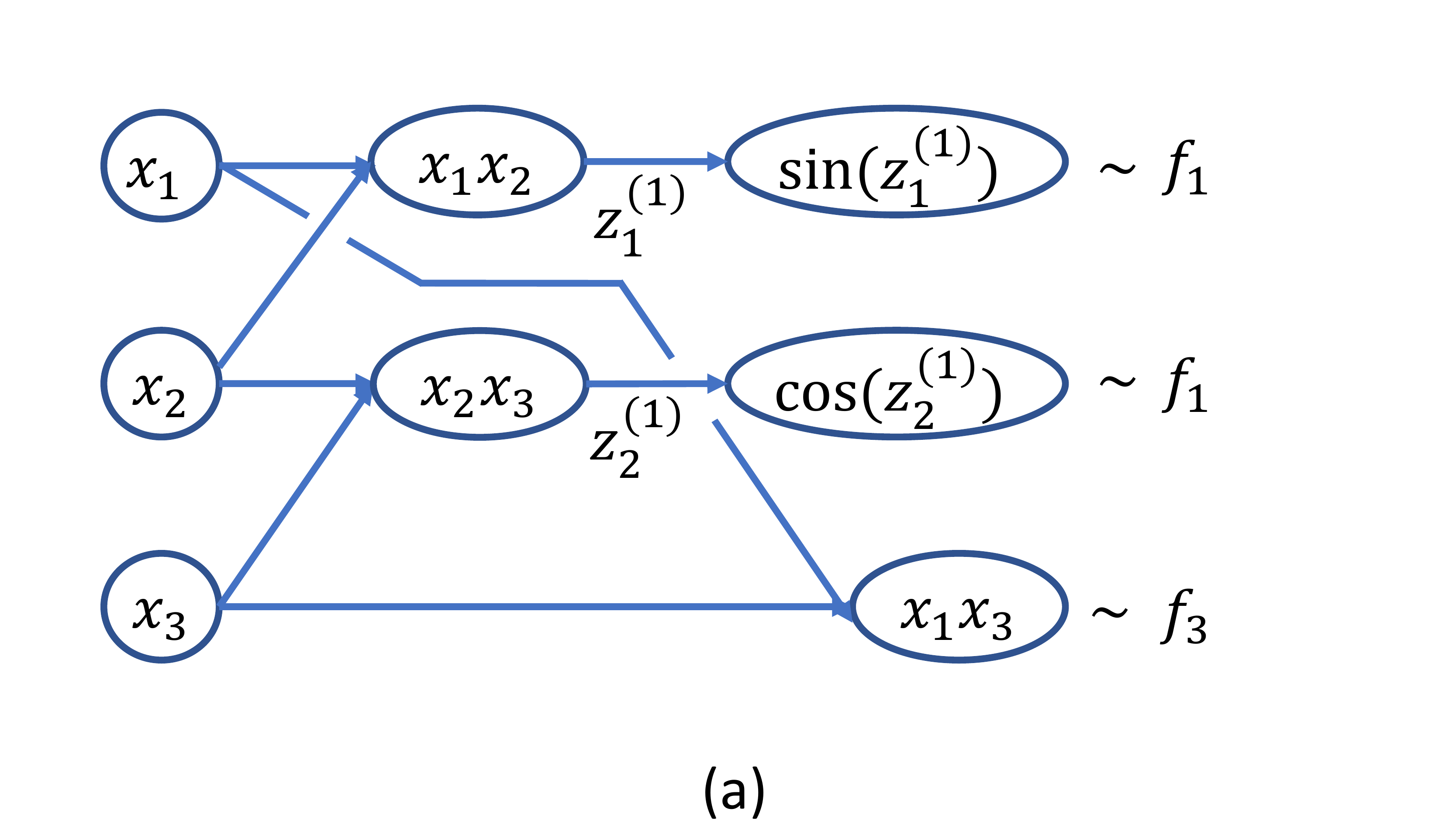} \includegraphics[width = 2.25in]{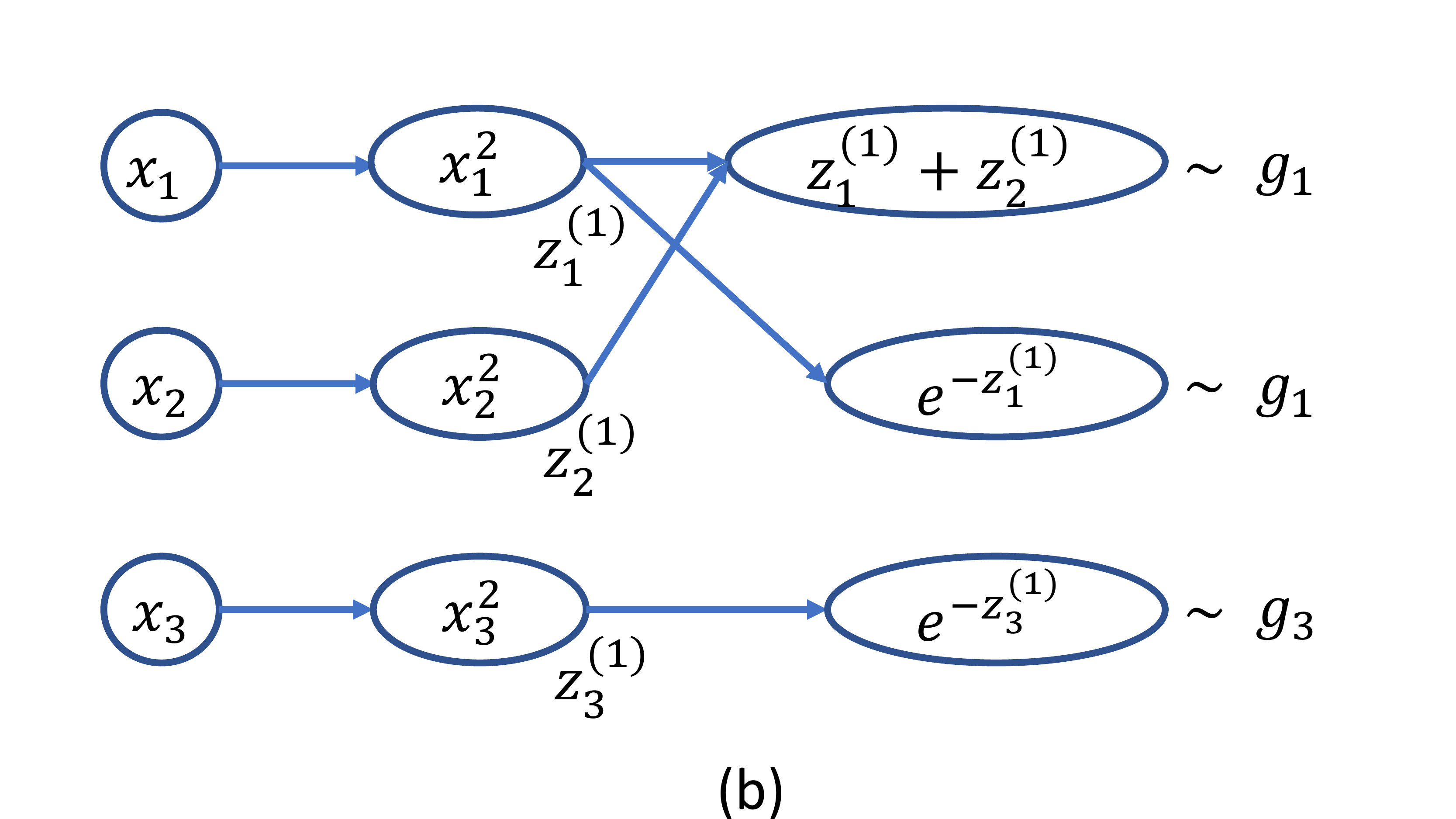}
\caption{Directed graphs representing the functions in (\ref{eq_compofunexmaple12}).}
\label{fig_closeoperation}
\end{figure}

\begin{figure}[!ht]
\centering
\includegraphics[width = 4.25in]{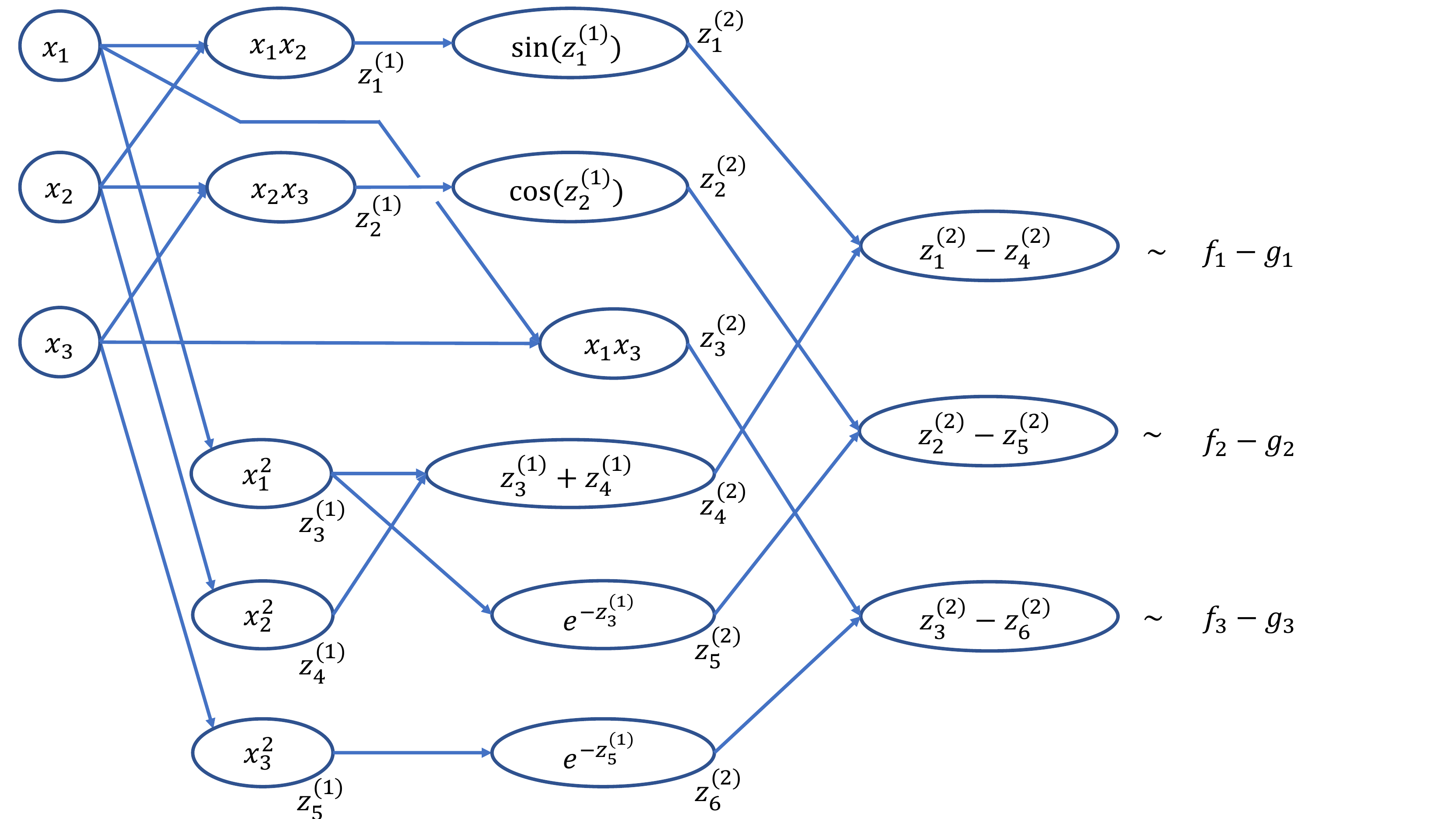} 
\caption{Directed graph representing $\bff-\bfg$, where $\bff$ and $\bfg$ are defined in (\ref{eq_compofunexmaple12}).}
\label{fig_closeoperation_addition}
\end{figure}

\begin{figure}[h!]
\centering
\includegraphics[width = 4.5in,angle=0]{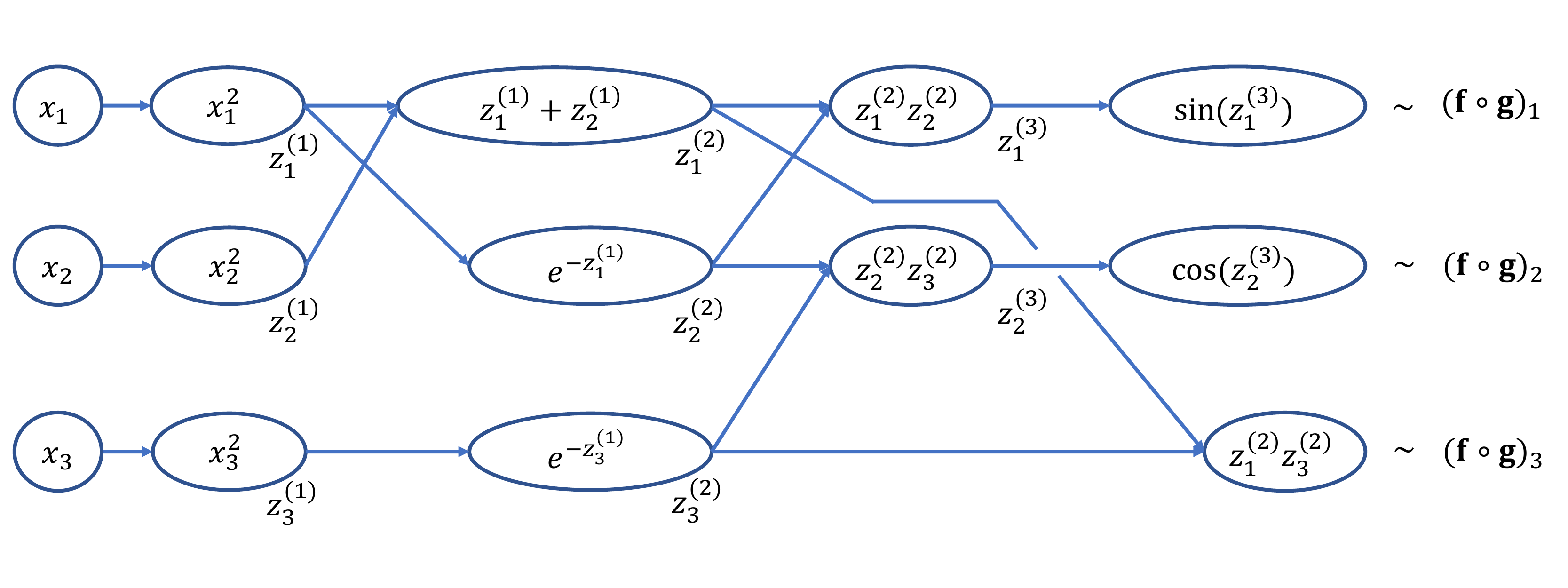} 
\caption{Directed graph representing $\bff\circ\bfg$, where $\bff$ and $\bfg$ are defined in (\ref{eq_compofunexmaple12}).}
\label{fig_closeoperation_compo}
\end{figure}

Let's consider functions generated through a network of multiple function compositions. A formal definition of compositional functions is given in the next section. Algebraic expressions of compositional functions can be cumbersome due to the complexity of the function. Instead, we use directed graphs. As an example, consider the following two vector valued functions
\EQ
\label{eq_compofunexmaple12}
\bff(x_1,x_2,x_3)=\left[\begin{array}{c} \sin (x_1x_2)\\ \cos (x_2x_3)\\x_1x_3 \end{array}\right], & \bfg(x_1,x_2,x_3)=\left[\begin{array}{c} (x_1^2+x_2^2)\\ e^{-x^2_1} \\e^{-x^2_3}\end{array}\right].
\EE
In this paper, we use bold letters, such as $\bfx$ and $\bff$, to represent a vector or a vector valued function and regular font to represent a scalar or a scalar valued function. The functions $\bff$ and $\bfg$ can be represented by a directed graph in which each node is a function with a relatively low input dimension (Figure \ref{fig_closeoperation}). These nodes are the elementary units of the function. 
Different from representations based on projection, the family of compositional functions is closed under both linear combinations and compositions. For example, $\bfh=\bff-\bfg$ is a compositional function; and every elementary unit (or node) of $\bff$ and $\bfg$ is an elementary unit of $\bfh$. The graph of $\bfh$ and its nodes are shown in Figure \ref{fig_closeoperation_addition}.  
If $\bfh=\bff\circ \bfg$, then $\bfh$ is a compositional function; the nodes of $\bff$ and $\bfg$ are also nodes of $\bfh$ (Figure \ref{fig_closeoperation_compo}). In addition to linear combinations and compositions, we will prove in the next section that the family of compositional functions is also closed under some other algebraic operations, including inner product, division, and substitution. 

Why is this interesting? In real-world applications, mathematical models of complicated problems oftentimes have a compositional structure in which the model can be represented by the composition of multiple functions that are relatively simple. The compositional structure reflects the interactive relations among basic components and sub-systems determined by physical laws or engineering design and thus contains valuable insights of the problem.  In addition to functions as input-output relations, many computational algorithms are compositional. For instance, an iterative algorithm is a process in which a function is repeatedly applied to the result from the previous step until the process converges to a small neighborhood of the target point. The iterative process is equivalent to a finite sequence of function compositions. For discrete dynamical systems, their trajectories are all compositional functions. Obviously, neural networks are also compositional functions. In fact, the family of  neural networks is closed under the operation of composition and linear combination. From the wide spectrum of subjects covered by compositional functions, we believe that an algebraic framework for compositional functions can provide a unified theoretical foundation for the study of neural networks in the approximation of not only functions as input-output relations, but also other subjects that have effective iterative algorithms, such as differential equations and optimal control. Because the family of compositional functions are closed under linear combinations and compositions, it simplifies the problem of tracking error propagations inside complicated compositional functions. 

Mathematical study on compositional functions has a rich history. In Hilbert’s 13th problem it was conjectured that the roots of some 7th order polynomials cannot be represented as compositions of functions of two variables. Hilbert’s conjecture was disproved by Arnold and Kolmogorov.  In \cite{Kolmogorov57} Kolmogorov showed that any continuous function defined on a d-dimensional cube can be exactly represented by a composition of a set of continuous one-dimensional functions as
$$
f(x_1,\cdots, x_d) = \sum_{q=1}^{2d+1}\phi_q\left(\sum_{p=1}^d\psi_{pq}(x_p)\right),
$$
where $\phi_q$ and $\psi_{pq}$ are continuous univariate functions. This astonishing result of Kolmogorov's representation theorem reveals that compositional structure could be viewed as an inherent property of any high-dimensional continuous functions. By relaxing Kolmogorov's exact representation to approximation and leveraging on techniques developed by Kolmogorov, Ref. \cite{kurkova92} gives a direct proof of a universal approximation theorem of multilayer neural networks for any continuous functions.

In deep learning, compositional functions have been addressed by many authors \cite{barron1,weinan, weinan0,mhaskar1,mhaskar2,poggio}. In \cite{barron1}, which is a study of statistic risk with a given sample size, the approach takes the advantage of the fact that deep networks with ramp activation functions can be represented as a tree rooted at the output node. In \cite{weinan,weinan0}, tree-like function spaces are introduced. They are considered as the natural function spaces for the purposes of approximation theory for neural networks. Compositional functions in \cite{mhaskar1,mhaskar2,poggio} are defined using directed acyclic graphs. Our definition of compositional functions is also based on directed acyclic graphs. However, we give an emphasis on the graph's layering, in addition to its nodes and edges. 

Before we introduce the main theorems, it is important to develop a theoretical foundation for error estimation. In Section \ref{sec_comp_def}, some basic concepts about compositional functions are defined. In Sections \ref{sec_3_2} and \ref{sec_3_3}, some algebraic properties of compositional functions are proved. These concepts and properties form a set of pillars that supports the approximation theory for compositional functions. In Section \ref{sec_4}, neural networks are introduced as a family of special compositional functions. The family of neural networks is closed under several algebraic operations including linear combination, composition, and substitution. This property enables the capability for one to compose complicated deep neural networks based on simple networks and algebraic operations. Also in Section \ref{sec_4}, a set of key features of compositional functions is identified. For neural network approximations, we prove an error upper bound that is determined by the key features.  Because iterative computational algorithms can be considered as compositional functions, the algebraic framework and the approximation theory are applicable to the solutions to differential equations, optimization, and optimal control. The results are proved in Sections \ref{sec_5} and \ref{sec_6}. 

\section{Algebraic properties of compositional functions}
\label{sec_3}
Compositional functions are functions consisting of multiple layers of relatively simple functions defined in low dimensional spaces. These simple functions are integrated together through algebraic operations such as function composition and linear combination. If compositional functions are estimated by neural networks, errors are propagated through layers of functions in a complicated way. Analyzing the estimation error requires a good understanding of the interplay between the error propagation and the internal structure of the compositional function. In this paper, some proofs of theorems on neural network solutions to differential equations and optimal control are impossible without first developing an algebraic foundation and approximation theory for the family of compositional functions. In the following, we first define a set of fundamental concepts and then prove some basic algebraic properties of compositional functions. 
In Table \ref{table1}, we summarize a list of repeatedly used notations in the following sections.  
\begin{table}[!ht]
\begin{center}
\begin{tabular}{|c|c||c|c|}
\hline
   \begin{minipage}[c][10mm][t]{0.1mm} \end{minipage} 
  \begin{tabular}{c}
   \hspace{-6mm} Notation  \hspace{-5mm}
  \end{tabular} &
  \begin{tabular}{c}
   Meaning
  \end{tabular} &
  \begin{tabular}{c}
  \hspace{-5mm} Notation  \hspace{-5mm}
  \end{tabular} &
  \begin{tabular}{c}
  Meaning  
  \end{tabular} 
  \\
  \hline
      \begin{minipage}[c][10mm][t]{0.1mm} \end{minipage} 
  \begin{tabular}{l}
 ${\cal G}^\bff$ 
  \end{tabular} &
  \begin{tabular}{l}
    a DAG associated with $\bff$
  \end{tabular} &
  \begin{tabular}{l}
    ${\cal L}^\bff(\cdot)$
  \end{tabular} &
   \begin{tabular}{l}
   a mapping: node $\rightarrow$ layer number
  \end{tabular} 
  \\
\hline
    \begin{minipage}[c][10mm][t]{0.1mm} \end{minipage} 
  \begin{tabular}{l}
 $\calV^\bff$ 
  \end{tabular} &
  \begin{tabular}{l}
     the set of nodes in $\calG^\bff$ 
  \end{tabular} &
  \begin{tabular}{l}
    $\calE^\bff$
  \end{tabular} &
   \begin{tabular}{l}
  the set of edges in $\calG^\bff$
  \end{tabular} 
  \\
\hline
    \begin{minipage}[c][10mm][t]{0.1mm} \end{minipage} 
  \begin{tabular}{l}
 $\calV^\bff_L$
  \end{tabular} &
  \begin{tabular}{l}
     the set of linear nodes in $\calG^\bff$
  \end{tabular} &
  \begin{tabular}{l}
    $\calV^\bff_G$ 
  \end{tabular} &
   \begin{tabular}{l}
  the set of general nodes in $\calG^\bff$  
  \end{tabular} 
  \\
\hline
    \begin{minipage}[c][10mm][t]{0.1mm} \end{minipage} 
  \begin{tabular}{l}
  $\calV^\bff_I$ 
  \end{tabular} &
  \begin{tabular}{l}
     the set of input nodes in $\calG^\bff$ 
  \end{tabular} &
  \begin{tabular}{l}
    $\calV^\bff_O$ 
  \end{tabular} &
   \begin{tabular}{l}
 the set of output nodes in $\calG^\bff$
  \end{tabular} 
  \\
\hline
    \begin{minipage}[c][10mm][t]{0.1mm} \end{minipage} 
  \begin{tabular}{l}
   $f_{i,j}$ 
  \end{tabular} &
  \begin{tabular}{l}
      the $j$-th node in the $i$-th layer  
  \end{tabular} &
  \begin{tabular}{l}
   $l_{max}^\bff$  
  \end{tabular} &
   \begin{tabular}{l}
 the layer number of output nodes
  \end{tabular} 
  \\
\hline
    \begin{minipage}[c][12mm][t]{0.1mm} \end{minipage} 
  \begin{tabular}{c}
    $d_{i,j}$\\  or $d_{i,j}^\bff$
  \end{tabular} &
  \begin{tabular}{l}
      the input dimension of $f_{i,j}$ 
  \end{tabular} &
  \begin{tabular}{c}
   $m_{i,j}$ \\ or $m^\bff_{i,j}$
  \end{tabular} &
   \begin{tabular}{c}
the smoothness of $f_{i,j}$
  \end{tabular} 
  \\
\hline
    \begin{minipage}[c][12mm][t]{0.1mm} \end{minipage} 
  \begin{tabular}{c}
   $R_{i,j}$ \\  or  $R^\bff_{i,j}$
  \end{tabular} &
  \begin{tabular}{c}
     $[-R_{i,j},R_{i,j}]^{d_{i,j}}$ is the domain \\ of $f_{i,j}$
  \end{tabular} &
  \begin{tabular}{c}
   $L_{i,j}$ \\ or $L^\bff_{i,j}$
  \end{tabular} &
   \begin{tabular}{c}
a Lipschitz constant associated \\ with $f_{i,j}$\
  \end{tabular} 
  \\
\hline
    \begin{minipage}[c][13mm][t]{0.1mm} \end{minipage} 
  \begin{tabular}{c}
  $\Lambda^\bff$
  \end{tabular} &
  \begin{tabular}{c}
     For all $f_{i,j}\in \calV^\bff_G$, the  largest \\ {\small $\max\{(R_{i,j})^{m_{i,j}},1\} \norm{f_{i,j}}_{W^{\infty}_{m_{i,j},d_{i,j}}}$} 
  \end{tabular} &
  \begin{tabular}{c}
   $L_{max}^\bff$
  \end{tabular} &
   \begin{tabular}{c}
the largest $L_{i,j}$ for $f_{i,j}\in \calV^\bff_G$
  \end{tabular} 
  \\
\hline
    \begin{minipage}[c][13mm][t]{0.1mm} \end{minipage} 
  \begin{tabular}{c}
$\bff$, $\bfg$, $\bfx$ \\  $f_{i,j}$, $x_i$
  \end{tabular} &
  \multicolumn{3}{|c|}
  { \begin{tabular}{c}
     bold letters represent a vector or a vector valued function  \\ 
     regular font represents a scalar or a scalar valued function 
  \end{tabular} }
\\
\hline
\end{tabular}
 \caption{Some frequently used notations.}
 \label{table1}
 \end{center}
\end{table}

\subsection{Definitions and notations}
\label{sec_comp_def}
\begin{figure}[h!]
\centering
\includegraphics[width = 2.25in]{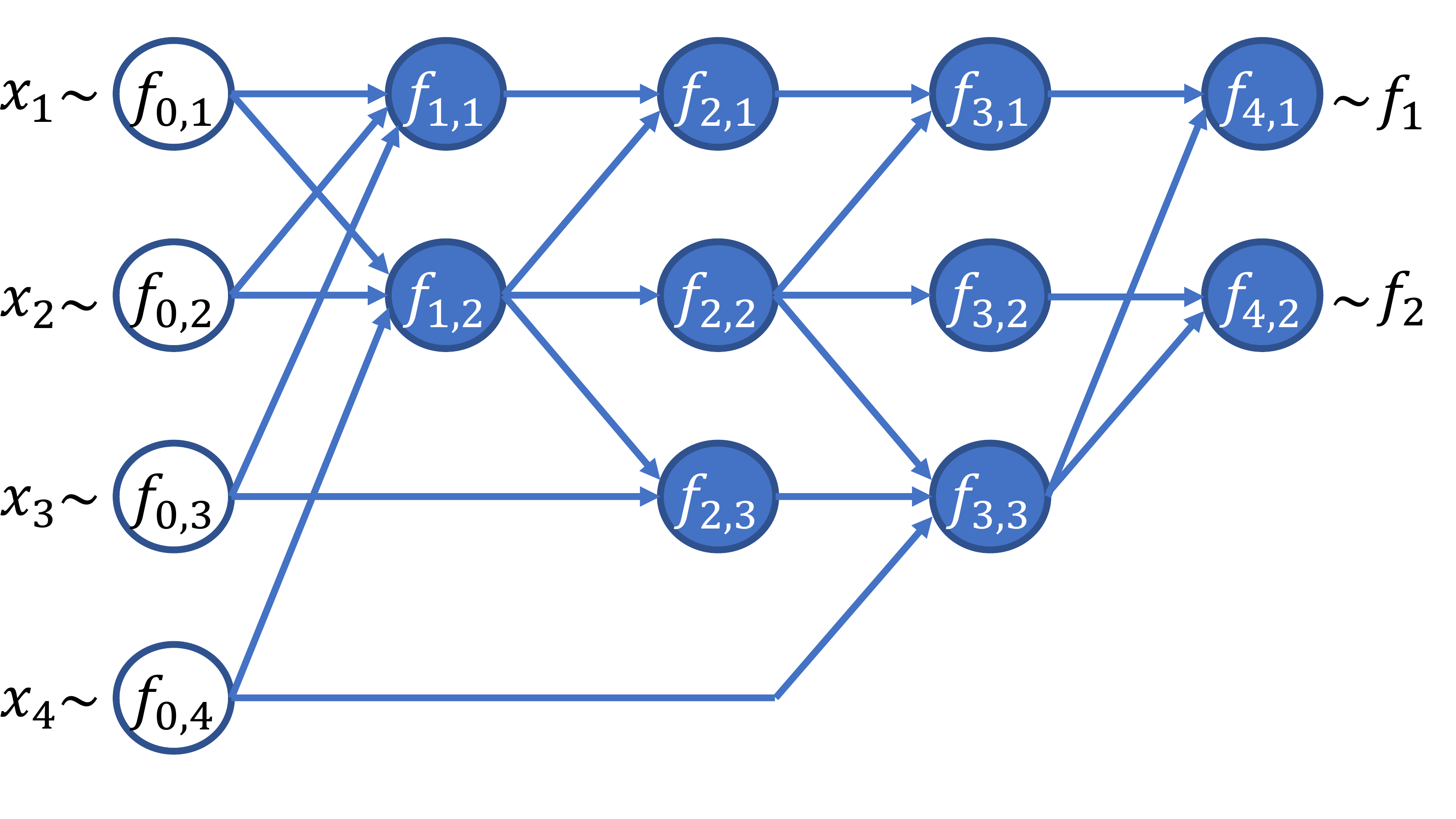}
\caption{An example of compositional function. The DAG has four input nodes representing, $x_1,x_2,x_3,x_4$,  and two output nodes, $f_{4,1},f_{4,2}$. }
\label{fig_DAG1}
\end{figure}

\begin{figure}[h!]
\centering
\includegraphics[width = 4.5in]{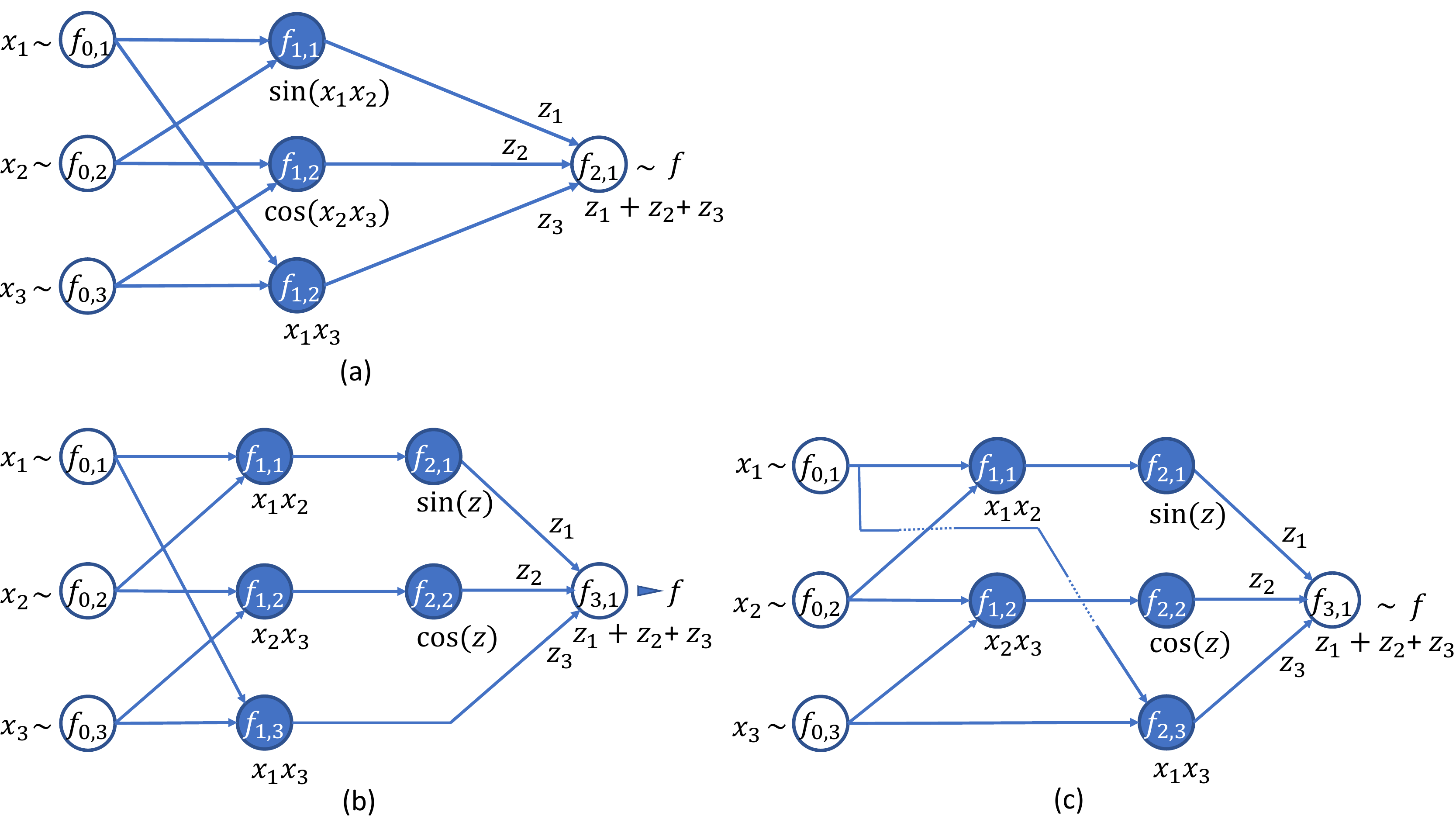}
\caption{Three DAGs associated with the same function in (\ref{eq_funexample3}). }
\label{fig_DAG_example1}
\end{figure}

A directed acyclic graph (DAG) is a directed graph without directed cycles. A DAG consists of a set of nodes and a set of directed edges. If each node is a function with input (inward edges) and output (outward edges), then the DAG represents a compositional function. Figure \ref{fig_DAG1} is an example of DAG associated with a function $\bff: \Real^4\rightarrow \Real^2$. The nodes that have no inward edges are call {\it input nodes}; those that have no outward edges are called {\it output nodes}.
DAGs allow layerings that agree with the direction of edges. More specifically, suppose $(h_i, h_j)$ is a directed edge in a DAG in which $h_i$ is a node in the $i$-th layer and $h_j$ is in the $j$-th layer, then $j>i$. For the compositional function in Figure \ref{fig_DAG1}, it has a total of five layers. In this paper, the set of input nodes is always the first layer, labeled as layer $0$; two layers next to each other are labeled using adjacent integers; the set of output nodes forms the terminal layer which has the highest layer number. We would like to point out that the layering of a DAG is not unique. Moreover, a function may have more than one associated DAGs. For example, consider the following function
\EQ
\label{eq_funexample3}
f(x_1,x_2,x_3)=\sin(x_1x_2)+\cos(x_2x_3)+x_1x_3.
\EE
In Figure \ref{fig_DAG_example1}, three DAGs are shown. As a function, they all represent $f$. Figure \ref{fig_DAG_example1}(a) has three layers. Both (b) and (c) have four layers. However, the set of nodes in two of the layers are different. In Figure \ref{fig_DAG_example1}, some nodes are colored and some are white. The reason will be explained following Definition \ref{def_comp}. 

\begin{definition} (Compositional function)
\label{def_comp}
A compositional function is a  triplet, $(\bff, \calG^\bff, \calL^\bff)$, consisting of a function, $\bff: \Real^d\rightarrow \Real^q$, a layered DAG, $\calG^\bff$, and a mapping, $\calL^\bff$.  If $h$ is a node in $\calG^\bff$, then $\calL^\bff(h)$ equals an integer that is the layer number of $h$. We also use subscripts to represent the layer number, i.e., $f_{i,j}$ is the $j$-th node in the $i$-th layer. Every node in $\calG^\bff$ that has at least one inward edge is a function, $f_{i,j}: \Real^{d_{i,j}}\rightarrow \Real$, where $d_{i,j} $ is the number of inward edges of the node. A node that has no inward edge is called an input node. It can take any value in the domain of $\bff$. All input nodes in $\calG^\bff$ are labeled as layer $\mathit{0}$, called the input layer. A node that has no outward edge is called an output node. All output nodes are located in the final layer, called the output layer. All layers between input and output layers are called hidden layers. We always assume that the ranges and domains of all nodes are compatible for composition, i.e.  if $(f_{i,j}, f_{l,k})$ is an edge in $\calG^\bff$, then the range of $f_{i,j}$ is contained in the interior of the domain of $f_{l,k}$. 
\end{definition}

For each node in a compositional function, the number of inward edges, $d_{i,j}$, defines the number of free variables of the node. Each node takes its value in $\Real$, i.e., all nodes are scalar valued functions. The set of nodes and edges in $\calG^\bff$ are denoted by ${\cal V}^\bff$ and ${\cal E}^\bff$, respectively. Given a node, $f_{i,j}$, if it is a 1st order polynomial, then it is called a {\it linear node}. Otherwise, it is called a {\it general node}. All input nodes are treated as linear nodes. In the figure of $\calG^\bff$, linear nodes have white color. The set of linear nodes are denoted by ${\cal V}^\bff_L$ and the set of general nodes are denoted by ${\cal V}^\bff_G$.  The set of input nodes, i.e., the first layer of $\calG^\bff$, is denoted by ${\cal V}_I^\bff$. Similarly, the layer formed by output nodes is ${\cal V}_O^\bff$. Therefore, $\calL^\bff({\cal V}_O^\bff)$ is the largest layer number in the graph. It is denoted by $l_{max}^\bff$.  Given an edge $(f_{i_1,j_1}, f_{i_2,j_2})$. If $i_2\geq i_1+2$, then we say that this edge skips layers. \\

\begin{example} \rm
A complicated function defined in high dimensional spaces can be a compositional function consisting of simple nodes. In a model of power systems \cite{anderson,qi}, the electric air-gap torque, ${\bf P}_e$, is a vector valued function that can be approximated by
\EQ
\label{eq_Pei}
({\bf P}_{e})_i=E_i^2G_{ii}+\ds\sum_{j=1, j\neq i}^m E_iE_j(G_{ij}\cos(\delta_i-\delta_j)+B_{ij}\sin(\delta_i-\delta_j)), &1\leq i\leq m,
\EE
where $m$ is the number of generators, $\delta_i$, $i=1,2,\cdots,m$, are rotor angles of the generators. Other notations represent constant parameters. This function is used to determine the equilibrium point of the power system. It is also a part of the torques applied to a rotor in a generator. If the system have $m=20$ generators, then the model is a function ${\bf P}_e: \Real^{20}\rightarrow \Real^{20}$. This function can be represented as a compositional function with simple nodes. Shown in Figure \ref{fig_powersystem} is $({\bf P}_{e})_i$ in a DAG with four layers. Except for the layer of $\sin(z)$ and $\cos(z)$, all nodes are linear. It will be shown later that linear nodes in a compositional function do not increase the complexity of the neural network that approximates the function. For all general nodes, their domains have a low dimension, $d_{i,j}=1$. Theorems in this paper help to reveal that a low input dimension of individual nodes implies low complexity when the function is approximated by neural networks. The complexity is not directly dependent on the overall dimension, in this case $d=20$.  \\
\begin{figure}[h!]
\centering
\includegraphics[width = 2.25in]{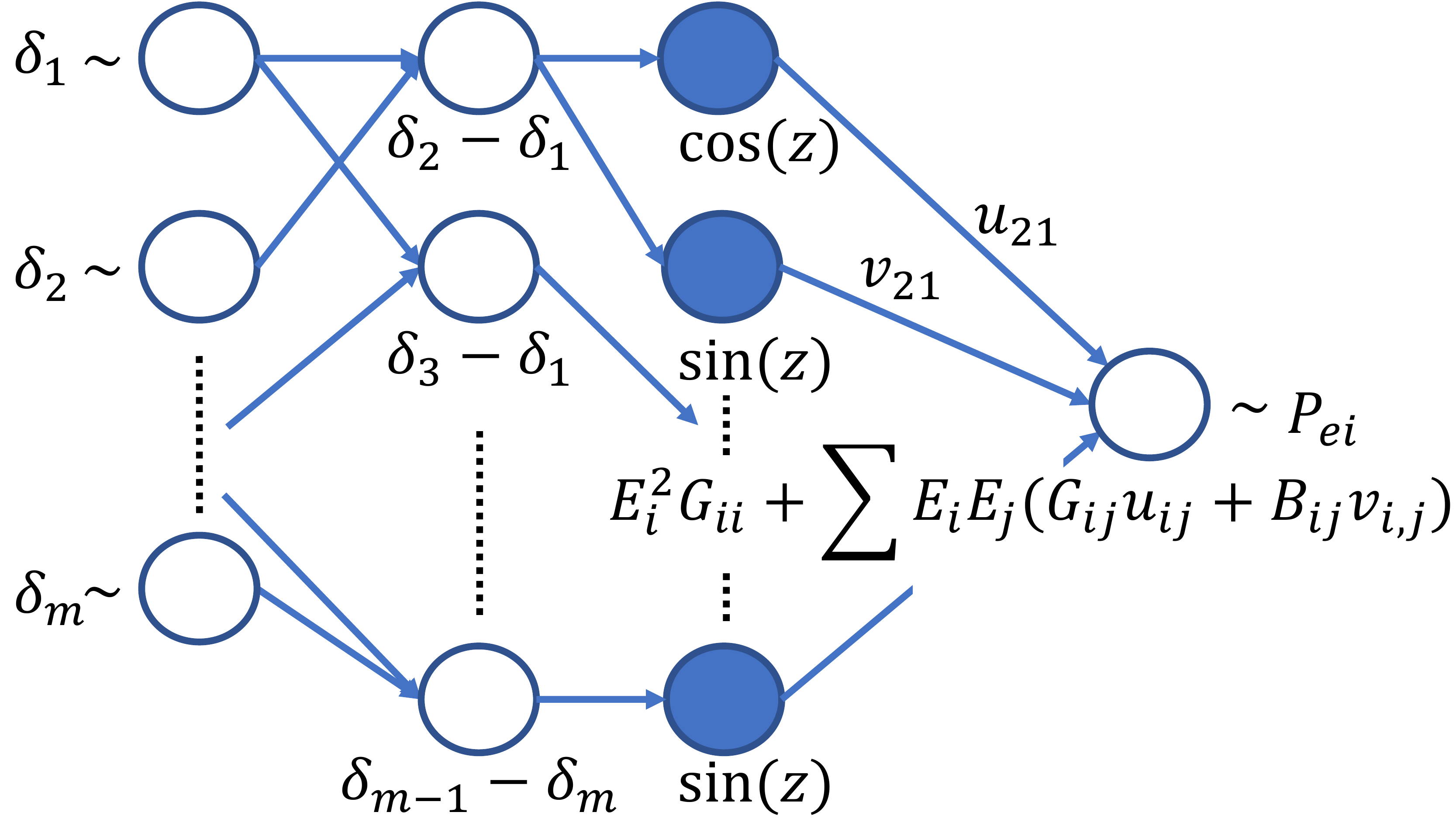}
\caption{DAG of $({\bf P}_{e})_i$ defined in (\ref{eq_Pei})}
\label{fig_powersystem}
\end{figure}
\end{example}

\begin{example} \rm
Many numerical algorithms are compositional functions. For instant, numerical algorithms of ODEs such as the Euler method and all explicit Runge-Kutta methods are compositional functions. Consider
\EQ
\dot \bfx=\bff(\bfx), & \bfx\in \Real^d, \, \bff(\bfx)\in \Real^d,\, t\in [0, T].
\EE
For any initial state $\bfx_0$, the following is a numerical solution based on the forward Euler method:
\EQ
\bfx_k=\bfx_{k-1}+h\bff(\bfx_{k-1}), & 1\leq k\leq K,
\EE
where $h=T/K$ is the step size. This algorithm is, in fact, a compositional function. The structure of its DAG is shwon in Figure \ref{fig_EulerExample}. Each box represents a layer that may have multiple nodes if $d>1$. In fact, the box of $\bff$ is a DAG if $\bff$ is also a compositional function. \\ 

\begin{figure}[!ht]
\centering
\includegraphics[width = 2.25in]{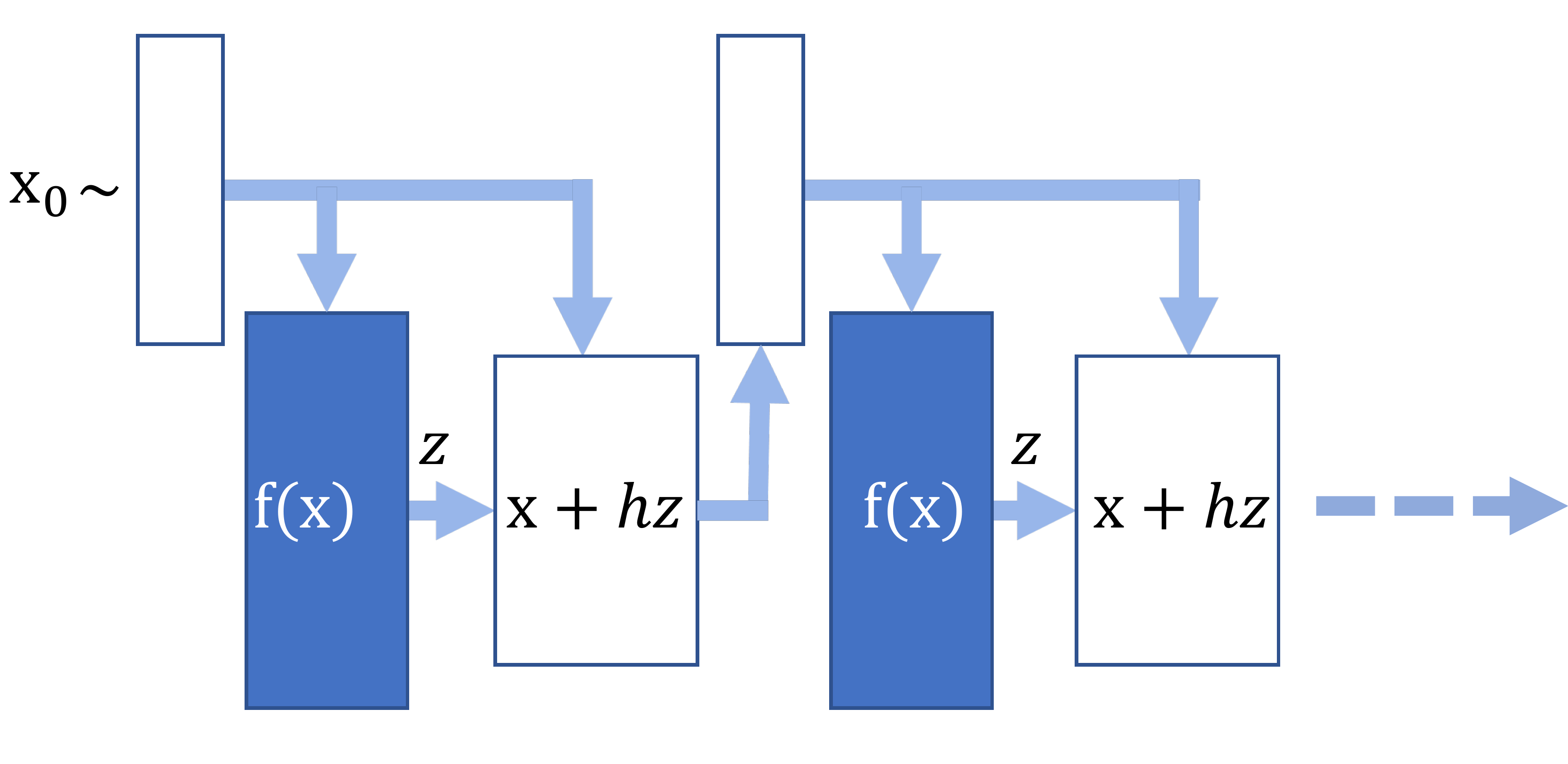}
\caption{DAG structure of the Euler method. If $\bff$ is a compositional function, then each box represents a layer that may have multiple nodes. The white box in the upper left represents the input layer of $\calG^\bff$. The box marked as ``$\bff (\bfx)$'' contains all hidden layers and the output layer of $\calG^\bff$.}
\label{fig_EulerExample}
\end{figure}
\end{example}

\begin{example} \rm
Neural networks are compositional functions. Let $\sigma: \Real \rightarrow \Real$ be an activation function. A function 
\EQ
f(\bfx)=\ds\sum_{j=1}^n a_j\sigma \left(\bfw_j^T \bfx +b_j \right), & \bfx\in\Real^d,
\EE
is a shallow neural network with a single hidden layer (Figure \ref{fig_shallowNN_2}), where $\bfw_j$, $a_j$ and $b_j$ are parameters. Similarly, deep neural networks are also compositional functions. 

\begin{figure}[h!]
\centering
\includegraphics[width = 2.25in]{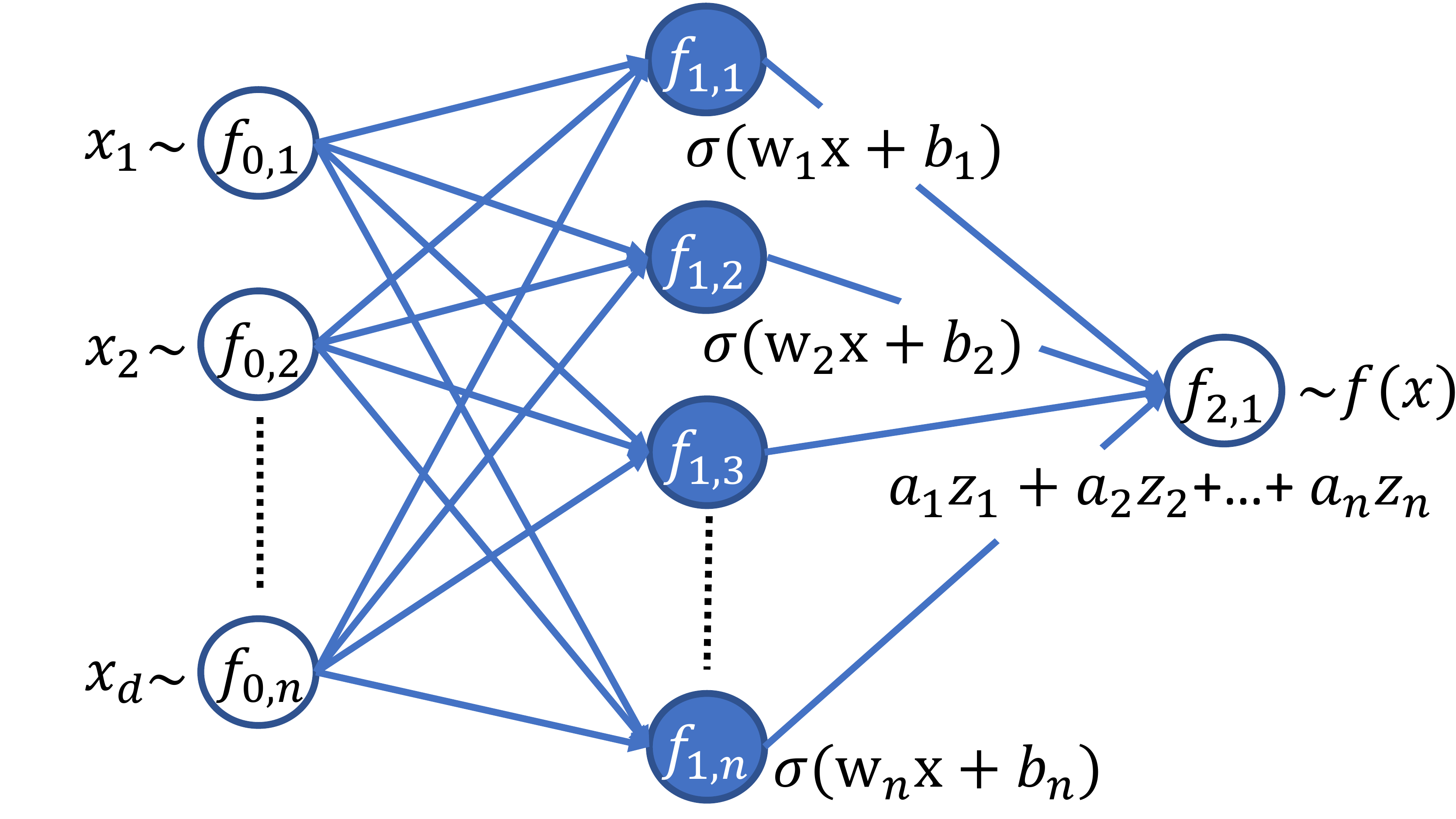}
\caption{DAG of a typical shallow neural network. In the hidden layer, $f_{1,j}(\bfx)=\sigma \left(\bfw_j^T \bfx +b_j \right)$. }
\label{fig_shallowNN_2}
\end{figure}
\end{example}

In this paper, the algebraic foundation and approximation theory on compositional functions are developed as a unified framework for various applications of neural networks. Before we introduce these applications, we must define some basic concepts about algebraic operations between compositional functions, such as linear combination, multiplication, division, substitution and composition. Many complicated functions and numerical algorithms, such as the solutions of differential equations or optimal control, are the results of a sequence of these algebraic operations of compositional functions.

\begin{figure}[!ht]
\centering
\includegraphics[width = 4.5in]{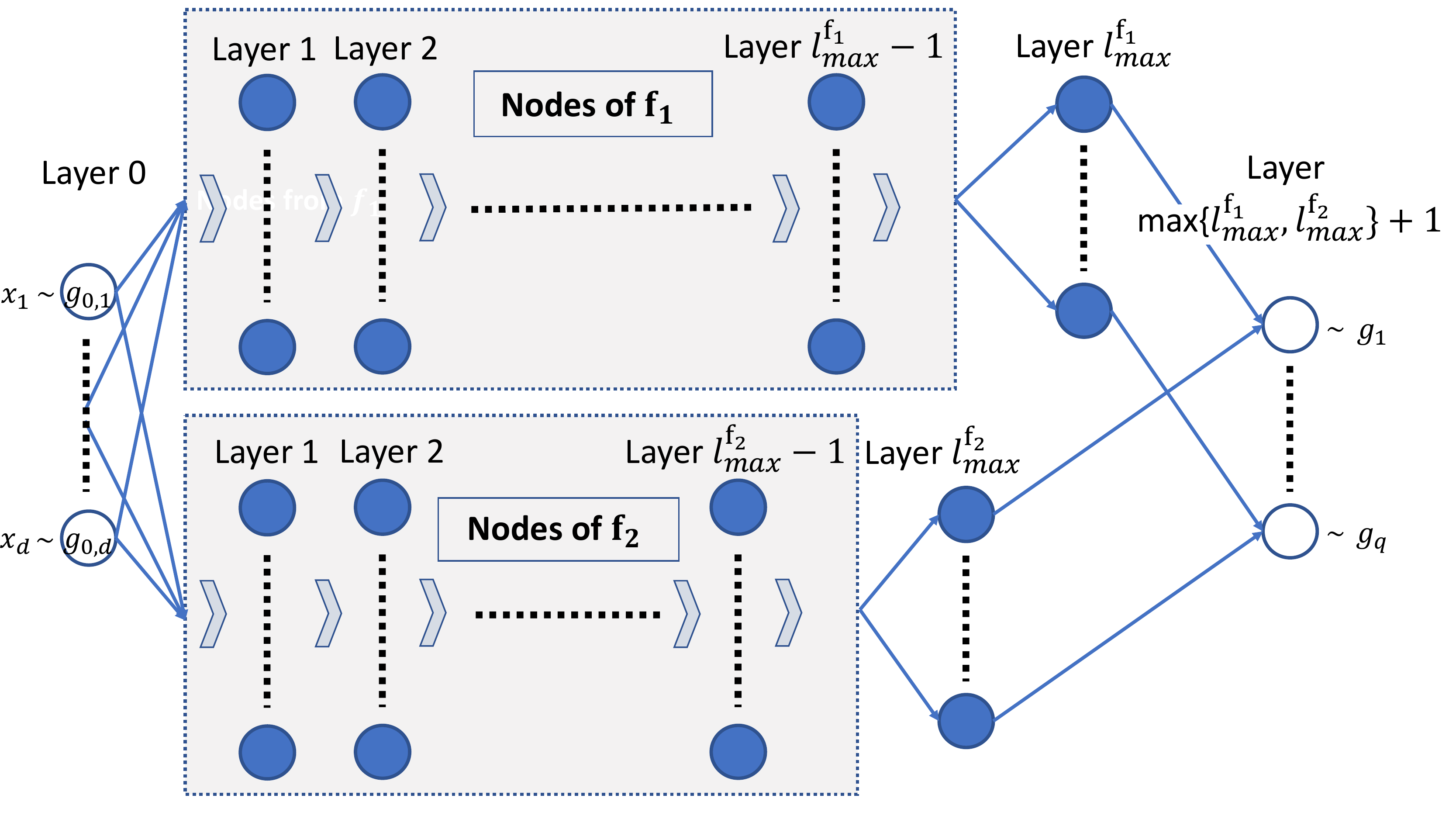}
\caption{Induced DAG of $a\bff_1+b\bff_2$. In the output layer, the $k$-th node is $af^{\bff_1}_{l^{\bff_1}_{max},k}+bf^{\bff_2}_{l^{\bff_2}_{max},k}$, where $f^{\bff_i}_{l^{\bff_i}_{max},k}$ is the $k$-th output node of $\bff_i$, $i=1,2$. For $\bff_1^T\bff_2$, the DAG is the same except that the output layer has one node only, which is $ \sum_{k=1}^d f^{\bff_1}_{l^{\bff_1}_{max},k}f^{\bff_2}_{l^{\bff_2}_{max},k}$. For $q=1$, the DAG of $f_1/f_2$ has an output layer that has one node, $f^{f_1}_{l^{f_1}_{max},1}/f^{f_2}_{l^{f_2}_{max},1}$. }
\label{fig_DAG_sum}
\end{figure}

\begin{definition}
\label{def_addition}
(Linear combintation, multiplication and division of compositional functions) 
Given compositional functions $(\bff_1,\calG^{\bff_1},\calL^{\bff_1})$ and $(\bff_2,\calG^{\bff_2},\calL^{\bff_2})$ in which both $\bff_1$ and $\bff_2$ are from $\Real^d$ into $\Real^q$. Let $a, b \in\Real$ be constant numbers. Define
\EQ
\bfg=a\bff_1+b\bff_2.
\EE
Then $\bfg$ is a compositional function associated with a layered DAG that is induced by $(\bff_1,\calG^{\bff_1},\calL^{\bff_1})$ and $(\bff_2,\calG^{\bff_2}, \calL^{\bff_2})$. The induced DAG and its layering are defined as follows (see Figure \ref{fig_DAG_sum} for an illustration).
\begin{enumerate}
\item Input layer: Let $\calV^{\bff_1}_I=\calV^{\bff_2}_I=\calV^\bfg_I$, i.e., the input layer of $\calG^\bfg$ is formed by overlapping the input nodes of $\bff_1$ and $\bff_2$. The outward edges of the input nodes in $\calG^{\bff_1}$ and $\calG^{\bff_2}$ are combined.  
\item Hidden layers:  all nodes in $\left( \calV^{\bff_1}\setminus\calV^{\bff_1}_I\right)\cup \left( \calV^{\bff_2}\setminus\calV^{\bff_2}_I\right)$ are nodes of $\calG^\bfg$. The layer number of each node is kept the same as that in their original DAG. The edges pointing to or from these nodes are kept unchanged.
\item Output layer: Create $q$ output nodes. Each one has two inward edges. For the $k$-th node, where $k=1,2,\cdots, q$, one inward edge starts from $f_{l^{\bff_1}_{max},k}$ and the other one from $f_{l^{\bff_2}_{max},k}$. Each output node is a function 
\EQ
\label{eq_alg1}
h=az_1+bz_2.
\EE
The layer number of output nodes in $\calG^\bfh$ is 
\EQ
\max\left\{ l_{max}^{\bff_1}, l_{max}^{\bff_2}\right\}+1.
\EE
\end{enumerate}
One can define the induced DAGs for the compositional functions $\bff_1^T\bff_2$ and $f_1/f_2$ (if $q=1$, $f_1$ and $f_2$ are scalar valued functions and $f_2\neq 0$) by modifying the output layer accordingly. 
\end{definition}

\begin{figure}[!ht]
\centering
\includegraphics[width = 4.5in]{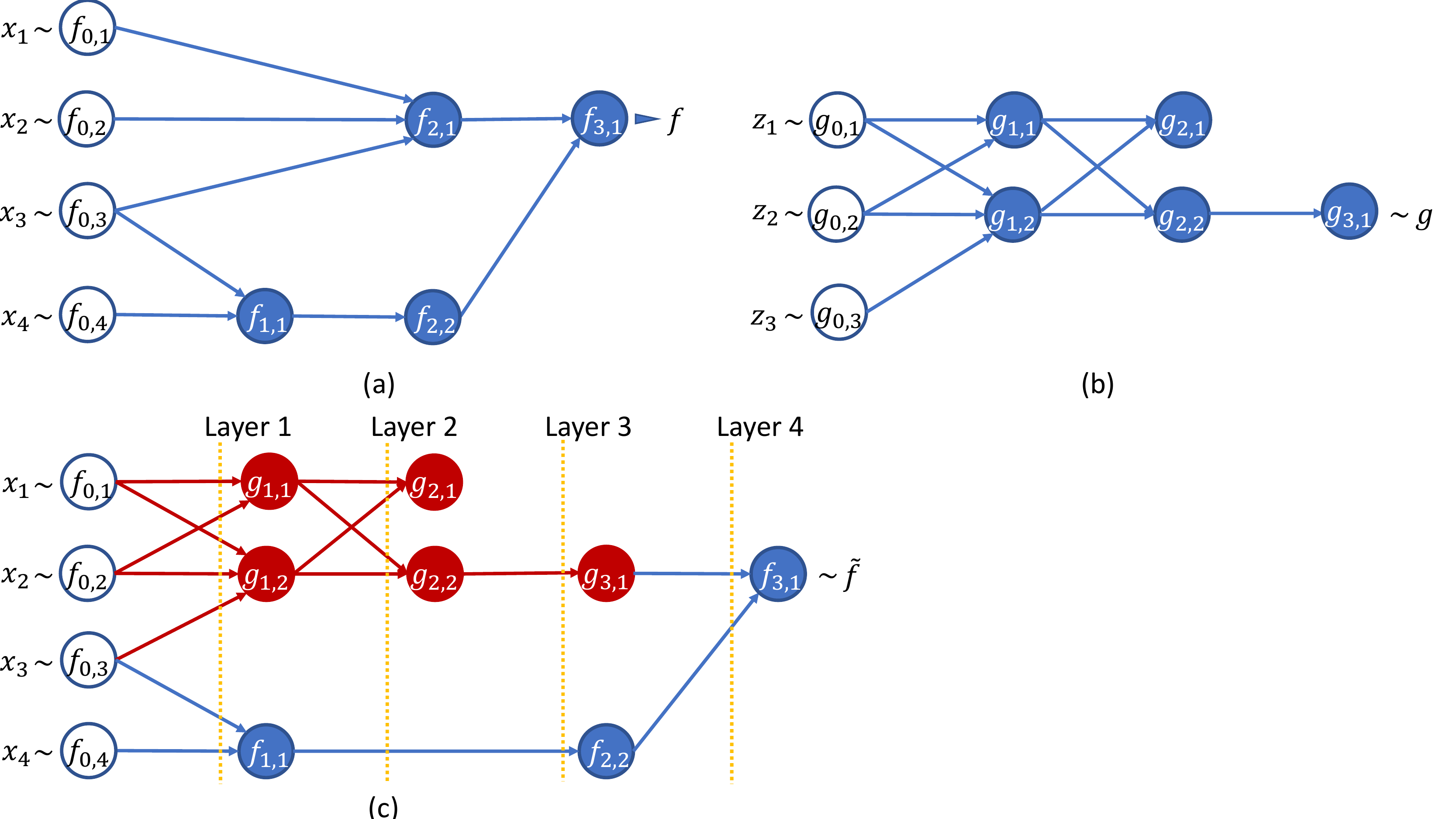}
\caption{Node substitution. The compositional function in (c) is the result of substituting $g$ in (b) for the node $f_{2,1}$ in (a).}
\label{fig_DAG_comp}
\end{figure}

This definition implies that the family of compositional functions is closed under inner product and division, in addition to linear combination and composition. Unless otherwise stated, in this paper we always use the induced DAG for functions that are generated through algebraic operations of compositional functions.  In the following, we define the induced DAG for a few more types of algebraic operations involving compositional functions. 
In Figure \ref{fig_DAG_comp}(a)-(b), two compositional functions, $f$ and $g$, are shown. We would like to substitute $g$ for the node $f_{2,1}$, a function that has the same number of inputs as $g$. Figure \ref{fig_DAG_comp}(c) shows a layered DAG of $\tilde f$, the function resulting from the substitution. The blue nodes and edges are those from $\calG^f$. The red nodes and edges are those from $\calG^g$. In the substitution, all input nodes of $\calG^g$ are removed; their outward edges are attached to the corresponding nodes in $\calG^f$; $f_{2,1}$ is replaced by the output node of $\calG^g$. The layering is changed depending on the depth of $\calG^g$.  

\begin{definition}(Node substitution)
\label{def_compsub}
Given a compositional function $(\bff, \calG^\bff, \calL^\bff)$. Let $f_{i,j}: \Real^{d_{i,j}}\rightarrow \Real$ be a node in the $i$-th layer, where $i\geq 1$. Let $(g, \calG^g, \calL^g)$, $g: \Real^{d_{i,j}}\rightarrow \Real$, be a compositional function. Substituting $g$ for $f_{i,j}$ results in a new compositional function, $\tilde \bff$. The induced DAG and its layering are defined as follows. 
\begin{enumerate}
\item $\calV^{\tilde \bff}=\left(\calV^\bff\setminus \{ f_{i,j}\}\right)\cup \left( \calV^g\setminus \calV^g_I\right)$ 
\item If $\{ v_1, \cdots, v_{d_{i,j}}\}$ is the ordered set of inward edges of $f_{i,j}$, then each $v_i$ is replaced by the set of outward edges of $g_{0,i}$. For the outward edges of $f_{i,j}$, their new starting node in $\calG^{\tilde \bff}$ is the output node of $g$.  All other edges in $\calG^\bff$ and $\calG^g$ are kept unchanged.
\item Let $\Dlt i$ be a positive integer so that  $i-\Dlt i$ be the highest layer number of layers in $\calG^\bff$ from which at least one node has an outward edge pointing to $f_{i,j}$. The layering in $\calG^{\tilde \bff}$ is defined as follows.
\EQ
\mbox{For } f_{l,k}\in \calV^\bff\setminus \{ f_{i,j}\}, & \calL^{\tilde \bff}(f_{l,k})=\left\{ \begin{array}{lll} l, & l \leq i-1 \\ l+\max\{0, l_{max}^g-\Dlt i\}, & l \geq  i \end{array}\right. ,\\
\mbox{For } g_{l,k}\in \calV^g\setminus \left( \calV^g_I\cup\calV^g_O\right),& \calL^{\tilde \bff}(g_{l,k})=l+i-\Dlt i,\\
\mbox{For } g_{l,k} \in \calV^g_O,  &  \calL^{\tilde \bff}(g_{l,k}) = \max\{i, i-\Dlt i+l^g_{max} \} .
\EE
\end{enumerate}
\end{definition}

Many numerical algorithms are iterative, i.e., the same function is repeatedly applied to the result from the previous iteration until the process converges to a small neighborhood of the target point. If each iteration is a compositional function, then the numerical process can be considered as a sequence of compositions of compositional functions. 

\begin{figure}[h!]
\centering
\includegraphics[width = 4.5in]{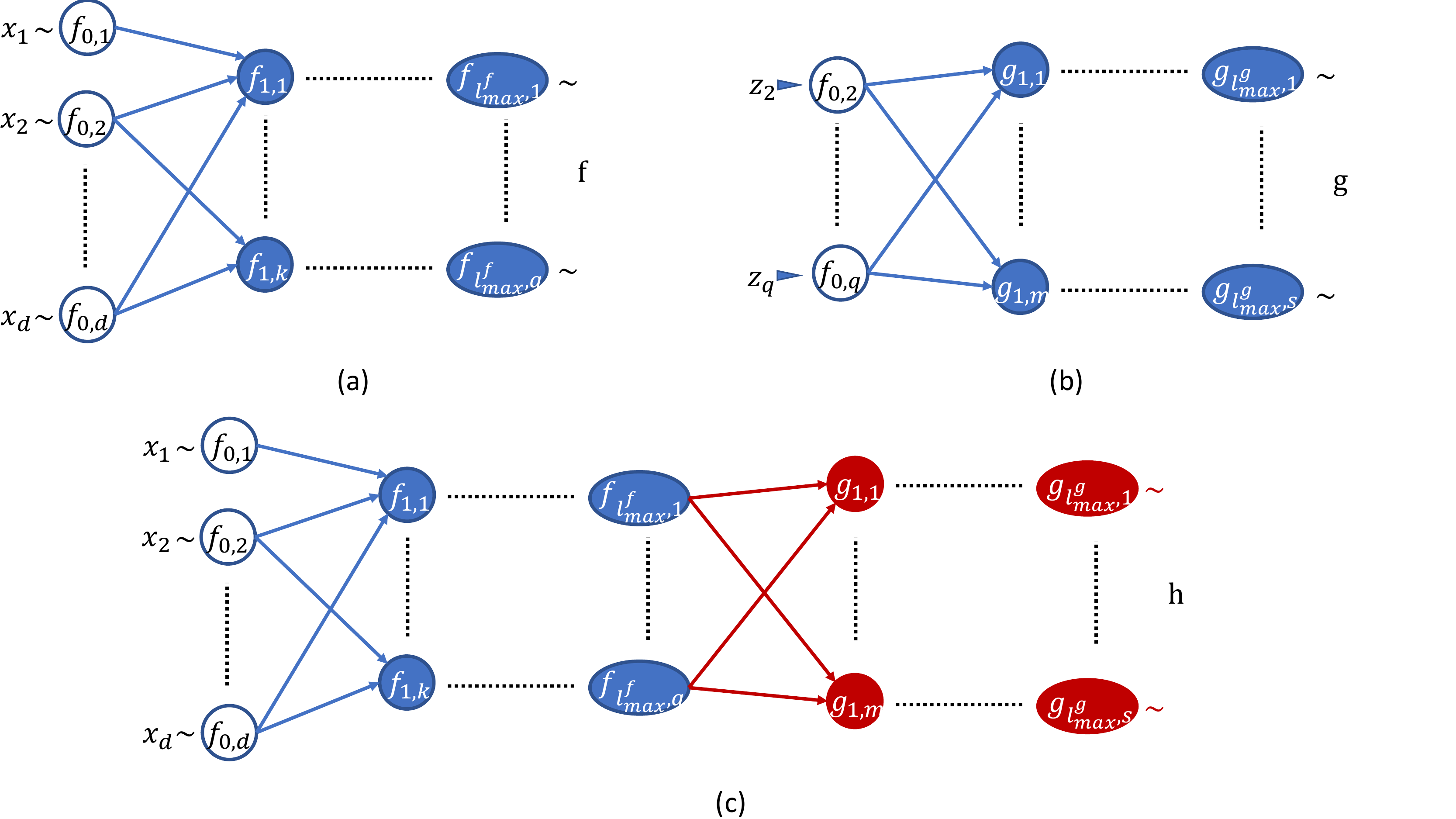}
\caption{Composition of compositional functions. The function in (c) is the composition of the functions in (a) and (b). }
\label{fig_DAG_comp2}
\end{figure}

\begin{definition} (Composition of compositional functions)
\label{def_compcomp}
Given positive integers $d$, $q$ and $s$. Consider two compositional functions $(\bff,\calG^\bff,\calL^\bff)$ and $(\bfg,\calG^\bfg,\calL^\bfg)$,  $\bff: \Real^d\rightarrow \Real^q$, $\bfg: \Real^q\rightarrow\Real^s$. Let $\bfh=\bfg\circ \bff: \bfx\in \Real^d \rightarrow g(f(\bfx))\in \Real^s$. Then $\bfh$ is a compositional function. Stacking $\calG^\bfg$ horizontally to the right of $\calG^\bff$ forms the induced DAG of $\bfh$ as illustrated in Figure \ref{fig_DAG_comp2}. More specifically, 
\begin{enumerate}
\item $\calV^\bfh=\calV^\bff\cup \left( \calV^\bfg\setminus \calV^\bfg_I\right)$. 
\item The outward edges from the input nodes in $\calG^\bfg$ are attached to the corresponding output nodes in $\calG^\bff$ as new starting nodes. 
\item The order of layers are kept unchanged, i.e.,
\EQ
\calL^\bfh(f_{i,j})=\calL^\bff(f_{i,j}), & \mbox{for } f_{i,j} \in \calV^\bff,\\
\calL^\bfh(g_{i,j})=l^\bff_{max}+\calL^\bfg(g_{i,j}), & \mbox{for } g_{i,j} \in \calV^\bfg\setminus \calV^\bfg_I.
\EE
\end{enumerate}
\end{definition}

\begin{figure}[h!]
\centering
\includegraphics[width = 2.25in]{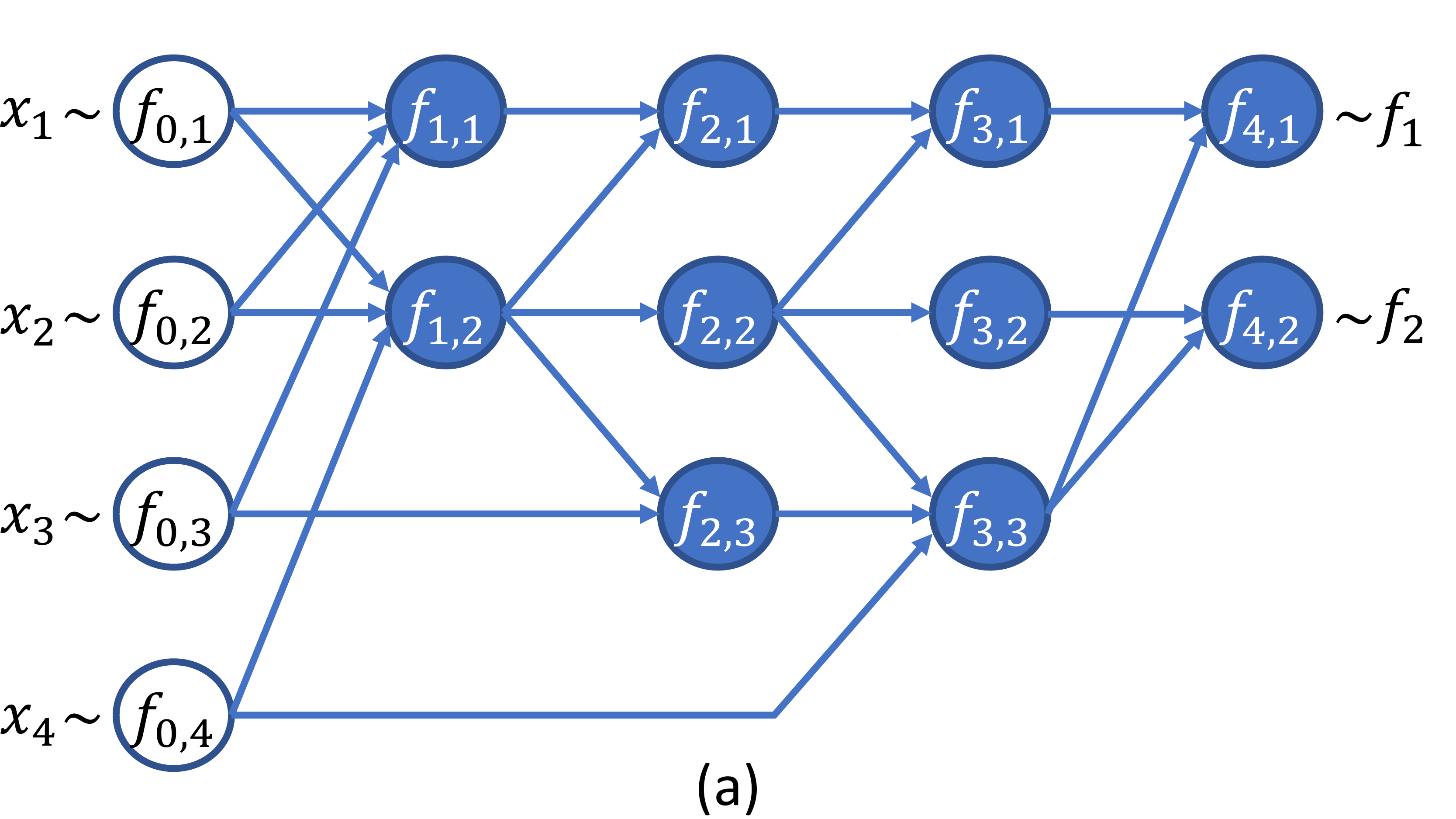} \includegraphics[width = 2.25in]{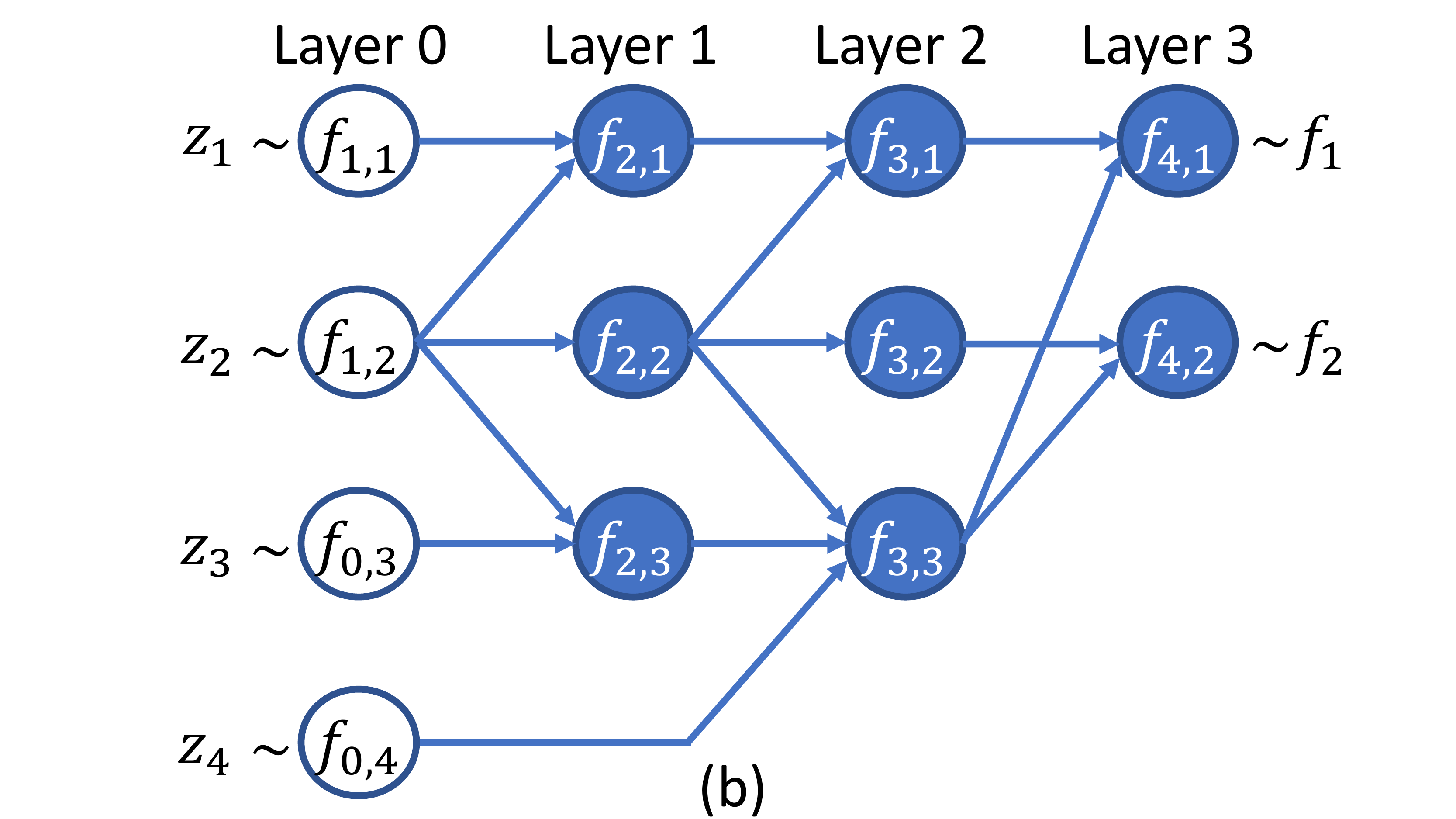}
\caption{Compositional truncation. The function in (b) is the compositional truncation of the function in (a) along layer $l=1$.}
\label{fig_DAG_truncation}
\end{figure}

In this study, we find that the layering of a DAG plays an important role in the propagation of approximation errors through a network of nodes. For example, given a node in the $i$-th layer of a compositional function. The variation of this node impacts the value of nodes in the layers $l\geq i+1$, but not those in the layers $l\leq i$. In the following, we introduce the concept of compositional truncation along a layer. The concept leads us to the definition of the Lipschitz constant associated with nodes. Consider a compositional function with five layers shown in Figure \ref{fig_DAG_truncation}(a). Its truncation along the layer $l=1$  is a function, $\bar \bff$, that is shown in Figure  \ref{fig_DAG_truncation}(b). In the truncation process, we first remove all edges that end at nodes in layers $l \leq 1$. Subsequently, $f_{0,1}$ and $f_{0,2}$ become isolated nodes, i.e., they have neither outward nor inward edge.  Remove them from the graph. Align all nodes in layers $l\leq 1$, $f_{1,1}$, $f_{1,2}$, $f_{0,3}$ and $f_{0,4}$, to form the input layer of the truncated function. In the new input layer, the nodes from layer $l=1$ in $\calG^\bff$ (in this example, $f_{1,1}$ and $f_{1,2}$) are kept in the same order; the nodes moved up from layer $l=0$ (in this case $f_{0,3}$ and $f_{0,4}$) are located at the lower half of the layer. Their order is not important. A set of dummy variables, $(z_1, z_2, z_3, z_4 )$, is introduced as the input of $\bar \bff$, the truncated function. It is important to clarify that the domain of $z_k$ is the intersection of the domains of all nodes that are directly connected with $z_k$ in $\bar \bff$. For example, $z_2$ in Figure \ref{fig_DAG_truncation} can take any value that is in the domains of $f_{2,1}$, $f_{2,2}$ and $f_{2,3}$ simultaneously. The value is not necessarily in the range of the original node $f_{1,2}$ in $\calG^\bff$. 

\begin{definition} (Compositional truncation)
Given a compositional function $(\bff,\calG^\bff,\calL^\bff)$. Let $i$ be an integer, $1\leq i\leq l_{max}^\bff-1$. The compositional truncation of $\bff$ along layer $i$ is another compositional function, $\bar \bff$. Its layered DAG is obtained by the following process: (i) removing all edges in $\calG^\bff$ whose ending nodes are located in layers $l\leq i$; (ii) removing all isolated nodes whose inward and outward edges are all removed in (i); (iii) forming the input layer of $\bar \bff$ by aligning all nodes in layers $l\leq  i$ in $\calG^\bff$ that are not removed in (ii). We find it convenient in discussions if the order of $f_{i,j}$'s in $\calG^\bff$ remains the same in the new input layer and they are located at the top portion of the layer. All other nodes in the new input layer are located at the bottom of the layer organized in any order one may choose. Given any $f_{l,k}\in \calG^\bff$, $l\geq i$, its layer number in $\calG^{\bar \bff}$ is $l-i$. The variables associate with the input layer of $\calG^{\bar \bff}$ are called dummy inputs. Each dummy input can take any value that is in the common domain of all nodes in $\calG^{\bar \bff}$ that have inward edges starting from this input node.
\end{definition}

The concept of the Lipschitz constant is essential in several theorems proved in this paper. In the following, we define the Lipschitz constant associated with nodes based on a given norm. For vectors in $ \Real^d$, its $p$-norm is 
\EQ
\norm{ \bfx}_p = \left\{ 
\begin{array}{ll}
 \left( \abs{x_1}^p+\abs{x_2}^p+\cdots +\abs{x_d}^p \right)^{1/p}, & 1\leq p < \infty   \\
 \max \{ |x_1|, |x_2|, \cdots, |x_d|\}, & p = \infty
\end{array}
\right.
\EE

\begin{definition} (The Lipschitz constant associated with a node)
\label{def_lipschitz}
Given a compositional function $(\bff,\calG^\bff,\calL^\bff)$ and a constant $1\leq p\leq \infty$. Consider a node, $f_{i,j}$, in the $i$-th layer, where $0\leq i \leq l^\bff_{max}-1$. Let $\bar \bff(z_1,\cdots, z_j,\cdots)$ be the truncation of $\bff$ along the $i$-th layer. If $L_{i,j}>0$ is a constant satisfying 
\EQ
\label{eq_lipschitz2}
\norm{\bar \bff(z_1,\cdots, z_j+{\it \Dlt} z, \cdots) - \bar \bff(z_1,\cdots, z_j, \cdots)}_p \leq L_{i,j} \abs {\Dlt z}
\EE
for all $(z_1,\cdots,z_j, \cdots )$ and $(z_1,\cdots,z_j+\Dlt z, \cdots) $ within the domain of $\bar \bff$, $L_{i,j}$ is called a Lipschitz constant (under the $p$-norm) associated with $f_{i,j}$. In some discussions, we denote the constant by $L^\bff_{i,j}$ to differentiate Lipschitz constants associated with different functions. 
\end{definition}
We should point out that a Lipschitz constant associated with a node, $f_{i,j}$, is different from the Lipschitz constant of $f_{i,j}$. The later is about the rate of change of $f_{i,j}(\bfz)$ when $\bfz$ is changed; the former is about the rate of change of $\bar \bff$, the compositional truncation, when the value of $f_{i,j}$ is changed, where the node itself is treated as a free variable.  The role of the Lipschitz constant of individual nodes in a DAG is studied in \cite{mhaskar1,mhaskar2},  where a good error propagation phenomenon is illustrated. Definition \ref{def_lipschitz} provides a different way of tracking error propagation. It leads to a feature (to be introduced in Section \ref{sec_4}) that has direct impact on an error upper bound of function approximations using neural networks.

\subsection{Some properties of Lipschitz constants}
\label{sec_3_2}
Algebraic operations may change Lipschitz constants associate with nodes. The propositions in this section are about the rules that Lipschitz constants follow in some algebraic operations of compositional functions. 

\begin{proposition}
\label{prop_lipschitz1}
Given a compositional function, $(\bff,\calG^\bff,\calL^\bff)$, and a node, $f_{i,j}$, in the $i$-th layer. Suppose $(g,\calG^g,\calL^g)$ is another compositional function in which $g$ is a scalar valued function that has the same domain as $f_{i,j}$. Let $(\tilde \bff, ,\calG^{\tilde \bff},\calL^{\tilde \bff})$ be the compositional function resulting from substituting $g$ for $f_{i,j}$, assuming that the range of $g$ is compatible for the substitution. For all nodes in $\calG^\bff$ in layers $l\geq i$, their associated Lipschitz constants can be carried over to $\calG^{\tilde \bff}$. Or equivalently, one can assign
\EQ
\label{eq_lipschitz1}
L^{\tilde \bff}_{l+\max\{0, l^g_{max}-\Dlt i\},k}=L^\bff_{l,k}, & l\geq i, 
\EE
where $i-\Dlt i$ is the highest layer number of layers in $\calG^\bff$ from which at least one edge points to $f_{i,j}$.
\end{proposition}

\noindent{\it Proof}. From Definition \ref{def_compsub}, substituting $g$ for $f_{i,j}$ does not change the nodes and edges above the $i$-th layer. The number of nodes in the $i$-th and higher layers in $\calG^\bff$ is the same as that in the layers $l\geq (i+\max\{0, l^g_{max}-\Dlt i\})$ in $\calG^{\tilde \bff}$. All edges in $\calG^\bff$ that end at nodes in the layers $l\geq (i+1)$ have a one-to-one correspondence with the edges in $\calG^{\tilde \bff}$ that end at nodes in the layers $l\geq (i+\max\{0, l^g_{max}-\Dlt i\})+1$. Therefore, the compositional truncation of $\bff$ along the $l$-th layer ($l\geq i$) is identical to the compositional truncation of $ \tilde \bff$ along the $(l+\max\{0, l^g_{max}-\Dlt i\})$-th layer. Because Lipschitz constants are defined based on compositional truncations, (\ref{eq_lipschitz1}) holds true.
$\blacklozenge$\\

The following result is about the change of Lipschitz constants after the composition of two compositional functions. 

\begin{proposition}
\label{prop_lipschitz3}
Given positive integers $d$, $q$ and $s$. Consider two compositional functions $(\bff,\calG^\bff,\calL^\bff)$,  $\bff: \Real^d\rightarrow \Real^q$,  and $(\bfg,\calG^\bfg,\calL^\bfg)$, $\bfg: \Real^q\rightarrow\Real^s$. Let $\bfh=\bfg\circ \bff: \bfx\in \Real^d \rightarrow g(f(\bfx))\in \Real^s$. Let $L^\bff$, $L^\bfg$ and $L^\bfh$ be some Lipschitz constants of $\bff$, $\bfg$ and $\bfh$ under a $p$-norm.
\begin{enumerate}
\item
Given any node $f_{i,j}$ in $\calG^\bff$. As a node in $\calG^\bfh$,  a Lipschitz constant associate with this node is
$L^\bff_{i,j}L^\bfg$.
\item 
A Lipschitz constant of $\bfh$ is $
L^\bfh=L^\bff L^\bfg$.
\end{enumerate}
\end{proposition}
\noindent{Proof}. Part (2) is well known. For part (1), 
let $\bar \bff(z_1,\cdots,z_j,\cdots)$ be the truncation of $\bff$ along the $i$-th layer in which $f_{i,j}$ is located. For $\bfh$, the truncation along the layer of the same node is $\bfg\circ\bar \bff$. For any $\Dlt z\in \Real$,
\EQ
\norm{\bfg(\bar \bff(z_1,\cdots,z_j+\Dlt z,\cdots))-\bfg(\bar \bff(z_1,\cdots,z_j,\cdots))}_p\\
\leq L^\bfg\norm{\bar \bff(z_1,\cdots,z_j+\Dlt z,\cdots)-\bar \bff(z_1,\cdots,z_j,\cdots)}_p\\
\leq  L^\bfg L^\bff_{i,j}\abs{\Dlt z}.
\EE
$\blacklozenge$

\subsection{Some properties of error propagation}
\label{sec_3_3}
For compositional functions in which all nodes have low input dimensions, one can approximate each node using a shallow NN. Substituting these neural networks for the corresponding nodes that they approximate, the process results in a deep neural network. The error of using this deep neural network as an approximation of the function depends on all ``local errors", i.e., the estimation error of each node. These local errors propagate through the DAG network interactively. In this section, we study the relationship between the Lipschitz constants associated with nodes and the propagation of local errors.  To measure errors, we need to introduce an appropriate norm.  Given a function, $f:D^f\subset \Real^d\rightarrow \Real$, where $D^f$ is a compact set. Its $L^\infty$-norm is the essential supremum  
\EQ
\norm{f}_{L^\infty} = \inf\left\{ M\geq 0; \;\abs{f(\bfx)}\leq M \mbox{ almost everywhere in } D^f\right\}.
\EE
Suppose $\tilde f$ is an approximation of $f$. We say that the approximation error is bounded by $\epsilon >0$ if 
\EQ
\norm{f-\tilde f}_{L^\infty} \leq \epsilon.
\EE
If both functions are continuous, this is equivalent to 
\EQ
\label{eq_errorbound}
\abs{f(\bfx)-\tilde f(\bfx)}\leq \epsilon, & \mbox{ for all } \bfx\in D^f.
\EE
In this paper, all approximations are measured using a uniformly bounded error like the one shown in (\ref{eq_errorbound}). If $\bff$ is a vector valued function, a $p$-norm is used to measure the estimation error
\EQ
\label{eq_errorbound1}
\norm{\bff(\bfx)-\tilde \bff(\bfx)}_p\leq \epsilon, & \mbox{ for all } \bfx\in D^\bff.
\EE

When deep neural networks are used to approximate compositional functions, two types of algebraic operations are frequently used. One is node substitution and the other  is composition. The following propositions are about the relationship between the approximation errors before and after an algebraic operation. 
\begin{proposition}
\label{prop_lipschitz2}
Let $(\bff,\calG^\bff,\calL^\bff)$ be a compositional function. Let $\{h_1,h_2,\cdots,h_K\} \subseteq \calV^\bff\setminus \calV^\bff_I$ be a set of nodes. Under a $p$-norm, let $L_j^\bff>0$ be a Lipschitz constant associated with $h_j$, $1\leq j\leq K$. Suppose 
\EQ
\left\{ (\tilde h_j, \calG^{{\tilde h}_j}, \calL^{{\tilde h}_j}); 1\leq j\leq K\right\}
\EE 
is a set of compositional functions in which $\tilde h_j$ has the same domain as $h_j$ and a compatible range for substitution. Assume that
\EQ
\label{eq_sub_1}
\abs{{\tilde h}_j(\bfw) - h_j(\bfw)} \leq \epsilon_j, &\mbox{for all } \bfw \mbox{ in the domain of } h_j,
\EE
where $\epsilon_j$, $j=1,2,\cdots, K$, are some positive numbers. Let $\tilde \bff$ be the function obtained by substituting $\tilde h_j$ for $h_j$, $j=1,2,\cdots,K$. Then 
\EQ
\label{eq_sub_2}
\norm{\tilde \bff (\bfx) -\bff(\bfx)}_p \leq \ds\sum_{j=1}^K L_j^\bff\epsilon_j, &\mbox{for all }  x \mbox{ in the domain of } \bff .
\EE
\end{proposition}

\noindent{\it Proof}. We prove the case for $K=1$ and $2$. The proof can be extended to arbitrary $K$ through mathematical induction. Let $K=1$. Let $\bar \bff$ and $\bar{\tilde \bff}$ be the compositional truncation of $\bff$ and $\tilde \bff$ along the layer in which $h_1$ is located. From the proof of Proposition \ref{prop_lipschitz1},  we know that $\bar \bff=\bar{\tilde \bff}$ . The following fact follows the definition of compositional truncations,
\EQ
\bff(\bfx)=\bar \bff (\bfz(\bfx)),\\
{\tilde \bff}(\bfx)=\bar{\tilde \bff}(\tilde \bfz(\bfx)),
\EE
where $\bfz(\bfx)$, the dummy input of the truncated function, takes the value of the corresponding nodes in $\calG^\bff$ before the truncation. Therefore, it depends on $\bfx$. The value of ${\tilde \bfz}(\bfx)$ is similarly defined. Without loss of generality, assume that $h_1$ is the first node in its layer. Then, $z_1$ takes the value of $h_1$ in the evaluation of $\bff(\bfx)$, and ${\tilde z}_1$ takes the value of ${\tilde h}_1$. The value of $z_1$ is different from that of $\tilde z_1$ because $h_1$ is replaced by $\tilde h_1$. All other components in $z$ and $\tilde z$ are equal because $h_1$ is the only node that is substituted. We have
\EQ
\begin{aligned}
\norm{\bff(\bfx)-\tilde \bff(\bfx)}_p&=\norm{\bar \bff(\bfz)-\bar{\tilde \bff}(\tilde \bfz)}_p\\
&=\norm{\bar \bff(\bfz)-\bar{ \bff}(\tilde \bfz)}_p\\
&\leq L^\bff_1\abs{z_1-\tilde z_1}\\
&=L^\bff_1\abs{h_1(\bfw)-\tilde h_1( \bfw)}, &  \\
&\leq L^\bff_1\epsilon_1,
\end{aligned}
\EE
where $\bfw$ takes the value of nodes that have edges pointing to  $h_1$.
If $K=2$, suppose $\{ h_1, h_2\}$ is ordered in consistent with the layer number of the nodes, i.e., $\calL^\bff(h_1)\leq \calL^\bff(h_2)$. Let $\bfg$ be the function obtained by substituting $\tilde h_1$ for $ h_1$ in $\calG^\bff$; and $\tilde \bff$ be the function obtained by substituting both  $\tilde h_1$  and $\tilde h_2$ for $h_1$ and $ h_2$, respectively. When the node $h_2\in \calV^\bff$ is considered as a node in $\calG^\bfg$, its associated Lipschitz constant as a node in $\calG^\bff$ can be carried over to $\calG^\bfg$ because  $\calL^\bff(h_1)\leq \calL^\bff(h_2)$ (Proposition \ref{prop_lipschitz1}). Therefore, $L_2^\bff$ is a Lipschitz constant associated with $h_2$ as a node in $\calG^\bfg$. Applying the proved result for $K=1$ to both $\bff(\bfx)-\bfg(\bfx)$ and $\bfg(\bfx)-\tilde \bff(\bfx)$, we have
\EQ
\norm{\bff(\bfx)-\tilde \bff(\bfx)}_p\\
\leq \norm{\bff(\bfx)-\bfg(\bfx)}_p+\norm{\bfg(\bfx)-\tilde \bff(\bfx)}_p\\
\leq L^\bff_1 \epsilon_1+L^\bff_2\epsilon_2
\EE
for all $x$ in the domain of $\bff$. 
$\blacklozenge$

Given a function $\bff: \Real^d\rightarrow \Real^d$. A sequence of repeated self-compositions is denoted by
\EQ
(\bff(\cdot))^k(\bfx):= \overbrace{\bff\circ \bff\circ \cdots \circ \bff}^k(\bfx).
\EE

\begin{proposition}
\label{prop4}
Consider compositional functions $(\bff,\calG^\bff,\calL^\bff)$, $\bff: \Real^d \rightarrow \Real^d$, $(\bfg,\calG^\bfg,\calL^\bfg)$, $\bfg: \Real^q \rightarrow \Real^d$, and $(\bfh,\calG^\bfh,\calL^\bfh)$, $\bfh: \Real^d \rightarrow \Real^q$. Suppose that the domains and ranges of them are compatible for compositions.  
Let $L^\bff$ and $L^\bfh$ be Lipschitz constants of $\bff$ and $\bfh$ under a $p$-norm. Suppose  $\tilde \bff$,  $\tilde \bfg$ and  $\tilde \bfh$ are  functions satisfying 
\EQ
\norm{\bff(\bfx)-\tilde \bff(\bfx)}_p\leq e_1, & \bfx \mbox{ in the domain},\\
\norm{\bfg(\bfx)-\tilde \bfg(\bfx)}_p\leq e_2, & \bfx \mbox{ in the domain},\\
\norm{\bfh(\bfx)-\tilde \bfh(\bfx)}_p\leq e_3, & \bfx \mbox{ in the domain},
\EE
for some $e_1>0$, $e_2>0$ and $e_3>0$. Given any integer $K>0$, we have 
\EQ
\label{eq_compoerror}
\norm{(\bff(\cdot))^K\circ \bfg(\bfx)-(\tilde \bff(\cdot))^K\circ {\tilde \bfg}(\bfx)}_p\leq  \Fr{(L^\bff)^K -1 }{L^\bff-1}e_1 + (L^\bff)^K e_2.
\EE
and
\EQ
\label{eq_compoerror1}
\norm{\bfh\circ(\bff(\cdot))^K(\bfx)-\tilde \bfh\circ(\tilde \bff(\cdot))^K(\bfx)}_p\leq L^\bfh \Fr{(L^\bff)^K -1 }{L^\bff-1}e_1 + e_3.
\EE
\end{proposition}
\noindent {\it Proof}. 
The inequality (\ref{eq_compoerror}) is true if $K=1$ because
\EQ
\norm{\bff\circ \bfg(\bfx)-\tilde \bff\circ {\tilde \bfg}(\bfx)}_p\\
\leq \norm{\bff\circ \bfg(\bfx)-\bff\circ {\tilde \bfg}(\bfx)}_p + \norm{\bff\circ {\tilde \bfg}(\bfx)-\tilde \bff\circ {\tilde \bfg}(\bfx)}_p\\
\leq L^\bff e_2+e_1.
\EE
Suppose the inequality holds if the power of $\bff(\cdot)$ is $K-1$. Then,
\EQ
\norm{(\bff(\cdot))^K\circ \bfg(\bfx)-(\tilde \bff(\cdot))^K\circ \tilde \bfg(\bfx)}_p\\
\leq \norm{\bff\circ (\bff(\cdot))^{K-1}\circ \bfg(\bfx)-\bff\circ(\tilde \bff(\cdot))^{K-1}\circ\tilde \bfg(\bfx)}_p+\norm{\bff\circ(\tilde \bff(\cdot))^{K-1}\circ\tilde \bfg(\bfx)-\tilde \bff\circ(\tilde \bff(\cdot))^{K-1}\circ\tilde \bfg(\bfx)}_p\\
\leq L^\bff  \norm{(\bff(\cdot))^{K-1}\circ \bfg(\bfx)-(\tilde \bff(\cdot))^{K-1}\circ \tilde \bfg(\bfx)}_p+e_1\\
\leq L^\bff \left( \Fr{(L^\bff)^{K-1} -1 }{L^\bff-1}e_1+(L^\bff)^{K-1}e_2\right)+e_1\\
=\Fr{(L^\bff)^K -1 }{L^\bff-1}e_1+(L^\bff)^K e_2.
\EE
The inequality (\ref{eq_compoerror1}) is implied by (\ref{eq_compoerror}) and the following inequality
\EQ
\norm{\bfh\circ(\bff(\cdot))^K(\bfx)-\tilde \bfh\circ(\tilde \bff(\cdot))^K(\bfx)}_p \\
\leq \norm{\bfh\circ(\bff(\cdot))^K(\bfx)- \bfh\circ(\tilde \bff(\cdot))^K(\bfx)}_p+\norm{\bfh\circ(\tilde \bff(\cdot))^K(\bfx)-\tilde \bfh\circ(\tilde \bff(\cdot))^K(\bfx)}_p
\EE
$\blacklozenge$

The following proposition is simply a version of the triangular inequality. It is useful when a sequence of approximations are applied to a function. 
\begin{proposition}
\label{prop_triangular}
Given a sequences of functions $\bff_j: D\subset \Real^d\rightarrow \Real^q$, $1\leq j\leq K$. Suppose $\{ e_j\}_{j=1}^{K-1}$ is a sequence of positive numbers satisfying
\EQ
\norm{\bff_{j+1}(\bfx)- \bff_j(\bfx)}_p\leq e_j, & \mbox{ for }\bfx\in D, \; 1\leq j\leq K-1.
\EE
Then, 
\EQ
\norm{\bff_1(\bfx)-\bff_K(\bfx)}_p\leq \ds\sum_{j=1}^{K-1} e_j.
\EE
\end{proposition}

\section{Neural networks as compositional functions}
\label{sec_4}

Let $W^{\infty}_{m,d}$ be the space of scalar valued functions whose derivatives up to $m$-th order exist almost everywhere in $D^f\subset \Real^d$ and they are all essentially bounded.  The Sobolev norm of $f\in W^{\infty}_{m,d}$ is defined by
\EQ
\norm{f}_{W^{\infty}_{m,d}}=\ds\sum_{0\leq \bfk, \abs{\bfk}\leq m}\norm{\Fr{\partial^{\abs{\bfk}} f}{\partial \bfx^{\bfk}}}_{L^\infty}
\EE
where $\bfk=\MT k_1, k_2, \cdots,k_d\EM$ represents a vector of integers and $\abs{\bfk}=\abs{k_1}+\cdots +\abs{k_d}$. Most theorems proved in this paper are based on similar assumptions. They are stated as follows. \\

\noindent \textbf{Assumption A1} (Compositional functions) {\it Given a compositional function $(\bff,\calG^\bff,\calL^\bff)$. We assume that $\bff$ is a function from  $[-R,R]^d$ to $\Real^q$ for some $R>0$ and positive integers $d$ and  $q$. Assume that the nodes,  $\{ f_{i,j}\}$, are functions in $W^\infty_{m_{i,j},d_{i,j}}$ for some integers $m_{i,j}\geq 1$ and $d_{i,j}\geq 1$. The domain of $f_{i,j}$ is a  hypercube with edge length $R_{i,j}>0$. The ranges and domains of all nodes are compatible for composition, i.e.,  if $(f_{i,j}, f_{l,k})$ is an edge in $\calG^\bff$, then the range of $f_{i,j}$ is contained in the interior of the domain of $f_{l,k}$. }\\

In the following, we introduce four key features of compositional functions. These features play a critical role in the estimation of the complexity of deep neural networks when approximating compositional functions. 

\begin{definition} (Features of compositional functions)
\label{def_feature}
Given a compositional function $(\bff,\calG^\bff,\calL^\bff)$,  $\bff: [-R,R]^d\rightarrow\Real^q$, that satisfies A1. The notations $r^\bff_{max}$, $\Lambda^\bff$ and $L^\bff_{max}$ represent positive numbers satisfying 
\EQ
\begin{aligned}
\label{eq_basicbounds}
 d_{i,j} /m_{i,j} &\leq r^\bff_{max}, \\
\max\{(R_{i,j})^{m_{i,j}},1\} \norm{f_{i,j}}_{W^{\infty}_{m_{i,j},d_{i,j}}}&\leq \Lambda^\bff, \\
\abs{L_{i,j}}&\leq L^\bff_{max}, 
\end{aligned}
\EE
for all nodes in $\calV^\bff_G$, where $L_{i,j}$ is defined under a fixed $p$-norm. These upper bounds, $r^\bff_{max}$, $\Lambda^\bff$ and $L^\bff_{max}$, together with $\abs{\calV^\bff_G}$ are called features of the compositional function. 
\end{definition}

\begin{example} \label{example:Lorenz96_rhs} \rm
The function $\bff: [-R,R]^{d} \subset \Real^{d}\rightarrow \Real^d$ in (\ref{eq:lorenz96_rhs}) is the vector field that defines the Loranz-96 model \cite{lorenz96}, 
\begin{eqnarray}\label{eq:lorenz96_rhs}
\bff & = & 
\left[
\begin{array}{c}
 x_0(x_{2}-x_{-1}) - x_1 +F   \\
  x_{1}(x_{3}-x_0) - x_2 +F \\
  \vdots  \\
 x_{i-1}(x_{i+1}-x_{i-2}) - x_i +F\\
 \vdots\\
 x_{d-1}(x_{d+1}-x_{d-2}) - x_d+F 
\end{array}
\right],
\end{eqnarray} 
where $x_{-1}=x_{d-1}$, $x_0=x_d$, $x_{d+1}=x_1$ and $F$ is a constant. Let's treat $\bff$ as a compositional function, $(\bff,\calG^\bff,\calL^\bff)$, with a DAG shown in Figure \ref{fig_Lorenz96}. 
\begin{figure}[ht!]
\centering
\includegraphics[width = 4in]{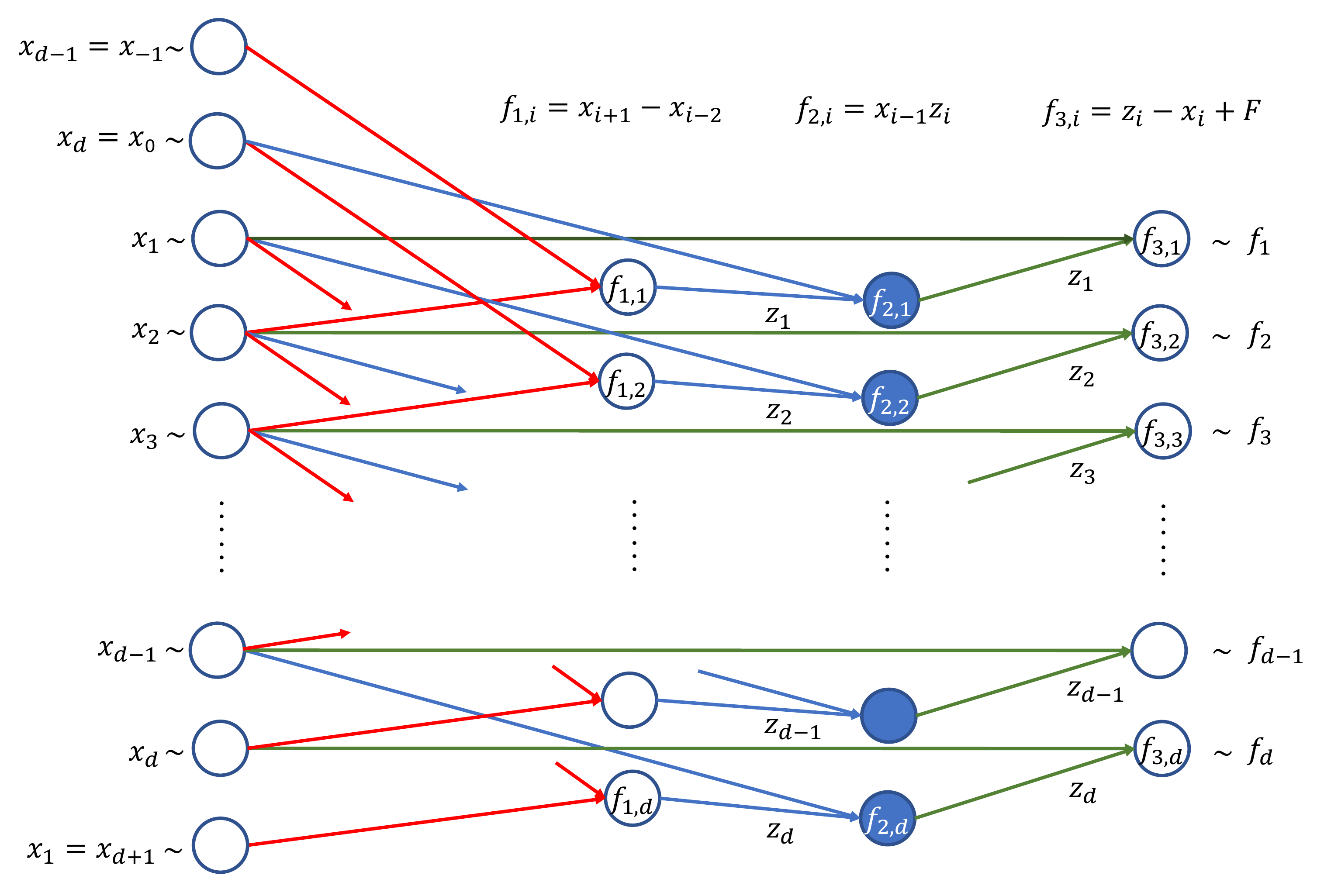}
\caption{DAG structure of the function (\ref{eq:lorenz96_rhs}). For a clear illustration, edges pointing to the first, the second, and the third layer are shown in red, blue and green respectively. }
\label{fig_Lorenz96}
\end{figure}
All general nodes in $(\bff,\calG^\bff,\calL^\bff)$ are located in the second layer. They are defined by
\EQ
f_{2,j}(x_{j-1},z_j) = x_{j-1}z_j, \ j=1,\cdots, d.
\EE
The formulae of linear nodes are shown in Figure \ref{fig_Lorenz96}. All general nodes have the dimension $d_{2,j}=2$ and the domain $[-2R,2R]^2$. To compute the Lipchitz constant associated with the general node, $f_{2,j}$, we construct the truncation of  $(\bff,\calG^\bff,\calL^\bff)$ along the second layer, which is given by
\EQ
\bar \bff(z_1,\cdots,z_{2d}) & = & 
\left[
\begin{array}{ccc}
  z_1-z_{d+1}   \\
  z_2 -z_{d+1}    \\
  \vdots \\
  z_d-z_{2d}  
\end{array}
\right],
\EE
where the dummy inputs $\ z_i \in [-2R,2R]$ for all $i = 1,\cdots, 2d$. Clearly $L_{2,j}=1$, for all $j=1,\cdots, d$. 
Since $f_{2,j}$ is smooth, we can set $m_{2,j}=m$ for any integer $m\geq 2$. Thus,
\[\| f_{2,j} \|_{W^\infty_{m,2}}  =  1+2R+4R^2. \]
Now it is straightforward to compute the features of the compositional function $(\bff,\calG^\bff,\calL^\bff)$:
\EQ \label{eq:Lorenz96_features}
r_{max}^\bff & = & 2/m,\\
\Lambda^\bff & = & \mbox{max}\{(2R)^m,1\}(1+2R+4R^2),\\
L^\bff_{max} & = & 1,\\
\abs{\calV^\bff_G} & = & d.
\EE
\end{example}

Shallow and deep neural networks are defined as follows. They are compositional functions in which all general nodes are based on a single function. 

\begin{figure}[h!]
\centering
\includegraphics[width = 2.25in]{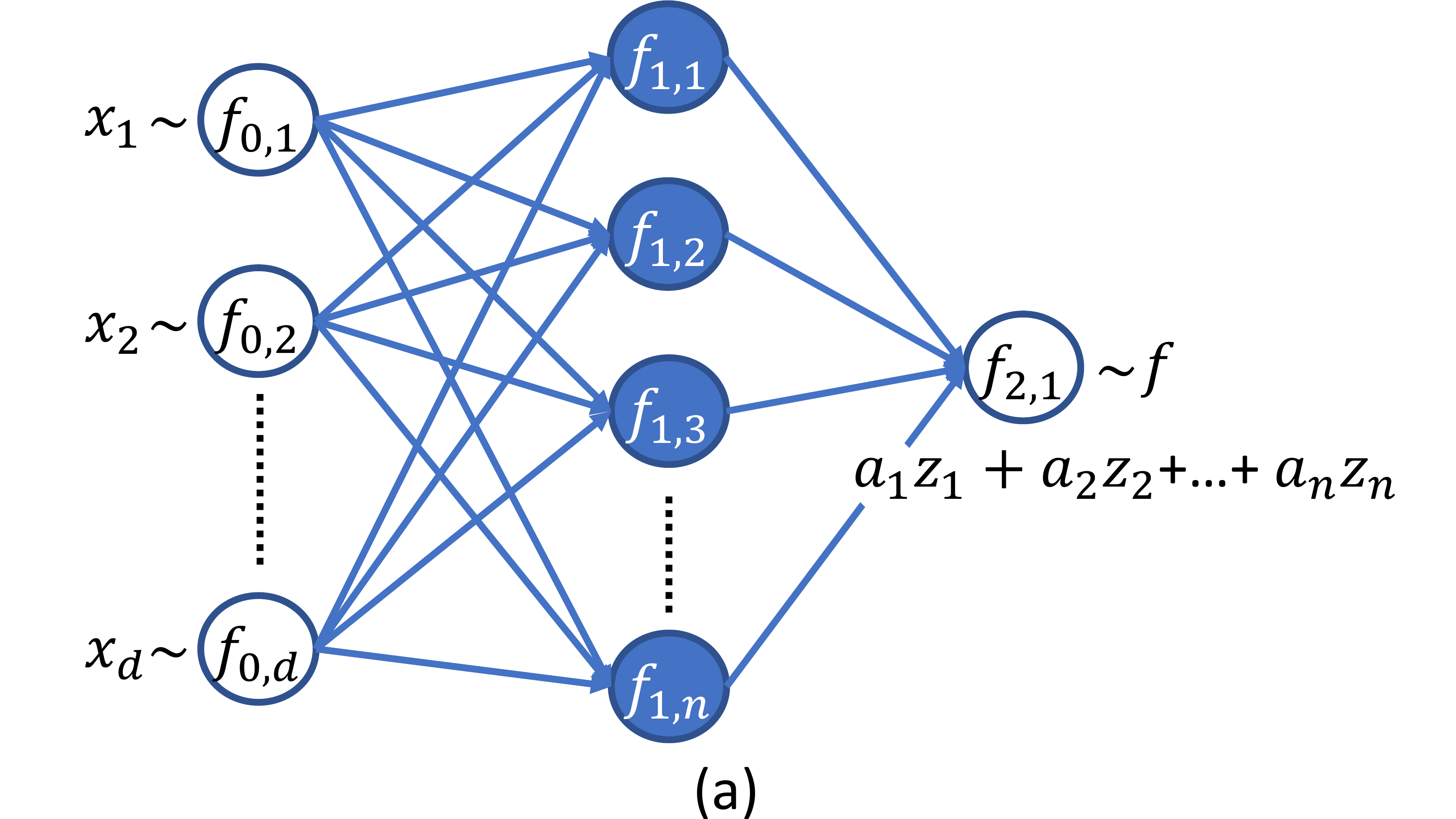} \;\;\; \includegraphics[width = 2.25in]{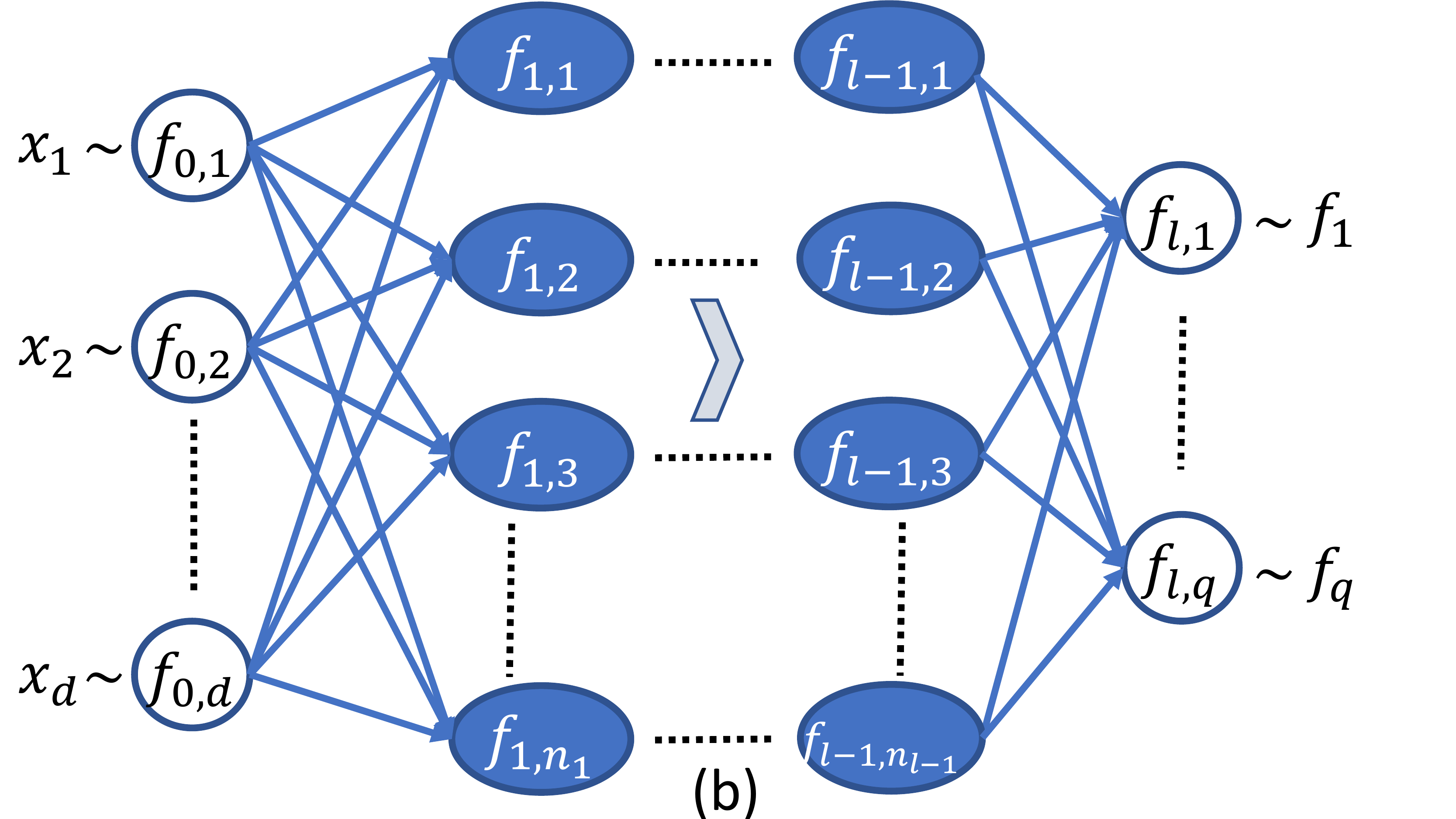} 
\caption{(a): DAG of typical shallow neural networks. (b): DAG structure of deep neural networks. }
\label{fig_NN}
\end{figure}

\begin{definition} (Neural network)
\label{def_NN}
A Neural network is a compositional function of a special type. Given a function $\sigma: \Real\rightarrow \Real$, which is called an activation function. 
\begin{enumerate}
\item A {\it shallow NN} is a function
\EQ
f(\bfx)=\ds\sum_{j=1}^n a_j\sigma \left(\bfw_j^T \bfx +b_j \right)
\EE
where $a_j, b_j\in \Real$ and $\bfw_j\in \Real^d$, $1\leq j\leq n$, are parameters. Its DAG is shown in Figure \ref{fig_NN}(a), which has one hidden layer and a linear output node. Each node in the hidden layer, $f_{1,j}$, is a function of the form $\bfx\rightarrow \sigma (\bfw^T\bfx+b)$.  In this paper, unless otherwise stated,  a shallow neural network is a scalar valued function. 
\item A deep neural network is a compositional function $(\bff,\calG^\bff,\calL^\bff)$, $\bff: \Real^d\rightarrow \Real^q$, that has multiple hidden layers (Figure \ref{fig_NN}(b)). The output layer consists of linear nodes. Every node in hidden layers is a function of the form $\bfz\rightarrow \sigma (\bfw^T\bfz+b)$. 
\item A node in a hidden layer is called a neuron. The complexity of a neural network or deep neural network is the total number of neurons in it. 
\end{enumerate}
\end{definition}

\begin{remark}
\label{remark0}
In many applications of neural networks, edges connecting neurons do not skip layers. This is different from the deep neural network defined in Definition \ref{def_NN} where edges can skip layers. It is addressed later in this section that one can construct neural networks without edges that skip layers by adding identity functions in the DAG. We are not the first ones in the study of DAG based neural networks. For instance, error analysis and potential computational advantages were addressed for DAG based neural networks in \cite{chiu,mhaskar1,mhaskar2,poggio} and references therein.  In this paper, all neural networks are DAG based. For the reason of simplicity, we do not always specify the layered DAG, $\calG^{\bff^{NN}}$ and $\calL^{\bff^{NN}}$, when introducing a neural network. 
\end{remark}

\begin{figure}[h!]
\centering
\includegraphics[width = 2.25in]{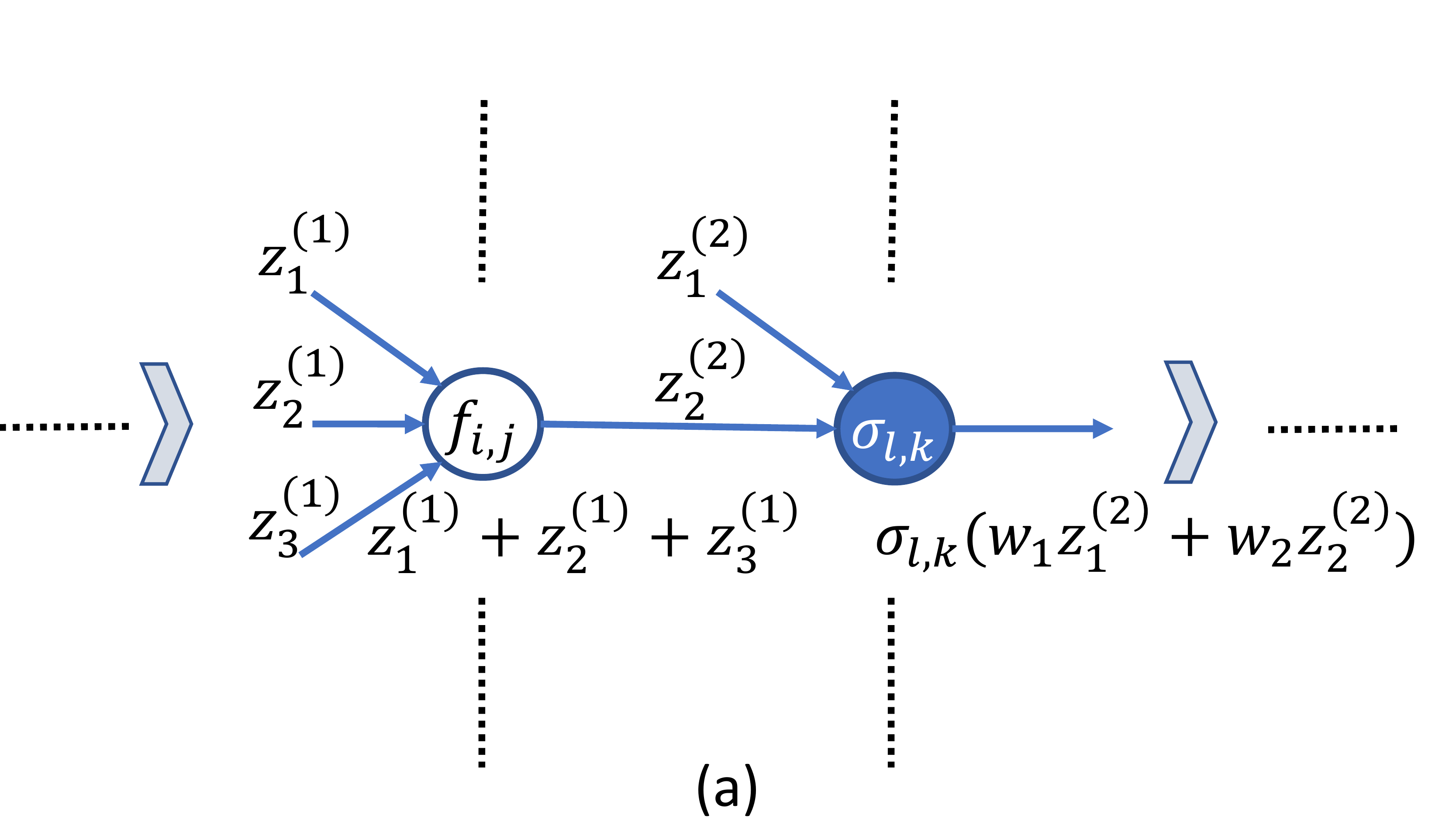} \;\;\; \includegraphics[width = 2.25in]{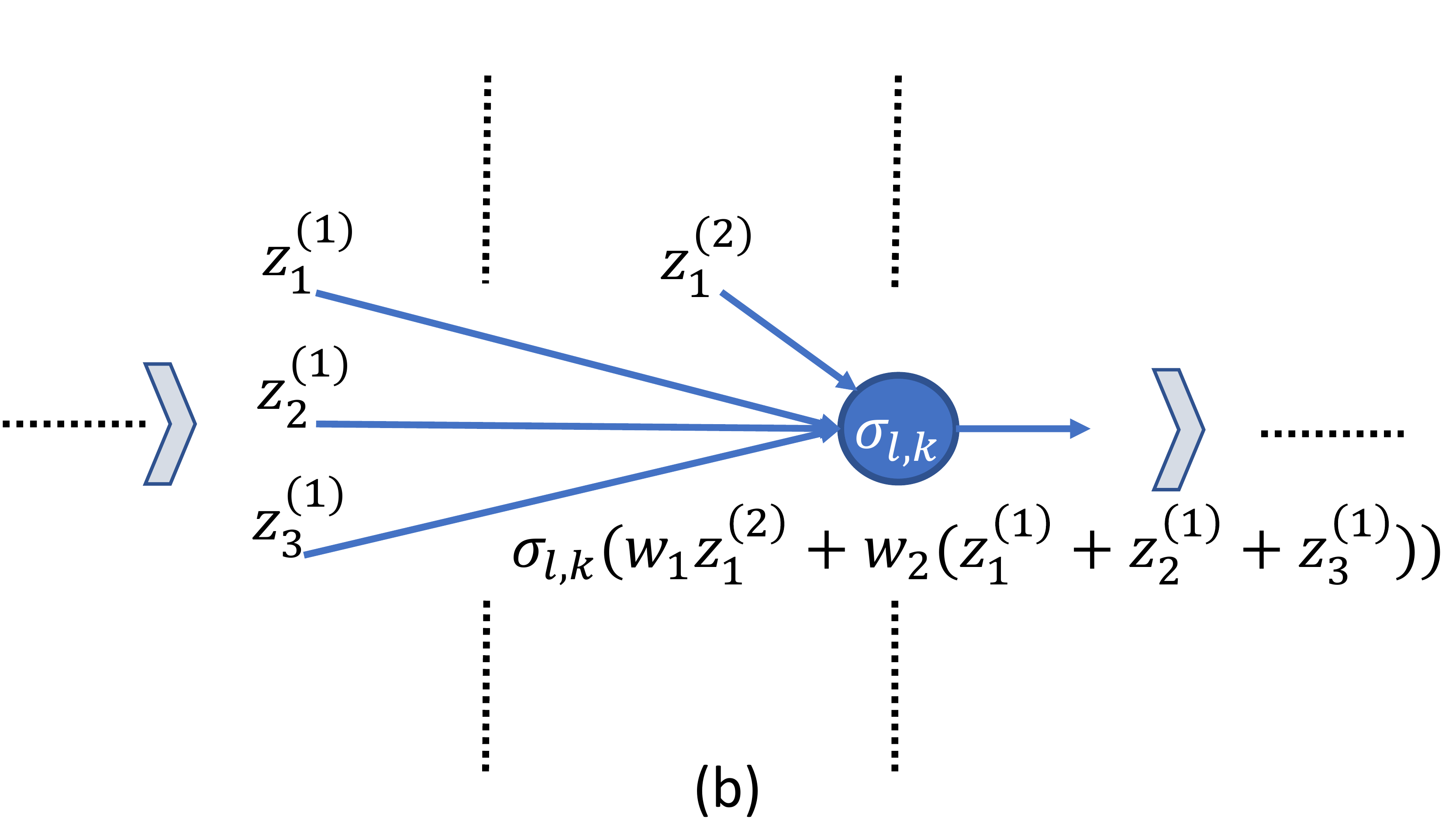}
\caption{An example of merging linear nodes into neurons. The linear node $f_{i,j}$ in (a) can be removed by merging it into the neuron $\sigma_{l,k}$. The new DAG is shown in (b). }
\label{fig_DAG_absorb}
\end{figure}

\begin{remark} 
\label{remark1}
In Definition \ref{def_NN}, there is no linear nodes in the hidden layers. This requirement is not essential because one can easily modify the DAG of a neural network to remove all linear nodes by merging them into neurons. Figure \ref{fig_DAG_absorb}(a) shows a portion of a DAG in which $f_{i,j}$ is a linear node. It evaluates the summation of the three inputs. This linear node is directly connected with a neuron $\sigma_{l.k}$.  We can remove the linear node; and then connect its inward edges directly to the neuron. Adjusting the neuron's parameters accordingly, the result is a simplified DAG shown in Figure \ref{fig_DAG_absorb}(b). Following the same process, one can remove all linear nodes in hidden layers without increasing the number of neurons. The compositional function resulting from this process is a deep neural network defined in Definition \ref{def_NN}. In the rest of the paper, we call a function a deep neural network even if it has linear nodes in addition to neurons, with the understanding that the function is equivalent to a deep neural network as defined in Definition \ref{def_NN} by merging all linear nodes into neurons. 
\end{remark}

Based on Remark \ref{remark1}, it is obvious that the family of neural networks is closed under several algebraic operations, including linear combination, substitution, and composition. 

\begin{proposition} Given two neural networks, $\bff^{NN}$ and $\bfg^{NN}$. Suppose that their domains and ranges are compatible for the operations discussed below. Then the following statements hold true. 
\begin{enumerate}
\item The linear combination, $a\bff^{NN}+b\bfg^{NN}$ for any constants $a$ and $b$, is a neural network.
\item The compositional function, $\bfg^{NN}\circ \bff^{NN}$, is a neural network. 
\item If a neuron in $\bff^{NN}$ is substituted by $g^{NN}$, the resulting compositional function is still a neural network.
\end{enumerate}
\end{proposition}

Based on this proposition, one can compose complicated deep neural networks using simple and shallow neural networks through sequences of algebraic operations. When this approach is applied to approximate functions, the approximation error of a complicated neural network depends on the accuracy of individual shallow neural networks that are used as building blocks
in the composition. The following theorem on the approximation error of shallow neural networks, which is proved in \cite{mhaskar}, is fundamental in the proof of several theorems about deep neural network approximations of various subjects, including the solutions of differential equations, optimization problems and optimal control. 

\begin{theorem} (Shallow neural network approximation \cite{mhaskar})
\label{thm1}
Let $m, d, n\geq 1$ be integers. Let $\sigma: \Real\rightarrow \Real$ be an infinitely differentiable function in some open interval in $\Real$. Assume that there is a point, $b\in \Real$, at which $\sigma$ and all its derivatives do not equal zero. For any positive integer, $n$, there exist vectors $\{ \bfw_j\}_{j=1}^n\subset \Real^d$ and a constant $C$ satisfying the following property. For any function, $f: [-1,1]^d\rightarrow \Real$ in $ W^{\infty}_{m,d}$, there exist coefficients $a_j(f)$ such that 
\EQ
\label{eq_bound1}
\abs{f(\bfx)-\ds\sum_{j=1}^n a_j(f)\sigma \left(\bfw_j^T\bfx +b \right)}\leq Cn^{-m/d}\norm{f}_{W^{\infty}_{m,d}}, & \mbox{ for all } \bfx \in  [-1,1]^d.
\EE
The constant $C$ is determined by $d$ and $m$ but is independent of $f$ and $n$. 
\end{theorem}

This theorem is based on the assumption that the domain of $f$ is a hypercube whose edge length is 2. However, the domain of each node in a compositional function do not necessarily satisfy this assumption. The following is an error upper bound for $f$ defined in a hypercube that has an arbitrary edge length. The result is proved for domains centered at the origin. Applying translation mappings, the result can be extended to hypercubes centered at any point. 

\begin{corollary}
\label{coro1}
Making the same assumptions as in Theorem \ref{thm1} except that the domain of  $f$ is  $[-R,R]^d$.  Then, there exist vectors $\{ \bfw_j\}_{j=1}^n\subset \Real^d$, $b\in \Real$, a constant $C$ and coefficients $a_j(f)$ such that 
\EQ
\label{eq_bound2}
\abs{f(\bfx)-\ds\sum_{j=1}^n a_j(f)\sigma \left(\bfw_j^T\bfx +b \right)}\leq Cn^{-m/d}\max\{R^m,1\}  \norm{f}_{W^{\infty}_{m,d}}, &  \bfx \in  [-R,R]^d.
\EE
The constant $C$ is determined by $d$ and $m$ but independent of $f$ and $n$. 
\end{corollary}

\noindent {\it Proof}. Define a mapping $\bfx=R \tilde \bfx$ and a new function $\tilde f(\tilde \bfx)=f(R\tilde \bfx)$. The domain of the new function is $[-1, 1]^d$. From Theorem \ref{thm1}, we have
\EQ
\label{eq_1}
\abs{\tilde f(\bfx)-\ds\sum_{j=1}^n a_j({\tilde f})\sigma \left(\bfw_j^T\bfx +b \right)}\leq Cn^{-m/d}\norm{\tilde f}_{W^{\infty}_{m,d}}
\EE
for all $\bfx$ in $[-1, 1]^d$. It is obvious that 
\EQ
\label{eq_2}
\norm{\Fr{\partial^{\abs{\bfk}} {\tilde f}}{\partial {\tilde \bfx}^{\bfk}}}_{L^\infty}=\norm{\Fr{\partial^{\abs{\bfk}} f}{\partial x^{\bf k}}}_{L^\infty}R^{\abs{\bfk}}.
\EE
Therefore,
\EQ
\label{eq_3}
\begin{aligned}
\norm{\tilde f}_{W^{\infty}_{m,d}}
&=\ds\sum_{0\leq {\bf k}, \abs{\bf k}\leq m}\norm{\Fr{\partial^{\abs{\bf k}} \tilde f}{\partial \tilde \bfx^{\bf k}}}_{L^\infty}\\
&=\ds\sum_{j=0}^m R^j\left( \sum_{\abs{\bf k}=j}\norm{\Fr{\partial^{\abs{\bf k}} f}{\partial \bfx^{\bf k}}}_{L^\infty}\right) \\
&\leq \ds\sum_{j=0}^m \max\{R^j,1\}\left( \sum_{\abs{\bf k}=j}\norm{\Fr{\partial^{\abs{\bf k}} f}{\partial \bfx^{\bf k}}}_{L^\infty}\right) \\
&\leq \max\{R^m,1\} \left(\ds\sum_{j=0}^m \sum_{\abs{\bf k}=j}\norm{\Fr{\partial^{\abs{\bf k}} f}{\partial \bfx^{\bf k}}}_{L^\infty}\right)\\
&= \max\{R^m,1\} \norm{ f}_{W^{\infty}_{m,d}}.
\end{aligned}
\EE
Then, (\ref{eq_1}) and (\ref{eq_3}) imply (\ref{eq_bound2}).
$\blacklozenge$

Consider a compositional function $(\bff,\calG^\bff,\calL^\bff)$ in which the nodes, $\{ f_{i,j}\}$, are functions in $W^\infty_{m_{i,j},d_{i,j}}$ for some positive integers $m_{i,j}, d_{i,j}\geq 1$. From Theorem \ref{thm1}, there exists a set of constant $C_{i,j}$ and a set of shallow neural networks, $\{ f_{i,j}^{NN}\}$, such that 
\EQ
\label{eq_sub2}
\abs{f_{i,j}(\bfx)-f_{i,j}^{NN}(\bfx)}\leq C_{i,j}n^{-m_{i,j}/d_{i,j}}\max\{R_{i,j}^{m_{i,j}},1 \}\norm{f_{i,j}}_{W^{\infty}_{m_{i,j},d_{i,j}}}, \mbox{ for all } \bfx \mbox{ in the domain,}
\EE
where $n$ is the complexity, i.e. the number of neurons in $f_{i,j}^{NN}$. Substituting the shallow neural networks for all general nodes in $\calG^\bff$ results in a new compositional function consisting of neurons and linear nodes, i.e. a deep neural network. It is an approximation of $\bff$. Its error upper bound is proved in the following theorem. 

\begin{theorem}
\label{thm2}
Consider a compositional function $(\bff,\calG^\bff,\calL^\bff)$,  $\bff: [-R,R]^d\rightarrow\Real^q$, and a set of its features (Definition \ref{def_feature}). Suppose that the function satisfies A1. 
\begin{enumerate}
\item For any integer $n_{width}>0$, there exists a set of shallow neural networks of width $n_{width}$ that approximates the nodes in $\calV^\bff_G$. The deep neural network, $\bff^{NN}$, resulting from substituting the shallow neural networks for all general nodes in $\calG^\bff$ has the following error upper bound,
\EQ
\label{eq_fNNbound0}
\norm{\bff(\bfx)-\bff^{NN}(\bfx)}_p\leq C (n_{width})^{-1/r^\bff_{max}}, &\mbox{for all } \bfx \in [-R, R]^d,
\EE
where $C$ is a constant. 
\item The constant, $C$, in (\ref{eq_fNNbound0}) depends on the features of $(\bff,\calG^\bff,\calL^\bff)$,
\EQ
\label{eq_fNNbound1}
C= C_1L_{max}^\bff\Lambda^\bff \abs{\calV^\bff_G} ,
\EE
where $C_1$ is a constant determined by $\{ d_{i,j}, m_{i,j}; \;\;  f_{i,j}\in \calV^\bff_G\} $.
\item The complexity of $f^{NN}$ is $\abs{\calV^\bff_G}n_{width}$.
\end{enumerate}
\end{theorem}

\begin{remark} 
Theorem \ref{thm2} implies that approximations using neural networks can avoid the curse of dimensionality, provided that the features of the compositional function do not grow exponentially with the dimension. More specifically, for a compositional function defined in a domain in $\Real^d$, the complexity of its neural network approximation depends directly on the features of the function, but not directly on the dimension $d$. Given a family of compositional functions for $d>0$. If the value of their features do not increase exponentially as $d\rightarrow \infty$, then neural network approximations of these compositional functions do not suffer the curse of dimensionality. We would like to point out that the inequality (\ref{eq_fNNbound0}) can be found in \cite{poggio}. Different from \cite{poggio}, we rigorously prove the inequality as well as the relationship between $C$ and the features of compositional functions. 
\end{remark}

\noindent{\it Proof of Theorem \ref{thm2}}. For every node $f_{i,j} \in \calV^\bff_G$, there exists a shallow neural network, $f_{i,j}^{NN}$, that approximates the node (Theorem \ref{thm1} and Corollary \ref{coro1}). If the number of neurons in $f_{i,j}^{NN}$ is $n_{width}$, we can find one that satisfies
\EQ
\abs{f_{i,j}(\bfz)-f_{i,j}^{NN}(\bfz)}\leq C_{i,j}(n_{width})^{-m_{i,j}/d_{i,j}}\max\{R^m_{i,j},1\}\norm{f_{i,j}}_{W^{\infty}_{m_{i,j},d_{i,j}}}, 
\EE
for all  $\bfz$ in the domain of $f_{i,j}$, 
where $C_{i,j}$ depends on $d_{i,j}$ and $m_{i,j}$ but not $f_{i,j}$. Suppose $C_1$ is the largest $C_{i,j}$. 
From Proposition \ref{prop_lipschitz2}, $\bff^{NN}$ satisfies
\EQ
\label{eq_error_10}
\norm{\bff(\bfx)-\bff^{NN}(\bfx)}_p\\
\leq \ds\sum_{f_{i,j}\in\calV^\bff_G}L_{i,j}^\bff C_{i,j}\max\{R^m_{i,j},1\}\norm{f_{i,j}}_{W^{\infty}_{m_{i,j},d_{i,j}}}(n_{width})^{-m_{i,j}/d_{i,j}}\\
\leq   \ds\sum_{f_{i,j}\in\calV^\bff_G}L_{max}^\bff C_1 \Lambda^\bff (n_{width})^{-1/r^\bff_{max}}\\
= C_1 L_{max}^\bff \Lambda^\bff\abs{\calV^\bff_G}(n_{width})^{-1/r^\bff_{max}}.
\EE
$\blacklozenge$\\

\begin{remark}
The expression of $C$ in (\ref{eq_fNNbound1}) is conservative. From the proof, an alternative error upper bound is in (\ref{eq_error_10}). More specifically, 
\EQ
\norm{\bff(\bfx)-\bff^{NN}(\bfx)}_p
\leq \ds\sum_{f_{i,j}\in\calV^\bff_G}L_{i,j}^\bff C_{i,j}\max\{R^m_{i,j},1\}\norm{f_{i,j}}_{W^{\infty}_{m_{i,j},d_{i,j}}}(n_{width})^{-m_{i,j}/d_{i,j}}
\EE
Moreover, the proof suggests that the number of hidden layers in the neural network (after all linear nodes in hidden layers are merged into neurons) is $l^\bff_{max}$. The output layer consists of all linear nodes. The width of layer $1\leq l\leq l^\bff_{max}$ in the neural network is bounded by $n_{width}\times n_l$, where $n_l$ is the number of general nodes in the $l$-th layer in $\calG^\bff$.  
\end{remark}

\begin{figure}[h!]
\centering
\includegraphics[width = 2.25in]{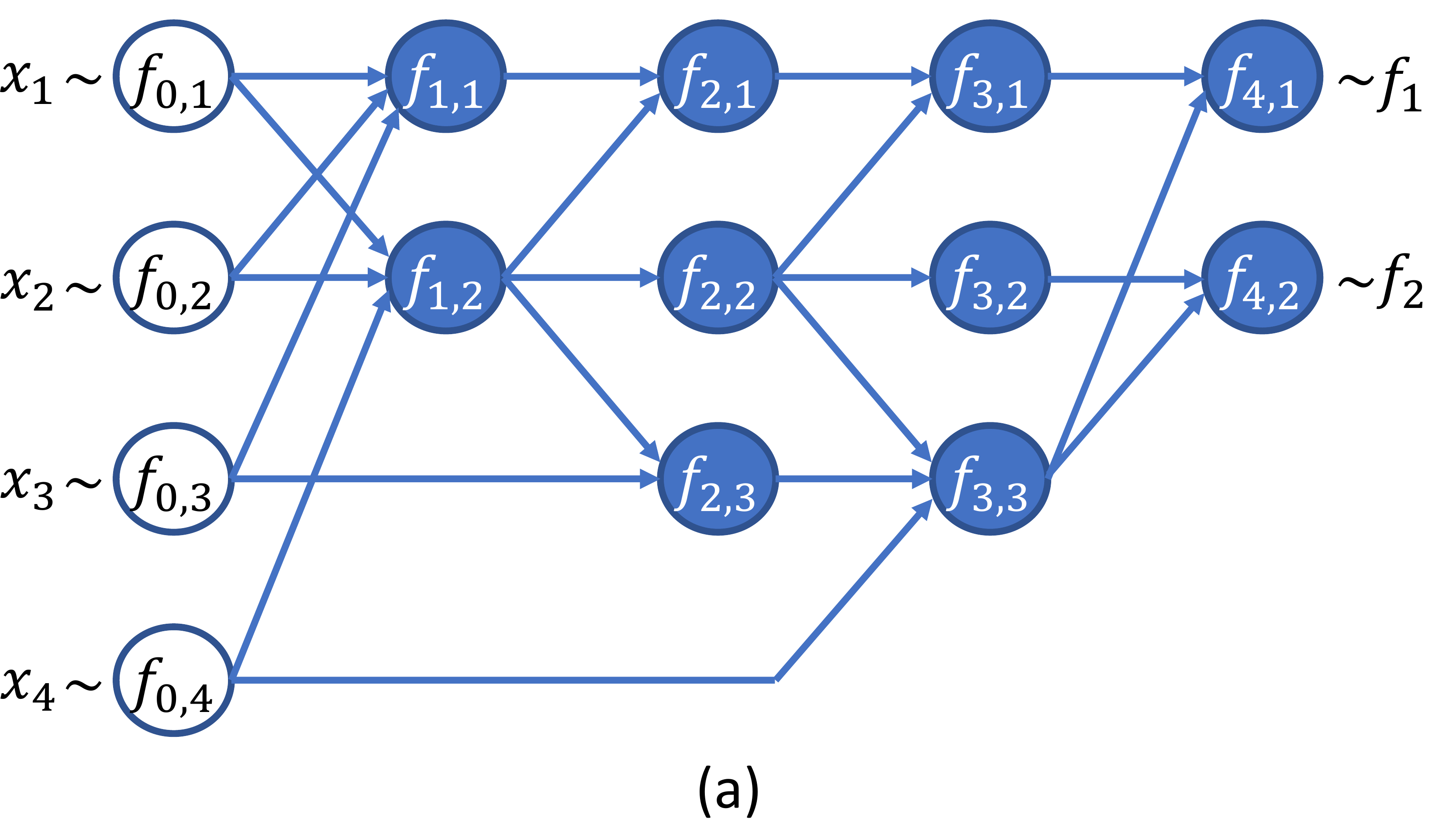} \;\;\; \includegraphics[width = 2.25in]{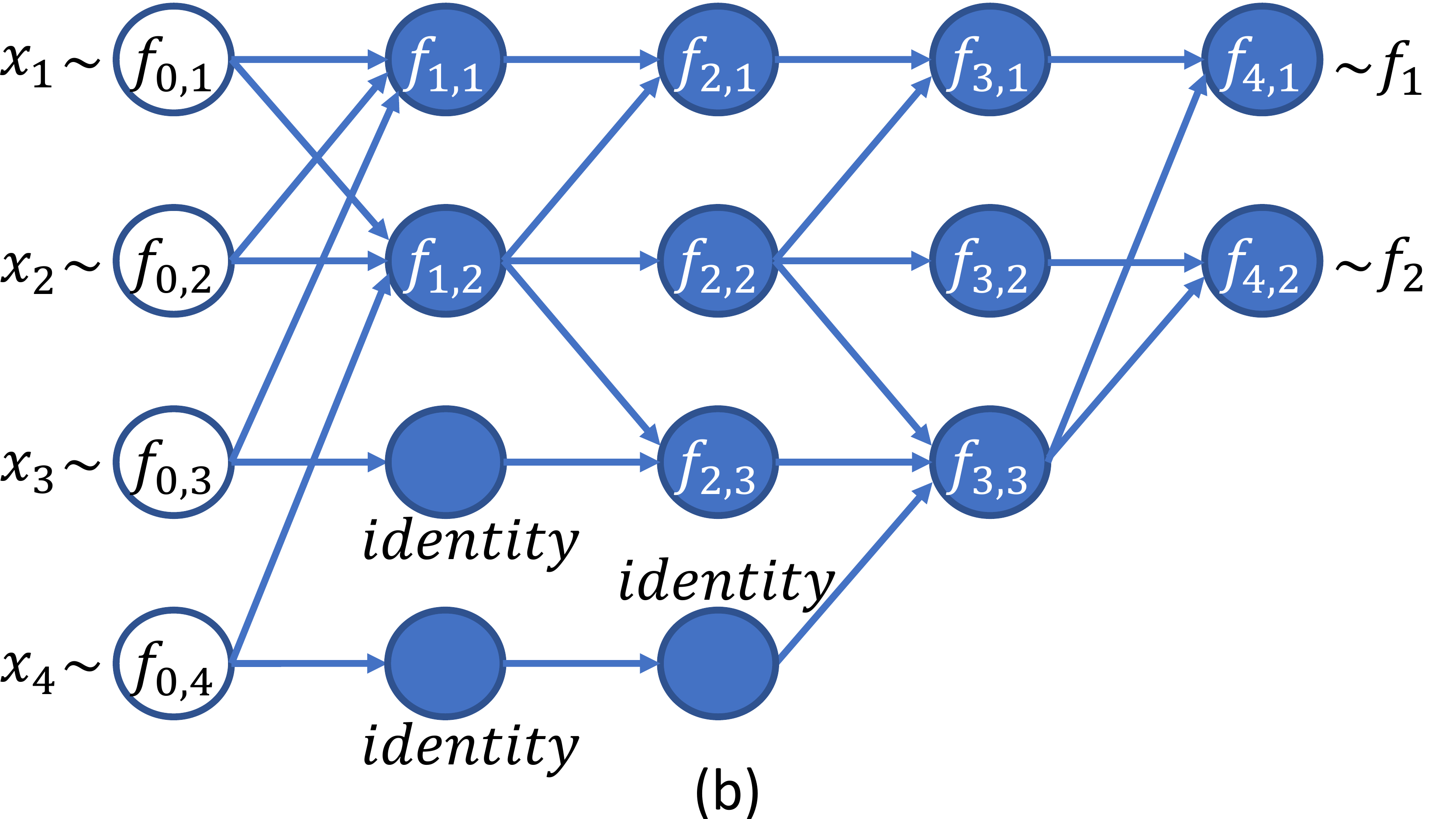}
\caption{Functions in (a) and (b), as input-output relations, are equal to each other. Adding identity nodes, the DAG in (b) has no edge that skips layers.}
\label{fig_idnode}
\end{figure}

\begin{remark} The edges in $\calG^{\bff^{NN}}$ in Theorem \ref{thm2} may skip layers, which is not in the traditional format of deep neural networks. Any compositional function, in fact, allows a DAG in which no edges skip layers. Adding identity functions as nodes is a way to achieve this. Shown in Figure  \ref{fig_idnode}(a) is the DAG of a compositional function in which two edges skip layers, $(f_{0,3}, f_{2,3})$ and $(f_{0,4}, f_{3,3})$. In Figure \ref{fig_idnode}(b), three identity functions are added as nodes to the DAG at each location where an edge skips a layer. Under the new DAG, no edges skip layers. In Theorem \ref{thm2}, if $\bff$ has no edges that skip layers and if we substitute shallow neural networks for all nodes in $\calV^\bff\setminus \calV^\bff_I$ (including identity and linear nodes), then the DAG of $\bff^{NN}$ in Theorem \ref{thm2} has no edges that skip layers. After merging linear nodes in $\calG^{\bff^{NN}}$ into neurons, one can achieve a traditional deep neural network in which edges do not skip layers. 
\end{remark}

\begin{corollary}
\label{thm3}
Consider a compositional function $(\bff,\calG^\bff,\calL^\bff)$ that satisfies the same assumptions made in Theorem \ref{thm2}. Then, there is an $\epsilon_0>0$. For  any $0<\epsilon \leq \epsilon_0$, there always exists a deep neural network, $\bff^{NN}$, satisfying
\EQ
\label{eq_fNNbound4}
\norm{\bff(\bfx)-\bff^{NN}(\bfx)}_p\leq \epsilon, & \bfx \in [-R, R]^d.
\EE
The complexity of $\bff^{NN}$ is
\EQ
\label{eq_fNNbound3}
n \leq C \left( L_{max}^\bff\Lambda^\bff\right)^{r^\bff_{max}}  \abs{\calV^\bff_G} ^{r^\bff_{max}+1} \epsilon^{-r^\bff_{max}},
\EE
where $C$ is a constant determined by $\{ d_{i,j}, m_{i,j}; \;\;  f_{i,j}\in \calV^\bff_G\} $.
\end{corollary}

\noindent{Proof}. 
From Theorem \ref{thm2}, there exists a deep neural network, denoted by $\bff^{NN}$, that satisfy (\ref{eq_fNNbound0})-(\ref{eq_fNNbound1}). Let $n_{width}$ be the smallest integer such that 
\EQ
\label{eq_fNNbound6}
CL_{max}^\bff\Lambda^\bff \abs{\calV^\bff_G}(n_{width})^{-1/r^\bff_{max}}\leq \epsilon.
\EE
Then (\ref{eq_fNNbound4}) holds.  As the smallest integer satisfying (\ref{eq_fNNbound6}), $n_{width}$ satisfies
\EQ
\label{eq_fNNbound5}
n_{width} \leq \left( CL_{max}^\bff\Lambda^\bff \abs{\calV^\bff_G} \right)^{r^\bff_{max}} \epsilon^{-r^\bff_{max}}+1.
\EE
Because $C^{r^\bff_{max}}$ is a constant depending on $\{ d_{i,j}, m_{i,j}; \;\;  f_{i,j}\in \calV^\bff_G\} $ and $n=\abs{\calV^\bff_G}n_{width}$,  (\ref{eq_fNNbound5}) implies 
\EQ
\label{eq_fNNbound7}
n \leq C \left( L_{max}^\bff\Lambda^\bff\right)^{r^\bff_{max}}  \abs{\calV^\bff_G} ^{r^\bff_{max}+1} \epsilon^{-r^\bff_{max}}+ \abs{\calV^\bff_G} .
\EE
If $\epsilon$ is small enough, then we can double the constant $C$ and remove the last term, $ \abs{\calV^\bff_G}$, i.e. (\ref{eq_fNNbound3}) holds. 
$\blacklozenge$\\

\begin{example} \rm
Consider the compositional function (\ref{eq:lorenz96_rhs}) in Example \ref{example:Lorenz96_rhs}, whose features are given in (\ref{eq:Lorenz96_features}). By Theorem \ref{thm2}, for any integer $n_{width}>0$, there exists a deep neural network, $\bff^{NN}$, such that 
\EQ
\norm{\bff(\bfx)-\bff^{NN}(\bfx)}_p\leq C_1 d \left(1+2R+4R^2\right) \mbox{max}\{(2R)^m,1\}  (n_{width})^{-m/2}, \nonumber
\EE
for all $\bfx \in [-R, R]^d$ and all integer $m\geq 2$,
where $C_1$ is a constant determined by $m$. The complexity of $f^{NN}$ is $d n_{width}$. The approximation error and the complexity depend on the dimension, $d$, linearly. If the value of $m$ and the domain size $R$ are set to be constants, the neural network approximation is curse-of-dimensionality free. 
\end{example}

Given a function $f:D\subset \Real^d\rightarrow \Real$. The radius of its range is defined by
\EQ
\label{eq_radius}
\left(\ds\max_{\bfx \in D} \{f(\bfx)\} -\min_{\bfx\in D}\{ f(\bfx)\}\right)/2.
\EE

\begin{theorem}
\label{thm5}
Consider two compositional functions $(\bff,\calG^\bff,\calL^\bff)$, $\bff: [-R, R]^d \rightarrow \Real^q$, and $(\bfg,\calG^\bfg,\calL^\bfg)$, $\bfg: [-R,R]^d \rightarrow \Real^q$. Suppose they both satisfy A1. Let $\bff^{NN}$ and $\bfg^{NN}$ be two neural networks with complexity $n_1$ and $n_2$, respectively. Suppose 
\EQ
\label{eq_e1e2}
\norm{\bff(\bfx)-\bff^{NN}(\bfx)}_p \leq e_1,\\
\norm{\bfg(\bfx)-\bfg^{NN}(\bfx)} _p\leq e_2,\\
\EE
for some positive numbers $e_1$, $e_2$ and for all $\bfx$ in the domain.
\begin{enumerate}
\item Consider $\bfh=a\bff+b\bfg $, where  $a,b\in \Real$ are constants. Then $\bfh^{NN}=a\bff^{NN}+b\bfg^{NN}$ is a neural network satisfying 
\EQ
\label{eq_ad_error}
\norm{\bfh(\bfx)-\bfh^{NN}(\bfx)}_p\leq \abs{a}e_1+\abs{b}e_2
\EE
for all $\bfx$ in the domain. The complexity of $\bfh^{NN}$ is $n_1+n_2$. 
\item Consider $h=\bff^T\bfg$. Let $p=2$ in (\ref{eq_e1e2}).
Then, for any integers $n_{width}>0$ and $m>0$,  there is a deep neural network, $h^{NN}$, satisfying
\EQ
\abs{h(\bfx)-h^{NN}(\bfx)}\\
\leq \ds\max_{\bfx\in [-R, R]^d}\{\norm{\bff(\bfx)}_2\}e_2+\ds\max_{\bfx\in [-R, R]^d}\{\norm{\bfg(\bfx)}_2\}e_1+e_1e_2+C\Lambda q (n_{width})^{-m/2},
\EE
where $C$ is a constant depending on $m$ and $q$. The parameter, $\Lambda$, is defined as follows based on $f_j$ and $g_j$ (the $j$-th component of $\bff$ and $\bfg$, respectively)
\EQ
\label{eq_psi_gamma}
\Lambda = \ds\max_{1\leq j\leq q}\left\{ \max\{(R_j)^m, 1\}(A_jB_j+A_j+B_j+2)\right\},\\
A_j=\ds\max_x\{\abs{f_j(\bfx)}\}, B_j=\ds\max_x\{\abs{g_j(\bfx)}\},\\
R_j=\max\{ \mbox{radii of the ranges of } f_j, g_j\}.
\EE
The complexity of $\bfh^{NN}$ is $n_1+n_2+qn_{width}$. 
\item If $f$ and $g$ are scalar valued compositional functions.  Suppose $g(\bfx)\neq 0$. Consider $h=f/g$. Suppose
\EQ
\label{eq_e22}
e_2\leq \Fr{1}{2}\ds\min_{\bfx \in [-R, R]^d}\{\abs{g(\bfx)}\}.
\EE
Then, for any integers $n_{width}>0$ and $m>0$, there is a deep neural network, $h^{NN}$, satisfying
\EQ
\abs{h(\bfx)-h^{NN}(\bfx)}\leq \Fr{2A}{B^2}e_2+\Fr{2}{B}e_1+C\Lambda (n_{width})^{-m/2},
\EE
where $C$ is a constant depending on $m$, 
\EQ
\label{eq_psi_gamma_2}
\Lambda =  \max\{R^m_1, 1\}\left( m!( A+B)\Fr{1-(1/B)^{m+1}}{B-1} \right),\\
A=\ds\max_{\bfx\in [-R, R]^d}\{\abs{f(\bfx)}\}, \; B=\ds\min_{\bfx\in [-R, R]^d}\{\abs{g(\bfx)}\},\\
R_1=\max\{ \mbox{radii of the ranges of } f, g\}.
\EE
The complexity of $h^{NN}$ is $n_1+n_2+n_{width}$. 
\end{enumerate}
\end{theorem}
\noindent {\it Proof}. The inequality (\ref{eq_ad_error}) is obvious. From Definition \ref{def_addition}, the last layer of the induced DAG for $a\bff^{NN}+b\bfg^{NN}$ consists of all linear nodes. It does not change the complexity of the deep neural network.  

\begin{figure}[h!]
\centering
\includegraphics[width = 2.25in]{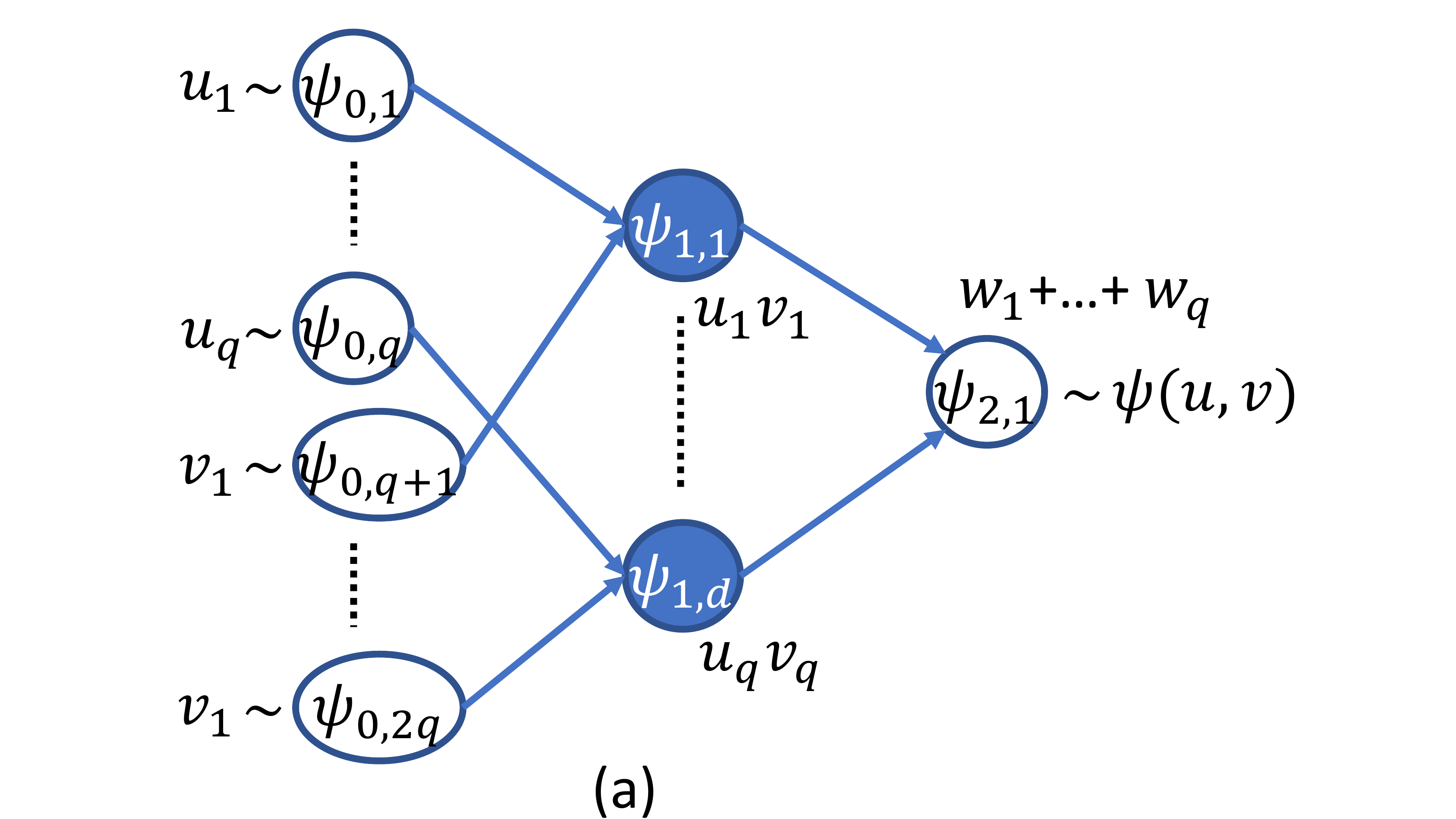} \includegraphics[width = 2.25in]{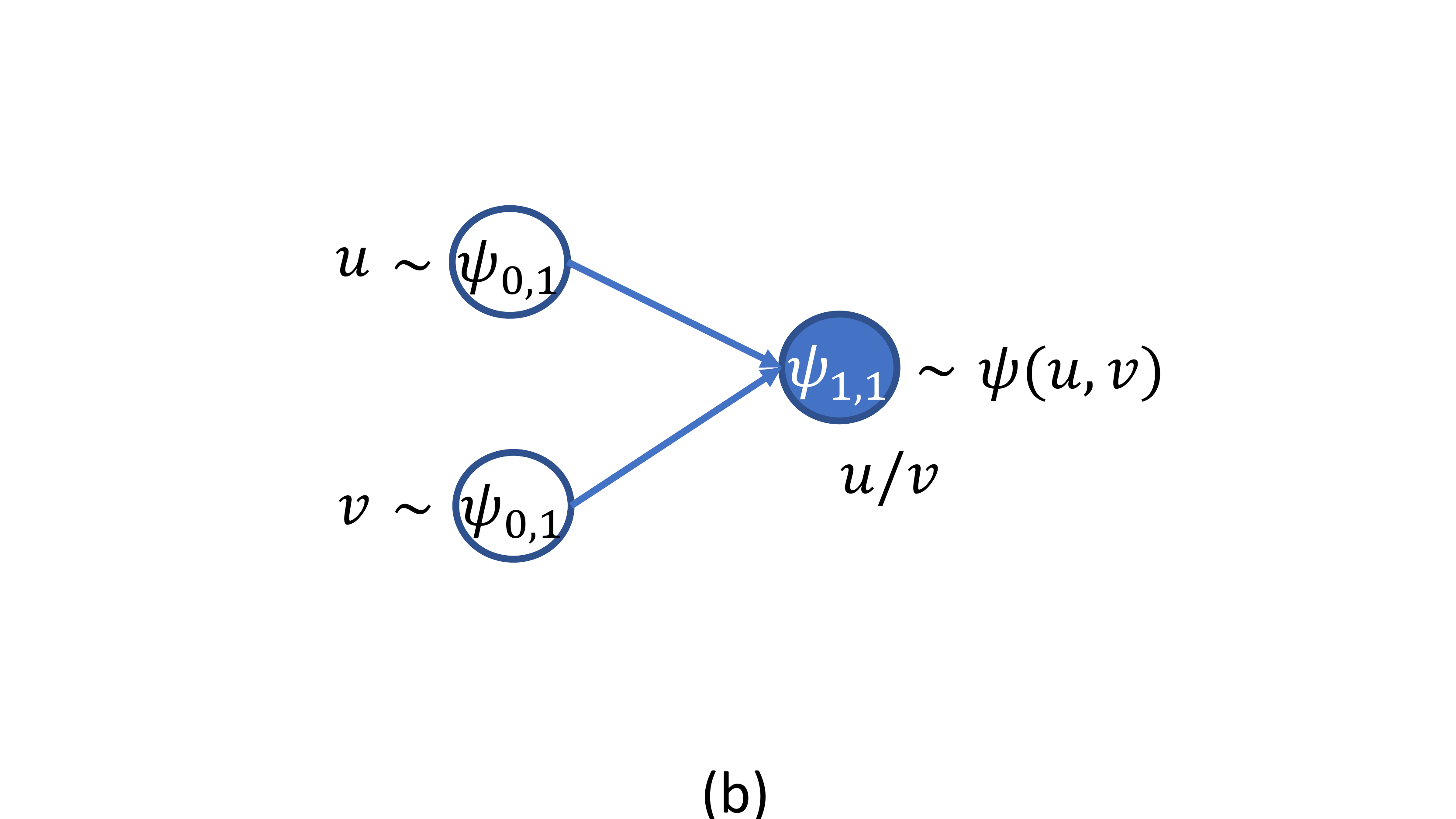} 
\caption{(a): DAG of inner product, $\bfu^T\bfv$, as a compositional function. (b) DAG of division, $u/v$, as a compositional function. }
\label{fig_Psi_1}
\end{figure}

To prove (2), we construct a compositional function $\psi: (\bfu,\bfv)\in \Real^q\times \Real^q\rightarrow \bfu^T\bfv\in \Real$ (Figure \ref{fig_Psi_1}(a)). Then
\EQ
h(\bfx)=\psi(\bff(\bfx),\bfg(\bfx)).
\EE
The Lipschitz constants associated with each general node in $\calG^\psi$ (all located in the layer $l=1$) is $L_{1,j}^\psi=1$. It is straightforward to show that
\EQ
\norm{\psi_{1,j}}_{W^\infty_{m,2}}\leq (A_jB_j+A_j+B_j+2),
\EE
for any $m\geq 1$. From Theorem \ref{thm2}, there exists a deep neural network, $\psi^{NN}$, so that 
\EQ
\abs{\psi (u,v)-\psi^{NN}(u,v)}\leq C\Lambda q (n_{width})^{-m/2},
\EE
where C is a constant depending on $m$ ($d_{i,j}=2$), $\Lambda$ is defined in (\ref{eq_psi_gamma}). Define $h^{NN}=\psi^{NN}\circ(f^{NN},g^{NN})$. It is a deep neural network. We have
\EQ
\abs{h(\bfx)-h^{NN}(\bfx)}\\
= \abs{\psi(\bff(\bfx),\bfg(\bfx))-\psi^{NN}(\bff^{NN}(\bfx),\bfg^{NN}(\bfx))}\\
= \abs{\psi(\bff(\bfx) , \bfg(\bfx))-\psi(f^{NN}(\bfx),\bfg^{NN}(\bfx))}+\abs{\psi(\bff^{NN}(\bfx),g^{NN}(\bfx))-\psi^{NN}(\bff^{NN}(\bfx),\bfg^{NN}(\bfx))}\\
\leq \abs{\bff(\bfx)^T\bfg(\bfx)-(\bff^{NN}(\bfx))^T\bfg^{NN}(\bfx)}+C\Lambda q (n_{width})^{-m/2}\\
\leq \abs{\bff(\bfx)^T\bfg(\bfx)-\bff(\bfx)^T\bfg^{NN}(\bfx)}+\abs{\bff(\bfx)^T\bfg^{NN}(\bfx)-(\bff^{NN}(\bfx))^T\bfg^{NN}(\bfx)}+C\Lambda q (n_{width})^{-m/2}\\
\leq \norm{\bff(\bfx)}_2\norm{\bfg(\bfx)-\bfg^{NN}(\bfx)}_2+\norm{\bff(\bfx)-\bff^{NN}(\bfx)}_2\norm{\bfg^{NN}(\bfx)}_2+C\Lambda q (n_{width})^{-m/2}\\
\leq \ds\max_\bfx\{\norm{\bff(\bfx)}_2\}e_2+(\ds\max_\bfx\{\norm{\bfg(\bfx)}_2\}+e_2)e_1+C\Lambda q (n_{width})^{-m/2}.
\EE

The proof of (3) is similar to the idea shown above but for $q=1$.  We construct a compositional function $\psi: (u,v)\in \Real\times \Real\rightarrow u/v\in \Real$. This is a compositional function with a single general node and two input nodes (Figure \ref{fig_Psi_1}(b)). Then
\EQ
h(\bfx)=\psi(f(\bfx),g(\bfx)).
\EE
A Lipschitz constant associated with the single general node in $\calG^\psi$ is $L_{1,1}=1$. Skipping the proof that is based on typical algebraic derivations, we claim that the norm of $\psi$ satisfies 
\EQ
A=\ds\max_\bfx\{\abs{f(\bfx)}\}, B=\ds\min_\bfx\{\abs{g(\bfx)}\},\\
\norm{\psi}_{W^\infty_{m,2}}\leq \left( m!( A+B)\Fr{1-(1/B)^{m+1}}{B-1} \right),\\
\EE
for any $m\geq 1$. From Theorem \ref{thm2}, there exists a deep neural network, $\psi^{NN}$, so that 
\EQ
\abs{\psi (u,v)-\psi^{NN}(u,v)}\leq C\Lambda (n_{width})^{-m/2},
\EE
where C is a constant depending on $m$, $\Lambda$ is defined in (\ref{eq_psi_gamma_2}). Define $h^{NN}=\psi^{NN}\circ(f^{NN},g^{NN})$. We have
\EQ
\abs{h(\bfx)-h^{NN}(\bfx)}\\
= \abs{\psi(f(\bfx),g(\bfx))-\psi^{NN}(f^{NN}(\bfx),g^{NN}(\bfx))}\\
= \abs{\psi(f(\bfx) , g(\bfx))-\psi(f^{NN}(\bfx),g^{NN}(\bfx))}+\abs{\psi(f^{NN}(\bfx),g^{NN}(\bfx))-\psi^{NN}(f^{NN}(\bfx),g^{NN}(\bfx))}\\
\leq \abs{f(\bfx)/g(\bfx)-f^{NN}(\bfx)/g^{NN}(\bfx)}+C\Lambda (n_{width})^{-m/2}\\
\leq \abs{f(\bfx)/g(\bfx)-f(\bfx)/g^{NN}(\bfx)}+\abs{f(\bfx)/g^{NN}(\bfx)-f^{NN}(\bfx)/g^{NN}(\bfx)}+C\Lambda  (n_{width})^{-m/2}\\
\leq \abs{f(\bfx)}\Fr{\abs{g(\bfx)-g^{NN}(\bfx)}}{\abs{g(\bfx)g^{NN}(\bfx)}}+\Fr{\abs{f(\bfx)-f^{NN}(\bfx)}}{\abs{g^{NN}(\bfx)}}+C\Lambda  (n_{width})^{-m/2}.
\EE
Because of (\ref{eq_e22}), we know
\EQ
\abs{g^{NN}(\bfx)}\geq  \Fr{B}{2}.
\EE
Therefore, 
\EQ
\abs{h(\bfx)-h^{NN}(\bfx)}\leq \Fr{2A}{B^2}e_2+\Fr{2}{B}e_1+C\Lambda (n_{width})^{-m/2}.
\EE
$\blacklozenge$

\section{Ordinary differential equations}
\label{sec_5}
In the following, we say that a function is $C^k$ if all derivatives of order less than or equal to $k$ are continuous in the domain. Consider a system of ordinary differential equations (ODEs) 
\EQ
\label{eq_ode1}
\dot \bfx=\bff(\bfx), & \bfx\in \Real^d, \bff(\bfx)\in \Real^d, t\in [0, T].
\EE
Without loss of generality, $\bff$  in (\ref{eq_ode1}) does not explicitly depend on $t$. If the right-hand side of the ODE is a function of $(t,x)$, one can add a new state variable to convert the equation to a time-invariant one. Let $\bfphi(t; \bfx_0)$ represent the solution of (\ref{eq_ode1}) satisfying $\bfx(0)=\bfx_0$.  In this section, we study the complexity of deep neural networks in the approximation of the function $\bfphi(T;\cdot): \bfx\rightarrow  \bfphi(T;\bfx)$. We assume that $\bff$, the right-hand side of (\ref{eq_ode1}),  is a compositional function that satisfies A1. After integration, the compositional structure of $\bfphi(t;\bfx)$ becomes unknown. 

\begin{figure}[h!]
\centering
\includegraphics[width = 2.25in]{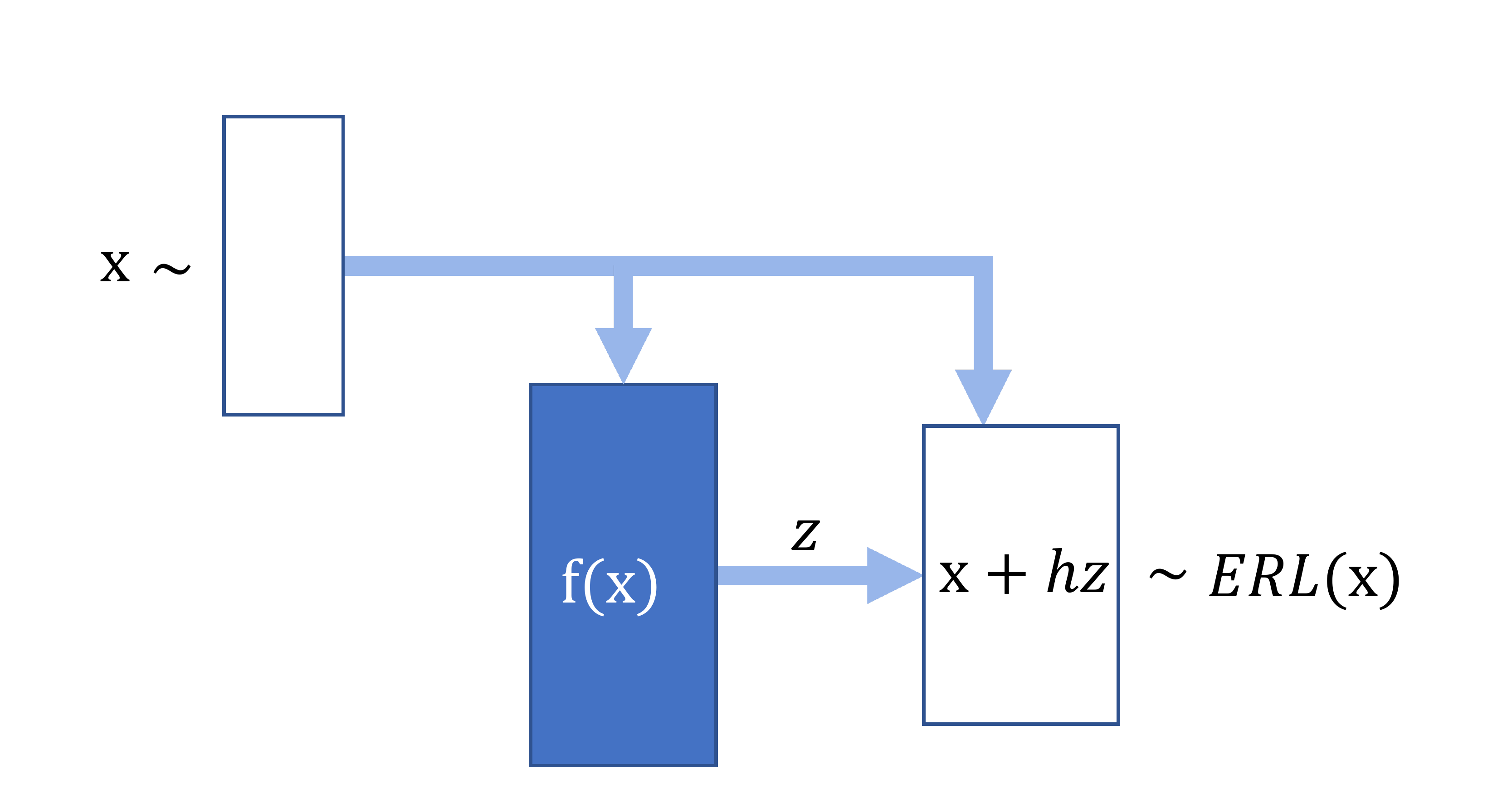}
\caption{DAG structure of an Euler iteration}
\label{fig_ELR}
\end{figure}

The solution of an ODE can be approximated using the Euler method. Let $ELR: \Real^d \rightarrow \Real^d$ represent the operator of the Euler method, i.e.,
\EQ
\label{eq_ELR}
ELR(\bfx)=\bfx+h\bff(\bfx).
\EE
Let $K>0$ be a positive integer and $h=T/K$. Then
\EQ
\label{eq_ELRN}
\bfphi(T; \bfx)\\
\approx \overbrace{ELR\circ\cdots ELR\circ ELR}^{K}(\bfx)\\
=(ELR(\cdot))^K(\bfx).
\EE
As a compositional function,  $ELR$ has a DAG, whose framework is shown in Figure \ref{fig_ELR}. The first box on the left represents the layer of input nodes. The blue box in the middle contains all nodes in $\calV^\bff\setminus \calV^\bff_I$ and their edges. The box on the right consists of output nodes, which are all linear nodes. Applying Definition \ref{def_compcomp},  $(ELR(\cdot))^K$ is a compositional function with an induced DAG.

\begin{theorem}
\label{thm4}
Consider the ODE (\ref{eq_ode1}) in which $(\bff,\calG^\bff,\calL^\bff)$ is a compositional function satisfying A1. Its features are defined in Definition \ref{def_feature}. Assume that all nodes in $\calG^\bff$ are $C^1$. Suppose $D^\bff$ is a closed set such that $\bfphi(t;\bfx) \in [-R, R]^d$ whenever $\bfx\in D^\bff$ and $t\in [0, T]$. Let $1\leq p\leq \infty$ be a real number.
\begin{enumerate}
\item There is a constant $C$ so that, for any integer $n_{width}>0$, there exists a deep neural network, $\bfphi^{NN}: \Real^d \rightarrow \Real^d$, satisfying 
\EQ
\label{eq_errorODEshort}
\norm{\bfphi(T;\bfx)-\bfphi^{NN}(\bfx)}_p \leq C(n_{width})^{-1/r^\bff_{max}} , & \bfx\in D^\bff.
\EE
\item Let $\alpha\in \Real$ be the constant so that a Lipschitz constant of $ELR$ has the form $(1+h\alpha)$. The constant, $C$, in (\ref{eq_errorODEshort}) satisfies 
\EQ
\label{eq_ODEerror}
C=C_1 \left(\max\{e^{\alpha T}, 1\}\right)T L_{max}^\bff\Lambda^\bff \abs{\calV^\bff_G}+ Ae^{BT}T,\\
A=\ds\max_{\bfx\in [-R, R]^d}\{ \norm{\bff(\bfx)}_p\},\\
B=\ds\max_{\bfx\in [-R,R]^d}\left\{\norm{\frac{\partial \bff}{\partial \bfx} (\bfx)}_p\right\},
\EE
where $C_1$ is a constant determined by $\{d_{i,j}, m_{i,j};\; f_{i,j}\in \calV^\bff_G\}$.
\item The complexity of $\bfphi^{NN}$ is bounded by
\EQ
\label{eq_complexitybound}
n\leq \left( (n_{width})^{1/r^\bff_{max}} +1\right)n_{width}\abs{\calV^\bff_G}.
\EE
\end{enumerate}
\end{theorem}

\begin{remark} The error upper bound in (\ref{eq_errorODEshort})-(\ref{eq_ODEerror}) increases at a polynomial rate with respect to the features of $\bff$. It also depends on the terms $e^{\alpha T}$ and $e^{BT}$.  If the Euler method is stable \cite{hairer}, then $\alpha <0$. The value of $e^{\alpha T}$ is always bounded. For a stable system, the error estimation using $e^{BT}$ is conservative. In this case the error does not increase exponentially with $T$ if a stable Euler method is applied.
\end{remark}

\noindent{\it Proof of Theorem \ref{thm4}}. For any integer $n_{width}>0$,  from Theorem \ref{thm2}, there exists a deep neural network approximation of $\bff$ satisfying
\EQ
\norm{\bff(\bfx)-\bff^{NN}(\bfx)}_p\leq C_1L^\bff_{max}\Lambda^\bff \abs{\calV^\bff_G}(n_{width})^{-1/r^\bff_{max}},
\EE
for some constant $C_1$ determined by $d_{i,j}$ and $m_{i,j}$. The complexity of the neural network is bounded by $n_{width}\abs{\calV^\bff_G}$.
Let $ELR^{NN}$ be the neural network obtained by substituting $\bff^{NN}$ for $f$. Then, 
\EQ
\norm{ELR(\bfx)-ELR^{NN}(\bfx)}_p,\\
=h\norm{\bff(\bfx)-\bff^{NN}(\bfx)}_p,\\
\leq hC_1L^\bff_{max}\Lambda^\bff \abs{\calV^\bff_G}(n_{width})^{-1/r^\bff_{max}}.
\EE
Define
\EQ
\bfphi^{NN}=(ELR^{NN}(\cdot))^K.
\EE
Applying Proposition \ref{prop4}, we have 
\EQ
\label{eq_errorELR2}
\norm{(ELR(\cdot))^K(\bfx)-\phi^{NN}(\bfx)}_p\\
\leq \Fr{(1+h\alpha)^K-1}{(1+h\alpha)-1} hC_1L^\bff_{max}\Lambda^\bff \abs{\calV^\bff_G}(n_{width})^{-1/r^\bff_{max}}\\
\leq  \left((1+h\alpha)^{K-1}+(1+h\alpha)^{K-2}+\cdots+1\right) hC_1L^\bff_{max}\Lambda^\bff \abs{\calV^\bff_G}(n_{width})^{-1/r^\bff_{max}}\\
\leq  K\left( \max\{(1+\frac{T}{K}\alpha)^{K-1}, 1\}\right)   \Fr{T}{K}C_1L^\bff_{max}\Lambda^\bff \abs{\calV^\bff_G}(n_{width})^{-1/r^\bff_{max}}   \\
\leq C_1 \left( \max\{e^{\alpha T}, 1\}\right) T L_{max}^\bff\Lambda^\bff \abs{\calV^\bff_G}(n_{width})^{-1/r^\bff_{max}}.
\EE
On the other hand, it is well known (see, for instance, \cite{hairer}) that the global error of the Euler method satisfies 
\EQ
\label{eq_errorELR1}
\norm{\bfphi(T;\bfx)-ELR^K(\bfx)}_p\leq Ae^{BT} \Fr{T}{K}, & \bfx\in D^\bff.
\EE
From Proposition \ref{prop_triangular} and the inequalities (\ref{eq_errorELR2}) and (\ref{eq_errorELR1}), we have
\EQ
\label{eq_ODEerror3}
\norm{\bfphi(T;\bfx)-\bfphi^{NN}(\bfx)}_p \leq C_1 \left(\max\{e^{\alpha T}, 1\}\right)T L_{max}^\bff\Lambda^\bff \abs{\calV^\bff_G}(n_{width})^{-1/r^\bff_{max}}+ Ae^{BT}T K^{-1} , & \bfx\in D^\bff.
\EE
Substituting $K$ by $K={\rm ceiling}((n_{width})^{1/r^\bff_{max}})$, where the value of ${\rm ceiling}(z)$ is the smallest integer that is greater than or equal to $z$,  we achieve the inequalities in (\ref{eq_errorODEshort})-(\ref{eq_ODEerror}) because 
\EQ
\label{eq_ceiling}
z\leq {\rm ceiling}(z) \leq z+1, &\mbox{for all } z\geq 0.
\EE
The calculation of the complexity is straightforward. There are a total of $K$ iterations in the Euler approximation. In each $\calG^{ELR}$, there is one copy of $\bff$, which is substituted by $\bff^{NN}$, a neural network whose complexity is $n_{width}\abs{\calV^\bff_G}$.  Thus, the total number of neurons in $\bfphi^{NN}$ is $Kn_{width}\abs{\calV^\bff_G}$, which implies (\ref{eq_complexitybound}) because $K\leq (n_{width})^{1/r^\bff_{max}}+1$.
$\blacklozenge$\\

\begin{example} \rm
Consider Lorenz-96 system \cite{lorenz96},
\begin{eqnarray}\label{eq:lorenz96}
\dot \bfx & = & \bff(\bfx),
\end{eqnarray}
where the vector field $\bff$ is the compositional function (\ref{eq:lorenz96_rhs}) in Example \ref{example:Lorenz96_rhs}.
For a suitable choice of the constant $F$, system (\ref{eq:lorenz96}) is chaotic; thus trajectories are bounded. Choose a sufficiently large  $R$ such that $\bfx(t)\in [-R,R]^{d}$ for all $t\in[0,T]$. 
By Theorem \ref{thm4} and the features of $\bff$ given in (\ref{eq:Lorenz96_features}), for any integers $n_{width}>0$ and $m\geq 2$, there exists a deep neural network, $\bfphi^{NN}: \Real^d \rightarrow \Real^d$, satisfying 
\EQ
\label{eq_errorODEshort_ad}
\norm{\bfphi(T;\bfx)-\bfphi^{NN}(\bfx)}_\infty \leq C(n_{width})^{-m/2},
\EE
where
\EQ
\label{eq_ODEerror_ad}
C=C_1 Td(1+2R+4R^2) \max\{e^{\alpha T}, 1\} \mbox{max}\{(2R)^m,1\} + Ae^{BT}T,\\
A=\ds\max_{\bfx\in [-R, R]^d}\{ \norm{\bff(\bfx)}_\infty\} \leq 2R^2+R+F ,\\
B=\ds\max_{\bfx\in [-R,R]^d}\left\{\norm{\frac{\partial \bff}{\partial \bfx} (\bfx)}_\infty\right\} \leq 1+4R,\\
\alpha\leq B,\\
C_1: \mbox{a constant determined by}\ $m$.
\EE
Moreover, the complexity of $\bfphi^{NN}$ is bounded by
\EQ
\label{eq_complexitybound_ad}
n\leq d\left( (n_{width})^{m/2} +1\right)n_{width}.
\EE
\end{example}

\section{Optimal control}
\label{sec_6}
Let $D$ be a bounded and closed set in $\Real^d$. Let $V(\bfx,\bfu): D\times \Real^q\rightarrow \Real$ be a cost function, where $\bfx$ is the state and $\bfu$ is the control input.  Suppose that $V$ is $C^2$. Let $H(\bfx,\bfu)$ represent the Hessian of $V$ with respect to $\bfu$, i.e.
\EQ
H(\bfx,\bfu)=\Fr{\partial^2 V}{\partial \bfu^2}.
\EE 
If $H(\bfx,\bfu)$ is positive definite, then $V$ is a convex function for every fixed $\bfx$. Consider the problem of optimization
\EQ
\label{eq_optim}
\ds\min_\bfu V(\bfx,\bfu).
\EE
Given any $\bfx\in D$, (\ref{eq_optim}) has a unique solution. The optimal control is denoted by $\bfu^\ast (\bfx)$. The minimum cost is called the value function,
\EQ
V^\ast(\bfx)=V(\bfx,\bfu^\ast (\bfx)).
\EE
Because of the convexity and the smoothness of $V$, it can be proved that $\bfu^\ast(\bfx)$ is continuous and the set of control inputs,  
\EQ
\label{eq_opitmset}
\{\bfu^\ast(\bfx);\; \bfx\in D\},
\EE
is bounded. Its diameter is denoted by 
\EQ
\gamma=\ds\max_{\bfx\in D}\{ u^\ast(x)\}.
\EE 
Let $\bfu_0 \in \Real^m$ be any fixed point satisfying
\EQ
\norm{\bfu_0-\bfu^\ast(\bfx)}_2\leq \gamma, & \bfx\in D.
\EE
In the following, we apply an iterative algorithm starting from $\bfu_0$ in the searching for $\bfu^\ast(\bfx)$. It is a fact that will become clear later that the following set is invariant under the searching process:
\EQ
\label{eq_calU}
{\cal U} = \{ \bfu\in \Real^m; \; \norm{\bfu-\bfu_0}_2\leq  2\gamma\} .
\EE
In the following, we prove an error upper bound for the approximation of $u^\ast$ using neural network.
\begin{theorem}
\label{thm7}
Suppose $V(\bfx,\bfu): D\times \calU\subset \Real^d\times\Real^q\rightarrow \Real$ is a $C^2$ function. Assume that $H(\bfx,\bfu)$ is positive definite in $D\times \calU$.  Let $\lambda_{min}>0$ and $\lambda_{max}>0$ be the lower and upper bounds of the eigenvalues of $H(\bfx,\bfu)$ in $D\times \calU$. Suppose $V^{NN}$ is a neural network satisfying 
\EQ
\abs{V(\bfx,\bfu)-V^{NN}(\bfx,\bfu)}\leq e_1, & (\bfx,\bfu)\in D\times \calU,
\EE
for some $e_1>0$. 
\begin{enumerate}
\item  Given any integer $K>0$ and real number $h>0$, there exists a deep neural network, $\bfu^{\ast NN}: \Real^d\rightarrow \Real^q$, that approximates $\bfu^\ast$ with the following error upper bound
\EQ
\label{eq_errorOptim00}
\norm{ \bfu^{\ast NN}(\bfx)-\bfu^\ast(\bfx)}_2 \leq  \gamma L^K+C_1 \sqrt{q}Kh + 2\sqrt{q} Kh^{-1}e_1, & \bfx\in D,
\EE
where $\gamma$ is the diameter of the set (\ref{eq_opitmset}), $L = 1-\lambda_{min}/\max \{ 1, 2\lambda_{max}\}$, and 
\EQ
 C_1=\ds\max_{\bfx\in D, \bfu\in \calU} \left\{\Fr{\partial^2 V}{\partial u_j^2}(\bfx,\bfu) \right\}, & 1\leq j\leq q.
\EE
\item  Let $n_1$ be the complexity of $V^{NN}$. The complexity of $\bfu^{\ast NN}$ satisfies
\EQ
\label{eq_complexityOptim}
n\leq 2n_1qK.
\EE
\end{enumerate}
\end{theorem}

\noindent{\it Proof}.
Define a function
\EQ
\label{eq_psi}
\bfpsi(\bfx,\bfu)=\bfu-\beta \Fr{\partial V}{\partial \bfu}(\bfx,\bfu),
\EE
where $\beta>0$ is chosen so that the matrix norm satisfies 
\EQ
\label{eq_Lpsi}
\norm{I-\beta \Fr{\partial^2 V}{\partial \bfu^2} }_2 \leq L <1, & \bfx\in D, \bfu\in \calU,
\EE
for some constant $L$. This is always possible because $\calU$ is bounded and $H(\bfx,\bfu)$ is positive definite and continuous. For instance, one can chose
\EQ
\beta =  \Fr{1}{\max\{1, 2\lambda_{max}\}}, 
\EE
In this case, we can prove that 
\EQ
L = 1-\Fr{\lambda_{min}}{\max \{ 1, 2\lambda_{max}\}}
\EE
satisfies (\ref{eq_Lpsi}). By the contraction mapping theorem, for every $\bfx\in D$, there is a fixed point $\bfu^\ast(\bfx)$ such that
\EQ
\bfpsi (\bfx,\bfu^\ast(\bfx))=\bfu^\ast (\bfx).
\EE
By the definition of $\bfpsi$ in (\ref{eq_psi}), this implies that
\EQ
\Fr{\partial V}{\partial \bfu} (\bfx,\bfu^\ast(\bfx))=0,
\EE
i.e., the fixed point is the optimal control. From Proposition \ref{prop_lipschitz3}, for any integer $K>0$, we have
\EQ
\label{eq_phiestimate1}
\abs{\left(\bfpsi(\bfx,\cdot)\right)^K(\bfu_0)-\bfu^\ast(\bfx)}\\
=\abs{\left(\bfpsi(\bfx,\cdot)\right)^K(\bfu_0)-\left(\bfpsi(\bfx,\cdot)\right)^K(\bfu^\ast(\bfx))}\\
\leq L ^K\norm{\bfu_0-\bfu^\ast(\bfx)}_2\\
\leq \gamma L^K.
\EE
This implies that $\left(\bfpsi(\bfx,\cdot)\right)^K(\bfu_0) \in \calU$. Let $\bfpsi^{NN}(\bfx,\bfu)$ be a deep neural network approximation (to be constructed) of $\bfpsi$ with an error bound
\EQ
\norm{\bfpsi(\bfx,\bfu)-\bfpsi^{NN}(\bfx,\bfu)}_2\leq e_2.
\EE
From Proposition \ref{prop4},
\EQ
\label{eq_phiestimate2}
\norm{\left(\bfpsi(\bfx,\cdot)\right)^K(\bfu_0)-\left(\bfpsi^{NN}(\bfx,\cdot)\right)^K(\bfu_0)}_2\\
\leq \Fr{L^K -1}{L -1}e_2.
\EE
The inequalities (\ref{eq_phiestimate1}) and (\ref{eq_phiestimate2}) and Proposition \ref{prop_triangular} imply
\EQ
\label{eq_optimerror1}
\abs{\left(\bfpsi^{NN}(\bfx,\cdot)\right)^K(\bfu_0)-\bfu^\ast(\bfx)}\\
\leq \gamma L ^K+\Fr{L^K -1}{L -1}e_2.
\EE
In the next, we construct $\bfpsi^{NN}$ and derive $e_2$. Let $h>0$ be a step size for finite difference. Then
\EQ
\abs{V_{u_j}-\Fr{1}{h}\left(V(\cdots u_j+h \cdots)-V(\cdots u_j\cdots)\right)}\leq C_1 h, &1\leq j\leq q,
\EE
where  
\EQ
C_1=\ds\max_{x\in D, u\in \calU}\left\{ \Fr{\partial^2 V}{\partial u_j^2}(\bfx,\bfu)\right\}, & 1\leq j\leq q.
\EE
Define
\EQ
\tilde \bfpsi (\bfx,\bfu)= \bfu-\beta\left[ \begin{array}{ccc}
\Fr{1}{h}\left(V(\cdots u_1+h \cdots)-V(\cdots u_1\cdots)\right)\\
\vdots\\ 
\Fr{1}{h}\left(V(\cdots u_j+h \cdots)-V(\cdots u_j\cdots)\right).\\
\vdots\end{array}\right]
\EE
Then
\EQ
\label{eq_optimerror3}
\norm{\bfpsi(\bfx,\bfu)-\tilde \bfpsi(\bfx,\bfu)}_2\leq \beta C_1\sqrt{q}h.
\EE
By assumption, $V^{NN}$ is a deep neural network that approximates $V$ satisfying
\EQ
\abs{V(\bfx,\bfu)-V^{NN}(\bfx,\bfu)} \leq e_1.
\EE
We define $\bfpsi^{NN}$ by substituting $V^{NN}$ for $V$ in $\tilde \bfpsi$. Then
\EQ
\label{eq_optimerror4}
\norm{\tilde \bfpsi(\bfx,\bfu)- \bfpsi^{NN}(\bfx,\bfu)}_2 \leq 2\Fr{\beta}{h}\sqrt{q} e_1.
\EE
From Proposition \ref{prop_triangular}, (\ref{eq_optimerror3}) and (\ref{eq_optimerror4}) imply
\EQ
\norm{\bfpsi(\bfx,\bfu)-\bfpsi^{NN}(\bfx,\bfu)}_2\\
\leq \beta  C_1\sqrt{q}h + 2\Fr{\beta}{h}\sqrt{q} e_1.
\EE
Using the right side of the above inequality as $e_2$, (\ref{eq_optimerror1}) implies
\EQ
\label{eq_errorOptim3}
\norm{\left( \bfpsi^{NN}(\bfx,\cdot)\right)^K(\bfu_0)-\bfu^\ast(\bfx)}_p\\
\leq \gamma L^K+\Fr{L^K -1}{L -1}\left( \beta  C_1\sqrt{q}h + 2\Fr{\beta}{h}\sqrt{q} e_1\right)\\
\leq \gamma L^K+K\left( \beta C_1\sqrt{q}h + 2\Fr{\beta}{h}\sqrt{q} e_1\right), & (\mbox{because } L <1)\\
\leq \gamma L^K+C_1 \sqrt{q}Kh + 2\sqrt{q} Kh^{-1}e_1,\\
\EE
The parameter $\beta$ is removed from the inequality because $\beta\leq 1$. 
If we define
\EQ
\bfu^{\ast NN}(\bfx):= \left(\bfpsi^{NN}(\bfx,\cdot)\right)^K,
\EE
then (\ref{eq_errorOptim3}) is equivalent to (\ref{eq_errorOptim00}). From the structure of $\tilde \bfpsi$, we know that the complexity of $\bfu^{\ast NN}$ is bounded by
\EQ
2n_1qK.
\EE
$\blacklozenge$

Consider a control system
\EQ
\label{eq_controlsys}
\dot \bfx=\bff(\bfx,\bfu), & \bfx\in D\subset\Real^d, \bfu\in \Real^q, t\in [0, T],
\EE
where $D \subset \Real^d$ is a bounded closed set. We assume a zero-order hold (ZOH) control input function. Specifically, given any positive integer $N_t>0$. Let $\Dlt t=T/N_t$ and $t_k=k\Dlt t$ for $0\leq k\leq N_t$. Then the control input $\bfu(t)$ is a step function that takes a constant value in each time interval $[t_{k-1}, t_{k}]$.  Let $\bfphi (t; \bfu(\cdot), \bfx_0)$ be the trajectory of the control system (\ref{eq_controlsys}) with the initial condition $\bfx(0)=\bfx_0$ and a step function $\bfu(t)$ as the control input. For a fixed $\Dlt t$, we can treat $\bfu(t)$ as a long vector $U\in \Real^{q N_t}$, where $U=\MT \bfu_1^T &\bfu_2^T&\cdots &\bfu_{N_t}^T\EM^T$ and $\bfu_k$ is the value of $\bfu(t)$, $t\in [t_{k-1}, t_k]$. Define 
\EQ
\label{eq_soln}
\begin{aligned}
\Phi(\bfx,U)&=\bfphi(T; \bfu(\cdot),\bfx)\\
&=\bfphi(\Dlt t; \bfu_{N_t},\cdot)\circ \bfphi(\Dlt t; \bfu_{N_t-1},\cdot)\circ\cdots\circ \bfphi(\Dlt t; \bfu_1,\bfx), & \bfx\in D.
\end{aligned}
\EE
If $f(\bfx,\bfu)$ is $C^2$ in $(\bfx,\bfu)$, then $\Phi(\bfx,U)$ is a $C^2$ function of $(\bfx,U)$. Let $\Psi:\Real^d\rightarrow \Real$ be a $C^2$ function. Define the cost function as follows:
\EQ
\label{eq_costJ}
J(\bfx,U)=\Psi \circ \Phi(\bfx,U).
\EE
The problem of ZOH optimal control is 
\EQ
\ds\min_{U\in \Real^{qN_t}} J(\bfx,U).
\EE
Given any $\bfx\in D$, let $U^\ast(\bfx)$ represent the optimal control. Let $\calU\subset \Real^{q N_t}$ be the set defined in (\ref{eq_calU}) where $\bfu$ is replaced by $U$.  Suppose $\bff(\bfx,\bfu)$ in (\ref{eq_controlsys}) is defined for all $\bfx\in [-R, R]^d$ where $R$ is large enough so that 
\EQ
\bfphi(\Dlt t; \bfu_{k},\cdot)\circ \bfphi(\Dlt t; \bfu_{k-1},\cdot)\circ\cdots\circ \bfphi(\Dlt t; \bfu_1,\bfx)\in [-R, R]^d, & 1\leq k\leq N_t,
\EE
for all $U\in \calU $ and $\bfx\in D$. For the simplicity of derivation in the following theorem, we assume $\Dlt t\leq 2$. 

\begin{theorem}
\label{thm8}
Suppose that $\bff$ in the control system (\ref{eq_controlsys}) and $\Psi$  in the cost function (\ref{eq_costJ}) are both compositional functions that satisfy A1.  Assume that all nodes in $\calG^\bff$ and $\calG^\Psi$ are $C^2$. Assume that the Hession of $ J (\bfx,U)$ with respect to $U$ is positive definite for all $(\bfx,U)\in D\times \calU$. Let $r_{max}=\max\{ r^\bff_{max}, r^\Psi_{max}\}$ and $\gamma$ be the diameter of $\calU$. Then, for any $\epsilon >0$, there exists a deep neural network, $U^{\ast NN}$ that approximates $U^\ast$. The estimation error is bounded by
\EQ
\norm{U^{\ast NN}(\bfx)-U^{\ast}(\bfx)}_2\leq 3\epsilon, & \bfx\in D.
\EE
The complexity of $U^{\ast NN}$ is bounded by 
\EQ
\label{eq_erroroptimalcontrolf}
n\leq C \left( q( \abs{\calV^\bff_G}+\abs{\calV^\Psi_G} )\right)^{(r_{max}+1+r_{max}/r^\bff_{max})}  \epsilon^{-(4r_{max}+1+4r_{max}/r^\bff_{max})}
\EE
for some constant $C>0$. 
\end{theorem}

\begin{remark}
The inequality (\ref{eq_erroroptimalcontrolf}) is a simplified version that contains only three explicitly presented parameters, $q$, $\calV^\bff_G$ and $\calV^\Psi_G$. For fixed $\Dlt t\leq 2$, the constant $C$ in (\ref{eq_erroroptimalcontrolf}) depends on $N_t$, $d_{i,j}^\bff$, $d_{i,j}^\Psi$ $m_{i,j}^\bff$, $m_{i,j}^\Psi$, $L^\bff_{max}$, $L^\Psi_{max}$, $\Lambda^\bff_{max}$, $\Lambda^\Psi_{max}$,  the Lipschitz constants of $\bff$, $\Psi$ and $\bfphi(\Dlt t; \cdot, \cdot)$ with respect to $\bfx$ for $(\bfx,\bfu)\in D\times \calU$, the norms of $\bff$ and its Jacobian, the eigenvalues and diagonal entries of the Hession matrix of $J(\bfx,U)$, and the diameter of $\calU$. We avoid showing an explicit expression of $C$ as a function of these parameters because such a formula would be too long. However, following the proof, an algebraic expression of the complexity upper bound in terms of all these parameters can be derived in detail if needed. 
\end{remark}

\noindent{\it Proof}. 
Suppose we can find a deep neural network, $J^{NN}(\bfx,U)$ (to be found later), that approximates $J(\bfx,U)$ with an error upper bound $e_1$, i.e.,
\EQ
\label{eq_JNN_e1}
\abs{J^{NN}(\bfx,U) -J(\bfx,U)}\leq e_1, & \bfx\in D, U\in \calU.
\EE
From Theorem \ref{thm7}, there is a deep neural network, $U^{\ast NN}$, that approximates the optimal control $U^\ast$ with an error upper bound
\EQ
\label{eq_erroru0}
\norm{U^{\ast NN}(\bfx)-U^\ast(\bfx)}_2\leq \gamma L^K+C_1 \sqrt{qN_t}Kh + 2\sqrt{qN_t}Kh^{-1}e_1, & \bfx\in D,
\EE
where $0< L<1$ is a constant determined by the eigenvalues of the Hession of $J(\bfx,U)$, $\gamma$ is the diameter of the set $\{ U^\ast(\bfx), \bfx\in D\}$ and $C_1$ depends on the diagonal of the Hession of $J(\bfx,U)$. Given any $\epsilon >0$. For the first term to be bounded by $\epsilon$, we need
$\gamma L^K\leq \epsilon$. This is obvious if $\epsilon \geq \gamma$, because $L<1$. If $\epsilon < \gamma$, $\gamma L^K= \epsilon$ implies
\EQ
K=\Fr{\ln \Fr{\gamma}{\epsilon}}{\ln \Fr{1}{L}}\leq \left(\ln \Fr{1}{L}\right)^{-1}\gamma \Fr{1}{\epsilon}
\EE
We can choose
\EQ
K= \bar C_1 \epsilon^{-1}
\EE
for some constant $\bar C_1>0$ that is determined by $\gamma$ and the eigenvalues of the Hessian of $J(\bfx,U)$. For the second term in (\ref{eq_erroru0}) to be bounded by $\epsilon$, we have
\EQ
C_1 \sqrt{qN_t}Kh =  C_1\bar C_1 \sqrt{qN_t} \epsilon^{-1} h\leq \epsilon.
\EE
We choose
\EQ
h= \Fr{\bar C_2}{\sqrt{qN_t}}\epsilon^2
\EE
for $\bar C_2=\Fr{1}{C_1\bar C_1}$. For the third term in (\ref{eq_erroru0}) to be less than or equal to $\epsilon$, we have
\EQ
2\sqrt{qN_t}Kh^{-1}e_1= 2\sqrt{qN_t}\bar C_1 \epsilon^{-1}  \Fr{\sqrt{qN_t}}{\bar C_2}\epsilon^{-2} e_1\leq \epsilon,
\EE
or  
\EQ
\label{eq_erroru2}
e_1\leq \Fr{\bar C_3}{qN_t}\epsilon^4,
\EE
where $\bar C_3=\Fr{1}{2C_1\bar C_1^2}$. To summarize, we proved that the existence of $J^{NN}$ satisfying (\ref{eq_JNN_e1}) and (\ref{eq_erroru2}) implies
\EQ
\norm{U^{\ast NN}(\bfx)-U^\ast(\bfx)}_2\leq  3\epsilon, & \bfx \in D
\EE
Now, we need to prove the existence of a $J^{NN}$ whose estimation error, $e_1$, satisfies (\ref{eq_erroru2}). Suppose we have deep neural networks that approximates $\Psi(\cdot)$ and $\bfphi(\Dlt t; \cdot,\cdot)$. Let $J^{NN}$ be the deep neural network obtained by substituting functions in (\ref{eq_soln})-(\ref{eq_costJ}) by $\Psi^{NN}(\cdot)$ and $\bfphi^{NN}(\Dlt t; \cdot,\cdot)$. From Proposition \ref{prop4}, the estimation error, $e_1$, of $J^{NN}$ is bounded by 
\EQ
e_1\leq \bar C_4 \ds\max_{\bfx\in [-R, R]^d, \bfu\in \calU}\norm{\bfphi(\Dlt t; \bfu,\bfx)-\bfphi^{NN}(\Dlt t; \bfu,\bfx)}_2+ \ds\max_{\bfz\in [-R, R]^d} \norm{\Psi(\bfz)-\Psi^{NN}(\bfz)}_2,
\EE
where $\bar C_4$ depends on $N_t$, and the Lipschitz constants of $\bfphi(\Dlt t, \cdot, \cdot)$ and $\Psi$ with respect to $\bfx$. From Theorem \ref{thm2} and Theorem \ref{thm4}, we can find $\Psi^{NN}(\cdot)$ and $\bfphi^{NN}(\Dlt t; \cdot,\cdot)$ so that
\EQ
\label{eq_erroru3}
e_1\leq \bar C_5\abs{\calV^\Psi_G} (n_{width})^{-1/r^\Psi_{max}}+\bar C_6 \abs{\calV^\bff_G}\Dlt t(n_{width})^{-1/r^\bff_{max}}\leq \bar C_7 \left( \abs{\calV^\Psi_G}+ \abs{\calV^\bff_G}\Dlt t \right)(n_{width})^{-1/r_{max}},
\EE
where $\bar C_7$ depends on $\bar C_4$, $d_{i,j}^\bff$, $d_{i,j}^\Psi$ $m_{i,j}^\bff$, $m_{i,j}^\Psi$, $L^\bff_{max}$, $L^\Psi_{max}$, $\Lambda^\bff_{max}$, $\Lambda^\Psi_{max}$ and the norms of $\bff$ and its Jacobian. It also depends on $e^{\alpha\Dlt t}$ (see (\ref{eq_ODEerror})). However, for a fixed $\Dlt t$, this term has an upper bounded determined by the norm of the Jacobian of $\bff$. To satisfy (\ref{eq_erroru2}), (\ref{eq_erroru3}) implies that we can choose 
\EQ
\label{eq_C9}
n_{width} = \bar C_8 q^{r_{max}} \left( \abs{\calV^\Psi_G}+ \abs{\calV^\bff_G}\Dlt t \right)^{r_{max}}\epsilon^{-4r_{max}},
\EE 
where $\bar C_8=(N_t \bar C_7/\bar C_3)^{r_{max}}$.
From Theorem \ref{thm2}, Theorem \ref{thm4} and (\ref{eq_C9}), the complexity of $J^{NN}$, denoted by $n_1$, satisfies
\EQ
n_1 \leq N_t((n_{width})^{1/r^\bff_{max}}+1)n_{width}\abs{\calV^\bff_G}+n_{width}\abs{\calV^\Psi_G}\\
\leq  (n_{width})^{1+1/r^\bff_{max}}\left(2N_t\abs{\calV^\bff_G}+\abs{\calV^\Psi_G}\right)\\
\leq \bar C_{9} q^{r_{max}(1+1/r^\bff_{max})}\left( \abs{\calV^\Psi_G}+ \abs{\calV^\bff_G}\Dlt t \right)^{r_{max}(1+1/r^\bff_{max})}\left(2N_t\abs{\calV^\bff_G}+\abs{\calV^\Psi_G}\right) \epsilon^{-4r_{max}(1+1/r^\bff_{max})}\\
\leq  \Fr{\bar C_9}{2N_t} q^{r_{max}(1+1/r^\bff_{max})}\left( \abs{\calV^\Psi_G}+ \abs{\calV^\bff_G} \right)^{r_{max}(1+1/r^\bff_{max})+1} \epsilon^{-4r_{max}(1+1/r^\bff_{max})}, 
\EE
where $\bar C_{9}=\bar C_{8}^{1+1/r^\bff_{max}}$. From (\ref{eq_complexityOptim}) in Theorem \ref{thm7}, the complexity of $U^{\ast NN}$ is bounded by
\EQ
n\leq 2n_1qN_tK\\
\leq  2  \Fr{\bar C_{9}}{2N_t} q^{r_{max}(1+1/r^\bff_{max})}\left(\abs{\calV^\bff_G}+\abs{\calV^\Psi_G}\right)^{r_{max}(1+1/r^\bff_{max})+1}\epsilon^{-4r_{max}(1+1/r^\bff_{max})}q N_t  \bar C_1 \epsilon^{-1}\\
\leq C q^{(r_{max}+1+r_{max}/r^\bff_{max})} \left(\abs{\calV^\bff_G}+\abs{\calV^\Psi_G}\right)^{(r_{max}+1+r_{max}/r^\bff_{max})} \epsilon^{-(4r_{max}+1+4r_{max}/r^\bff_{max})}
\EE
where $C=\bar C_1 \bar C_9$. 

\section{Conclusions}
In this paper, we introduce new concepts and develop an algebraic framework for compositional functions. In addition to functions defined in $\Real^d$, iterative numerical algorithms and dynamical systems fall into the category of compositional functions as well. The algebraic framework forms a unified theoretical foundation that supports the study of neural networks with applications to a variety of problems including differential equations, optimization, and optimal control. It is proved that  the error upper bound of approximations using neural networks is determined by a set of key features of compositional functions. In the approximation of functions, differential equations and optimal control, the complexity of neural networks is bounded by a  polynomial function of the key features and error tolerance. The results could also shed light on the reason why using neural network approximations may help to avoid the curse of dimensionality. As shown in examples, increasing the dimension of a system may not increase the value of the key features.  

The study in this paper raises more questions than answers. All neural networks used in the proofs of theorems have layered DAGs that are similar to the DAG of the function being approximated. What is the underlying connection between the shallow neural networks that approximate individual nodes and the neural network that approximates the overall function? How can this connection help to generate initial guess of parameters for neural network training? Instead of $C^1$ functions, to what extent one can apply the idea in this paper to a larger family of functions to include $C^0$ or even discontinuous functions? This is important because optimal control problems with constraints are likely to have $C^0$ or discontinuous feedbacks. The DAG associated with a given function is not unique. What types of DAGs are advantageous for more accurate approximation with less complexity? In general, many problems of nonlinear control such as finding Lyapunov functions, designing output regulators, $H_\infty$ control and differential game do not have tractable numerical solutions if the dimension of state space is high. Finding compositional function representations of these problems may lead to innovative and practical solutions based on deep learning.

\end{document}